\documentclass[times,twocolumn,final,authoryear]{elsarticle}

\usepackage{prletters}
\usepackage{framed,multirow}

\usepackage{amssymb}
\usepackage{latexsym}

\usepackage[usenames,dvipsnames,svgnames,table]{xcolor}
\usepackage[hyphens]{url}
\usepackage[hidelinks]{hyperref}

\definecolor{newcolor}{rgb}{.8,.349,.1}

\journal{Pattern Recognition Letters}

\usepackage{mathtools,cuted}

\usepackage{graphicx}
\usepackage{rotating}
\usepackage{bbm}
\usepackage{latexsym}
\usepackage[eulergreek]{sansmath} 
\usepackage{amsmath}               
\usepackage{amssymb}               
\usepackage{psfrag} 
\usepackage{wasysym}
\usepackage{subcaption}
\usepackage{adjustbox}
\usepackage{balance}

\usepackage{tikz}
\usetikzlibrary{decorations.pathreplacing}
\usetikzlibrary{decorations.markings}
\usetikzlibrary{patterns}
\usetikzlibrary{positioning}
\usetikzlibrary{shapes,arrows}
\usetikzlibrary{calc}
\usetikzlibrary{intersections}
\usetikzlibrary{plotmarks}
\usepackage{pgfplots}
\usepgflibrary{plotmarks}
\pgfplotsset{compat=1.12}
\usetikzlibrary{pgfplots.groupplots}
\usetikzlibrary{external}
\usetikzlibrary{arrows.meta}

\usepackage[ruled,vlined]{algorithm2e}

\usepackage{booktabs}      
\usepackage{multirow}

\definecolor{brightred}{rgb}{1.0,0.1,0.1}
\definecolor{brightblue}{rgb}{0.0,0.0,0.8}
\definecolor{darkblue}{rgb}{0.0,0.0,0.5}
\definecolor{darkgreen}{rgb}{0.0,0.3,0.0}
\definecolor{brightgreen}{rgb}{0.0,0.8,0.0}
\definecolor{darkblack}{rgb}{0.0,0.0,0.0}
\definecolor{grey}{rgb}{0.3,0.3,0.3}

\newcommand{\changedOld}[1]{#1}

\newcommand{\changed}[1]{#1}

\newcommand{\removed}[1]{}

\newcommand{\mycomment}[1]{}

\graphicspath{{./}}
\newcommand\numberthis{\addtocounter{equation}{1}\tag{\theequation}}

\newcommand{\dblone}{\hbox{$1\hskip -1.2pt\vrule depth 0pt height 1.6ex width 0.7pt\vrule depth 0pt height 0.3pt width 0.12em$}}

\newcommand{\Ncal}{{\cal N}}

\newcommand{\FF}{{\cal F}}

\newcommand{\HH}{{\cal H}}

\newcommand{\JJ}{{\cal J}}

\newcommand{\LL}{{\cal L}}

\newcommand{\RRR}{\mathbbm{R}}

\newcommand{\eps}{\epsilon}

\newcommand{\ct}{\tilde{c}}

\newcommand{\muVec}{\vec{\mu}}

\newcommand{\disT}{\textstyle}
\newcommand{\disS}{\displaystyle}

\newcommand{\DKL}[2]{D_{\mathrm{KL}}\big(#1,#2\big)}

\newcommand{\yVec}{\vec{y}}
\newcommand{\yVecN}{\vec{y}^{\,(n)}}

\newcommand{\backHalf}{\hspace{-1.5mm}}
\newcommand{\backOne}{\hspace{-3mm}}

\newcommand{\KK}{{\cal K}}
\newcommand{\KKn}{\KK^{(n)}}

\newcommand{\KKNew}{\KK^{\mathrm{new}}}

\newcommand{\KKOld}{\KK^{\mathrm{old}}}

\newcommand{\comment}[1]{}

\newcommand{\qn}{q^{(n)}}

\newcommand{\beq}{\begin{equation}}
\newcommand{\eeq}{\end{equation}}
\newcommand{\beqo}{\begin{displaymath}}
\newcommand{\eeqo}{\end{displaymath}}
\newcommand{\bea}{\begin{eqnarray}}
\newcommand{\eea}{\end{eqnarray}} 
\newcommand{\beao}{\begin{eqnarray*}}
\newcommand{\eeao}{\end{eqnarray*}}

\newcommand{\One}{\dblone}

\newcommand{\BOX}{$\square$}

\newcommand{\myvanish}[1]{}

\newcommand{\algBreak}{\vspace{1.5mm}\\}

\newcommand{\rcn}{r_c^{(n)}}

\newcommand{\rCN}{r_{1:C}^{(1:N)}}
\newcommand{\scn}{s_c^{(n)}}

\newcommand{\qcn}{q_c^{(n)}}

\newcommand{\qCN}{q_{1:C}^{(1:N)}}

\newcommand{\cNew}{c^{\mathrm{new}}}

\newcommand{\remove}[1]{}

\begin{document}

\begin{frontmatter}

\title{$k$-means as a variational EM approximation of Gaussian mixture models\vspace{-2mm}}

\author[1]{J\"org L\"ucke\corref{cor1}\vspace{-1mm}} 
\cortext[cor1]{Corresponding author: 
  Tel.: +49-441-798-5486;  
  fax: +49-441-798-3902;}
\ead{joerg.luecke@uni-oldenburg.de}
\author[1]{Dennis Forster}

\address[1]{Machine Learning Lab, University of Oldenburg, Ammerl\"ander Heerstr. 114-118, 26129 Oldenburg, Germany}

\received{}
\finalform{}
\accepted{}
\availableonline{}
\communicated{}

\begin{abstract}
We show that $k$-means (Lloyd's algorithm) is obtained as a special case when truncated variational EM approximations are applied to Gaussian Mixture Models (GMM)
with isotropic Gaussians.
In contrast to the standard way to relate $k$-means and GMMs, the provided derivation shows that it is not required to consider Gaussians with small variances
or the limit case of zero variances. There are a number of consequences that directly follow from our approach: (A)~$k$-means can be shown to increase a free energy associated with truncated distributions
and this free energy can directly be reformulated in terms of the $k$-means objective; (B)~$k$-means generalizations can directly be derived by considering the 2nd closest, 3rd closest etc.\ cluster in
addition to just the closest one; and (C)~the embedding of $k$-means into a free energy framework allows for theoretical interpretations of other $k$-means generalizations in the literature.
In general, truncated variational EM provides a natural and rigorous quantitative link between $k$-means-like clustering and GMM clustering algorithms which may be very relevant for future theoretical and empirical studies.
\end{abstract}


\end{frontmatter}
%
%
%
\setlength{\abovedisplayskip}{6pt}
\setlength{\belowdisplayskip}{6pt}
%
\section{Introduction}
\label{SecIntro}
Clustering is the task of associating a set of $N$ data points with a set of $C$ clusters (typically with $C\ll{}N$), where
such an association is defined by a high similarity of points within one cluster compared to the similarity of any two points of
different clusters. Different criteria for data point similarity and
different algorithmic properties have led to the development of a large variety of clustering algorithms in the course of more than half a century.
Two of the presumably most influential classes of algorithms are $k$-means-like algorithms \citep[][and many more]{Lloyd1982,Jain2010} and Gaussian Mixture
Models (GMMs).

{\em $k$-means.} The $k$-means algorithm and its many variants \citep[e.g.][]{Steinley2006} have been used since the 1950's and are often considered
as the most popular clustering algorithms \citep[][]{Berkhin2006}. If we denote by $\yVec^{\,(1:N)}=\yVec^{\,(1)},\ldots,\yVec^{\,(N)}$ the data points \changed{(with $\yVecN\in\RRR^D$}) and
by $\muVec_{1:C}=\muVec_1,\ldots,\muVec_C$ the cluster centers \changed{(with $\muVec_c\in\RRR^D$}), \changed{then} the most common form of $k$-means is given by Alg.\,\ref{AlgKMeans}, with 
\mbox{$\|\cdot\|$} as Euclidean metric.
After initialization of $\muVec_{1:C}$, Alg.\,\ref{AlgKMeans} increases the $k$-means objective given by
%
%
%
\begin{align}
\ \nonumber\\[-8mm]
\JJ(s^{(1:N)}_{1:C},\muVec_{1:C}) = \disS\sum_{n=1}^{N}\sum_{c=1}^C \scn\, \| \yVecN \,-\,\muVec_c \|^{2}\,.  \label{EqnKMObjective}
%
%
\end{align}
The updates of $\scn$ and $\muVec_c$ in Alg.\,\ref{AlgKMeans} are usually derived from (\ref{EqnKMObjective}).
%
Because of its few elementary algorithmic steps, $k$-means is easy to implement, and it has been observed to work very well in practice \citep[e.g.][]{Duda2007}.\vspace{-4mm} 
\SetAlCapHSkip{0.2em}
\DecMargin{-0.2em}
%
%
%
\begin{algorithm}[h!]\vspace{0.5mm}
%
\Repeat{$\muVec_{1:C}$ have converged\vspace{1.5mm}}
{
\For{$c=1,\ldots,C$ and $n=1,\ldots,N$}{
$
\scn = \left\{\begin{array}{cl}
1 & \mbox{if }\forall{}c'\neq{}c:\| \yVecN -\muVec_c \|\hspace{0mm} < \hspace{0mm}\| \yVecN -\muVec_{c'} \|\\[1mm]
0 & \mbox{otherwise;}
\end{array}
\right.\hspace{-4mm}
$
}
\For{$c=1,\ldots,C$}{
$
%
\muVec_c = \sum_{n=1}^N\scn\,\yVecN/\sum_{n=1}^N\scn
$;
}
}
\caption{$k$-means. \label{AlgKMeans}}
\end{algorithm}
\ \\[-9mm]

{\em GMM.} GMM-based clustering algorithms \citep[e.g.][]{McLachlanBasford1988} are derived from a probabilistic data model $p(\yVec\,|\,\Theta)$.
While general GMMs allow for different mixing proportions and multivariate Gaussian distributions, we will for the purposes of this study consider
equal mixing proportions and equally sized, isotropic Gaussians:
\begin{align}
%
%
%
%
\backHalf{}p(c\,|\,\Theta) &= \frac{1}{C}\,,\ \ \disS p(\yVec\,|\,c,\Theta) = \disT (2\pi\sigma^2)^{-\frac{D}{2}} \exp\!\big(\hspace{-0.8mm}-\frac{1}{2\sigma^2}\|{}\yVec-\muVec_c\|{}^2\big),\label{EqnGMMIso}
\end{align}
i.e., we will use a `flat' prior $p(c\,|\,\Theta)$ and equal and isotropic variance $\sigma^{2}$ of the clusters.
The most standard form to update the GMM model parameters $\Theta=(\muVec_{1:C},\sigma^2)$ is derived using expectation maximization \citep[EM;][]{DempsterEtAl1977}, which results for GMM  (\ref{EqnGMMIso}) in Alg.\,~\ref{EqnGMMAlgo} \citep[][\& refs.\,therein]{Barber2012}.
%
%
\vspace{-1mm}
\begin{algorithm}[h]\vspace{1.5mm}
%
\Repeat{parameters $\Theta$ have converged\vspace{1.5mm}}
{
\For{$c=1,\ldots,C$ and $n=1,\ldots,N$}{
$
\rcn = \frac{\disT \exp\!\big(-\frac{1}{2\sigma^2}\|{}\yVecN-\muVec_c\|{}^2\big)} {\disT\sum_{c'=1}^C\exp\!\big(-\frac{1}{2\sigma^2}\|{}\yVecN-\muVec_{c'}\|{}^2\big)}; \phantom{iiii} \mbox{}
$
}
\For{$c=1,\ldots,C$}{
$
\begin{array}{lll}
%
\muVec_c &= \sum_{n=1}^N\rcn\yVecN/\sum_{n=1}^N\rcn; &\backHalf\mbox{}\algBreak
%
%
\end{array}
$
}
$
\begin{array}{lll}
%
\sigma^2 &= \frac{1}{DN}\,\!\!\sum_{n,\,c=1}^{N,C}\rcn\,\|\yVecN-\muVec_c\|^2;&\backHalf\mbox{}\algBreak
\end{array}
$
}
\caption{EM for GMM. \label{AlgGMM}\label{EqnGMMAlgo}}
\end{algorithm}
\vspace{-2mm}

After initialization of $\Theta=(\muVec_{1:C},\sigma^2)$, the algorithm maximizes the data log-likelihood given by:
\begin{align}
\LL(\Theta) = \frac{1}{N}\sum_{n=1}^N\log\!\Big(\sum_{c=1}^C \frac{1}{C}\,\Ncal\big(\yVecN;\muVec_c,\sigma^2\One\big)\Big)\,,\label{EqnGMMLikelihood}
%
%
\end{align}
with $\Ncal(\yVecN;\muVec_c,\sigma^2\One)$ as given in (\ref{EqnGMMIso}).
Note that (\ref{EqnGMMLikelihood}) is normalized by the number of data points for this study.
As is customary for GMMs, we refer to the posteriors $p(c\,|\,\yVecN,\Theta)$ as {\em responsibilities} (abbreviate by $\rcn$). Computing the $\rcn$ in Alg.\,\ref{AlgGMM} is referred to as {\em E-step}, while updates of parameters $\muVec_{1:C}$ and $\sigma^2$ in Alg.\,\ref{AlgGMM} are referred to as {\em M-step}.\\[3mm]
%
%
%
{\bf Related Work and Our Contribution.}
The popularity of $k$-means and GMM algorithms has resulted in many theoretical as well as empirical studies
of their functional and theoretical properties.
Considerable progress using novel versions could be made, and much insight could be gained for $k$-means \citep[][]{HarPeledSadri2005,ArthurVassilvitskii2006,ArthurEtAl2009,BachemEtAl2016b}
and GMMs \citep[][]{ChaudhuriEtAl2009,KalaiEtAl2010,MoitraValiant2010,BelkinSinha2010,XuHsuMaleki2016} relatively recently.
%
%
Because of their similarity, $k$-means and GMMs have long been formally related to each other. It is thus well-known \citep[see, e.g.][\& refs.\ therein]{MacKay2003,Barber2012}
that $k$-means (Alg.\,1) can be obtained as a limit case of EM for GMM (\ref{EqnGMMIso}).
This limit is given by considering increasingly small $\sigma^2$, i.e., $\sigma^2\rightarrow{}0$.
The responsibilities $\rcn$ in Alg.\,\ref{AlgGMM} then become equal to one for the closest cluster and zero otherwise, and the $k$-means algorithm (Alg.\,\ref{AlgKMeans}) is recovered.
Furthermore, approaches using algorithms which modify EM algorithms by introducing additional `hard' assignment steps of data points to clusters have been used to relate $k$-means and
GMM clustering. Given a generative model, such approaches are often referred to as `hard' EM \citep[e.g.][]{SegalEtAl2002,OordEtAl2014}, as `classification EM' \citep[CEM; e.g.][]{CeleuxGovaert1992} for GMMs, or as `Viterbi training' for HMMs \citep[e.g.][]{AllahverdyanGalstyan2011}. For data distributions with negligible cluster overlap (a setting which is closely related to the limit $\sigma^2\rightarrow{}0$), `hard' assignment algorithms can be shown to be equivalent to standard EM \citep[e.g.][]{CeleuxGovaert1992}. `Hard' assignments can also be informally interpreted as a variational approach but (to the knowledge of the authors) neither proofs nor quantitative results have been provided (compare Suppl.\,\ref{sec:Difference}). 
\changedOld{In contrast to `hard' cluster assignments,} the assignment is `soft' in EM for GMMs. `Hard' assignments have sometimes been considered disadvantageous
as the relative importance of the clusters for the data points is not taken into account. Different $k$-means generalizations have therefore been suggested, \changed{e.g.}, with aims to 
enhance $k$-means \changed{convergence} \cite[][]{HarPeledSadri2005} or to relax its `hard' cluster assignment \changedOld{\citep[e.g.][]{Bezdek1981,CeleuxGovaert1992,MacKay2003,MiyamotoEtAl2008}}.
\changed{As for clustering in general, $k$-means also remained of interest in the probabilistic Machine Learning community, and notably in the field of non-parametric approaches.
\citet[][]{WellingKurihara2006} suggested `Bayesian $k$-means', for instance, and used variational Bayesian approximations in order to obtain $k$-means-like run time
behavior for model selection. Later on, \citet[][]{KulisJordan2011} also used a Bayesian treatment, and combined it with the relation of $k$-means to GMMs obtained in the
limit $\sigma^2\rightarrow{}0$. In this way they derived new `hard assignment' algorithms based on a Gibbs sampler used within a non-parametric approach \citep[also compare][]{BroderickEtAl2013}.}
%
%

In this work, we derive the $k$-means algorithm from a novel class of variational EM algorithms applied to GMMs. Most notably $k$-means is obtained cleanly and rigorously without any assumptions on $\sigma^2$. 
%
%
Variational EM seeks to optimize a lower bound (the free energy; \citeauthor{NealHinton1998},\citeyear{NealHinton1998}) of the data log-likelihood by making use of variational distributions that approximate full posterior probabilities. \changed{The free energy is also frequently referred to as the evidence lower bound \citep[ELBO; e.g.][]{HoffmanEtAl2013}.}
For our study, we apply truncated posteriors \citep[][]{LuckeEggert2010} as variational distributions in their fully variational formulation \citep[][]{Lucke2018}.
After having shown that $k$-means is a variational approximation, $k$-means and its generalizations can be quantitatively related to GMMs without taking the limit to zero cluster variances or without assuming $\sigma$ to be small compared to cluster-to-cluster distances.
Furthermore, the observation that $k$-means is a variational optimization implies that it optimizes a lower bound of a GMM log-likelihood.
Hence, we can derive lower free energy bounds for $k$-means and its generalizations that quantify the link between the $k$-means and the GMM objective. As
such we provide a closer theoretical link between these two central classes of clustering \changed{methods} than has previously been established.

Truncated approaches have been applied to mixture models before. Work by \citet[][]{DaiLucke2014} used truncated approximations for a position invariant mixture model, and \citet[][]{ForsterEtAl2018} for a hierarchical Poisson mixture. Work by \citet[][]{SheltonEtAl2014} was the first to apply truncated EM to standard GMMs, followed by \citet[][]{HughesSudderth2016} who additionally used a constraint likelihood optimization to find cluster centers for truncated posteriors. None of these contributions has derived $k$-means as a variational EM algorithm for GMMs nor did any contribution provide quantitative free energy results or the links to generalizations of $k$-means derived in this study.
\section{Truncated variational EM and GMMs}
\label{SecTGMM}
The basic idea of truncated EM is the use of truncated approximations of exact posterior distributions \citep[e.g.][]{LuckeEggert2010,SheikhEtAl2014}.
In the notation as used for GMMs above, the truncated approximation takes the form:
\begin{align}
\rcn \approx \qcn = \frac{p(c,\yVecN\,|\,\Theta)}{\disT\sum_{c'\in\,\KKn}p(c',\yVecN\,|\,\Theta)}\, \delta(c\in\KKn)\,,
\label{EqnTruncated}
\end{align}
where $\KKn$ is a set of cluster indices (containing different clusters $c$ associated with data point $\yVecN$). Suppl.\,\ref{sec:Illustration} and Fig.\,\ref{FigVarGMM} provide an example.
The set of all $\KKn$ we denote by $\KK$, i.e., $\KK=(\KK^{(1:N)})$. As is customary for truncated distributions \citep[][]{LuckeEggert2010,DaiLucke2014,SheltonEtAl2014,HughesSudderth2016}, we take
the sizes of all $\KKn$ to be equal, $|\KKn|=C'$, with $1\leq C'\leq C$. The truncated approximation (\ref{EqnTruncated}) is a good approximation if $\KKn$ contains all those clusters with significant posterior mass $p(c\,|\,\yVecN,\Theta)$ (i.e., significant non-zero responsibilities $\rcn$). Truncated approaches can represent very accurate approximations for many data sets, as typically most responsibilities are negligible.

In order to derive a learning algorithm for GMMs based on truncated distributions, we have to answer the question how the parameters $\KKn$ and $\Theta$ are to be updated.
For our purposes we will here make use of a recent study which addressed this question for general models (with discrete latents) by embedding truncated distributions into a fully variational optimization framework \citep[][]{Lucke2018}. More specifically, we use the result of \citet[][]{Lucke2018} that the free energy as a lower bound of the data likelihood is monotonically increased if: (A)~the parameters $\Theta$ are updated using standard M-steps, with exact posteriors replaced by
truncated posteriors; and (B)~that the sets $\KKn$ can be found using a simplified expression for the free energy.

For GMMs, this means that we can use the standard M-steps of Alg.\,\ref{EqnGMMAlgo} and replace $\rcn$ with the truncated approximations $\qcn$
in (\ref{EqnTruncated}). For the GMM (\ref{EqnGMMIso}), the truncated responsibilities and M-steps are thus:
\begin{align}
 \qcn &= \frac{\disT \exp\!\big(-\frac{1}{2\sigma^2}\|{}\yVecN-\muVec_c\|{}^2\big)} {\disT\sum_{c'\in\KKn}\exp\!\big(-\frac{1}{2\sigma^2}\|{}\yVecN-\muVec_{c'}\|{}^2\big)}\,\delta(c\in\KKn) \label{EqnQCN} \\
\muVec^{\mathrm{\,new}}_c &= \frac{\sum_{n=1}^N{}\qcn\yVecN}{\sum_{n=1}^N{}\qcn} 
%
, \,\,\,\,\, \sigma_{\mathrm{new}}^2 = \frac{1}{DN}\!\!\sum_{n,c=1}^{N,C}\qcn\,\|\yVecN-\muVec^{\mathrm{\,new}}_c\|^2 \label{EqnMStep} 
%
%
\end{align}

The parameters $\KKn$ of the truncated distributions $\qcn$ have to be found in the variational E-step. In order to do so, we use the simplified
free energy derived in \citep[][Prop.\,3]{Lucke2018}, which takes for our GMM (\ref{EqnGMMIso}) the following form:
\begin{align}
\FF(\KK,\Theta) &= \disT \frac{1}{N}\sum_{n=1}^{N}\log\!\Big(\!\sum_{c\in\KKn} p(c,\yVecN\,|\,\Theta)\Big) \nonumber\\
 &=  \disT \frac{1}{N}\sum_{n=1}^{N}\log\!\Big(\!\sum_{c\in\KKn} \frac{1}{C}\,\Ncal\big(\yVecN;\,\muVec_c,\sigma^2\One\big)\Big). 
\label{EqnTruncatedF}
\end{align}

The truncated variational E-step (TV-E-step) first optimizes $\FF(\KK,\Theta)$ w.r.t.\ $\KK$ and the obtained truncated responsibilities $\qcn$ are then used in the M-step (\ref{EqnMStep}) to optimize $\FF(\KK,\Theta)$ w.r.t.\ $\Theta$.
The form of the free energy (\ref{EqnTruncatedF}) and the result that it is monotonically increased by iterating TV-E-step and M-step are the crucial theoretical results by \citet[][]{Lucke2018} that are used in this study.
Neither of these two results is straight-forward:
(A)~truncated distributions themselves depend on the model parameters $\Theta$, and 
(B)~it requires a number of derivations exploiting specific properties of truncated distributions to obtain the \changedOld{concise form used for}  
expression (\ref{EqnTruncatedF}). 

The TV-E-step now 
requires finding sets $\KKn$ which increase $\FF(\KK,\Theta)$. The free energy (\ref{EqnTruncatedF})
is computationally tractable, so a new $\KK$ could in principle be found by directly comparing $\FF(\KKNew,\Theta)$ of a new $\KKNew$ with $\FF(\KKOld,\Theta)$ of the current $\KKOld$.
We can slightly reformulate the problem by considering a specific data point $n$ and cluster $\ct\in\KKn$ for which we ask when any other replacing cluster $c\not\in\KKn$ would increase the free energy $\FF(\KK,\Theta)$.
By virtue of the properties of GMM (\ref{EqnGMMIso}) and due to the specific structure of the free energy (summation and concavity of the logarithm in Eqn.\,\ref{EqnTruncatedF}), we can then show:\\
\ \\
\noindent{}{\bf Proposition 1}\\
Consider the GMM (\ref{EqnGMMIso}) and the free energy (\ref{EqnTruncatedF}) for 
$n=1:N$ data points $\yVecN\in\RRR^D$. Furthermore, consider for a fixed $n$ the replacement of a cluster $\ct\in\KKn$ by
a cluster $c\not\in\KKn$. Then the free energy $\FF(\KK,\Theta)$ increases if and only if
\begin{align}
\|\yVecN\,-\,\muVec_c\| < \|\yVecN\,-\,\muVec_{\ct}\|\,. \label{EqnEuclid}
\end{align}
\ \\[-2mm]
{\bf Proof}\\
First observe that the free energy is increased if $p(c,\yVecN\,|\,\Theta)>p(\ct,\yVecN\,|\,\Theta)$ because of the summation over $c$ in
(\ref{EqnTruncatedF}) and because of the concavity of the logarithm. Analogously, the free energy stays constant or decreases for $p(c,\yVecN\,|\,\Theta)\leq{}p(\ct,\yVecN\,|\,\Theta)$. If we use the GMM (\ref{EqnGMMIso}), we obtain for the joint:\vspace{-1mm}
\begin{align}
\disT p(c,\yVec\,|\,\Theta)=\frac{1}{C}(2\pi\sigma^2)^{-\frac{D}{2}}\exp\!\big(-\frac{1}{2\sigma^2}\|{}\yVec-\muVec_c\|{}^2\big).
\end{align}
The first two factors are independent of the data point and cluster. The criterion for an increase of the free energy can therefore be reformulated
as follows:\vspace{-2mm}
\begin{align*}
&&p(c,\yVec\,|\,\Theta) &> p(\ct,\yVec\,|\,\Theta) \\
\Leftrightarrow{} && \disT \exp\!\big(-\frac{1}{2\sigma^2}\|\yVec-\muVec_c\|^2\big)& \disT> \exp\!\big(-\frac{1}{2\sigma^2}\|\yVec-\muVec_{\ct}\|^2\big) \\
%
%
\Leftrightarrow{}&& \disT \|\yVec-\muVec_c\|&<\|\yVec-\muVec_{\ct}\|\,. \nonumber\vspace{-4mm}
\end{align*}
\BOX\\[2mm]
Prop.\,1 means that we have to replace clusters in $\KKn$ that are relatively distant from $\yVecN$ by those closer to $\yVecN$
in order to increases the free energy $\FF(\KK,\Theta)$. Any such procedure gives with M-step (\ref{EqnMStep}) rise to a variational EM algorithm
that monotonically increases the lower bound (\ref{EqnTruncatedF}) of likelihood (\ref{EqnGMMLikelihood}). 
%
For an arbitrary generative model, the degree how much $\FF(\KK,\Theta)$ is increased or how long one should seek new clusters in the E-step is a design choice of the
algorithm.
%
In the case of GMMs (and other mixture models) we can exhaustively enumerate all clusters such that $\FF(\KK,\Theta)$ can be fully maximized.\\[3mm]
%
%
\noindent{}{\bf Corollary 1}\\
Same prerequisites as for Prop.\,1. The free energy $\FF(\KK,\Theta)$ is maximized w.r.t.\ $\KK$ (with fixed $\Theta$) if and only if
for all $n$ the set $\KKn$ contains the $C'$ clusters closest to data point $\yVecN$.\\[2mm]
%
%
\noindent{}{\bf Proof}\\
We assume that there are no equal distances among all pairs of data points and cluster centers.
If $\KKn$ contains the $C'$ closest clusters, it applies: $\forall{}c\in\KKn,\ \forall{}\ct\not\in\KKn:
\|{}\yVecN\,-\,\muVec_c\|{} < \|{}\yVecN\,-\,\muVec_{\ct}\|{}$. If we now consider an arbitrary $n$ and replace an arbitrary $c\in\KKn$ by an arbitrary $\cNew\not\in\KKn$ it applies
$\|{}\yVecN\,-\,\muVec_{\cNew}\|{} > \|{}\yVecN\,-\,\muVec_c\|{}$ such that by virtue of Prop.\,1 $\FF(\KK,\Theta)$ decreases. As any arbitrary such replacement (any change of $\KK$) results in a decrease of the free energy, $\FF(\KK,\Theta)$ is maximized if $\KK$ contains the $C'$ closest clusters.\\
\BOX\\[3mm]
We can now formulate a truncated variational EM (TV-EM) algorithm for GMM (\ref{EqnGMMIso}), here referred to as $k$-means-$C'$ (Alg.\,\ref{AlgTVEMforGMM}).\vspace{0mm}

%
%
%
%
\vspace{-4pt}
\begin{algorithm}[h]\vspace{0.5mm}
set $|\KKn|=C'$ for all $n$ and init $\muVec_{1:C}$ and $\sigma^2$;\algBreak
%
\Repeat{$\muVec_{1:C}$ and $\sigma^2$ have converged\vspace{1.5mm}}
{
%
\For{$n=1,\ldots,N$}{
define $\KKn$ such that $\forall{}c\in\KKn\ \forall{}\ct\not\in\KKn$:\\
$\|{}\yVecN\,-\,\muVec_c\|{}  <  \|{}\yVecN\,-\,\muVec_{\ct}\|{}$;
}
%
compute $\qcn$ for all $c$ and $n$ using (\ref{EqnQCN});\vspace{1.0mm}\\
update $\muVec_{1:C}$ and $\sigma^2$ using (\ref{EqnMStep});\vspace{0.5mm}
}
%
\caption{The $k$-means-$C'$ algorithm.
\label{AlgTVEMforGMM}}
\end{algorithm}
%
%
%
%
\ \\[-15mm]
\section{$k$-means and truncated variational EM for GMMs}
\label{SecKMeans}
TV-EM for GMMs (Alg.\,\ref{AlgTVEMforGMM}) increases the similarity between $k$-means and standard EM for GMMs in two ways:
(A)~it relates Euclidean distances to a variational free energy and thus to the GMM likelihood; and (B)~it introduces `hard' zeros in the updates of model parameters (some or many $\qcn$ are zero). 
Crucial remaining differences are, however, (A)~the weighted updates of the cluster centers in Eqn.\,\ref{EqnMStep} compared to the $k$-means update, and
(B)~the update of the cluster variance $\sigma^2$ in Eqn.\,\ref{EqnMStep} along with the cluster centers for Alg.\,\ref{AlgTVEMforGMM} which does not have a correspondence in $k$-means.
By considering the first difference, the obvious next step is to consider a boundary case of Alg.\,\ref{AlgTVEMforGMM}
by demanding that the sets $\KKn$ shall contain just one element, i.e., we set $C'=1$.
All derivations above apply for all $1\leq{}C'\leq{}C$, and while standard EM for the GMM (\ref{EqnGMMIso}) is recovered for $C'=C$, we find that for $C'=1$ standard $k$-means (Alg.\,1) is recovered.
%
%
\\[3mm]
\noindent{}{\bf Proposition 2}\\
Consider the TV-EM algorithm (Alg.\,\ref{AlgTVEMforGMM}) for the GMM (\ref{EqnGMMIso}) with $\sigma^2>0$. If we set
$C'=1$, then the TV-EM updates of the cluster centers $\muVec_c$ (\ref{EqnMStep}) become independent of the variance $\sigma^2$ and are given by the standard $k$-means algorithm in Alg.\,\ref{AlgKMeans}.\\[3mm]
{\bf Proof}\\
If we choose $|\KKn|=C'=1$ for all $n$, then each $\KKn$ computed in the TV-E-step of Alg.\,\ref{AlgTVEMforGMM} contains according to
Corollary\,1 the index of the cluster center closest to $\yVecN$ as only element. If we denote these centers by $c_o^{(n)}$, we get $\KKn=\{c_o^{(n)}\}$
and obtain for the truncated responsibilities $\qcn$ in (\ref{EqnQCN}):
\begin{align}
\qcn & = \frac{\disT \exp\!\big(\!-\!\frac{1}{2\sigma^2}\|{}\yVec-\muVec_c\|{}^2\big)\,\delta(c=c_o^{(n)})} {\disT\sum_{c'\in\{c_o^{(n)}\}}\exp\!\big(\!-\!\frac{1}{2\sigma^2}\|{}\yVec-\muVec_{c'}\|{}^2\big)}
 = \left\{\begin{array}{@{\hspace{2pt}}l@{\hspace{7pt}}l@{\hspace{-2pt}}}
1 & \mbox{if\ \ } c=c_o^{(n)}\\
0 & \mbox{otherwise}
\end{array}\right.,
\end{align}
which is identical to $\scn$ in Alg.\,1. By using $\qcn=\scn$ for the M-step, we consequently obtain:
\begin{align}
\phantom{i}\backOne\muVec^{\mathrm{\,new}}_c \!= \frac{\disT\sum_{n=1}^N{}\scn\yVecN}{\disT\sum_{n=1}^N{}\scn}
%
, \,\,\, \sigma_{\mathrm{new}}^2 = \frac{1}{DN}\!\sum_{n,c=1}^{N,C}\!\scn\,\|\yVecN-\muVec^{\mathrm{\,new}}_c\|^2.\ \label{EqnMStepKMeans} 
\end{align}
Now observe that the computation of $\qcn=\scn$ and the updates of the $\muVec_c$ do not involve
the parameter $\sigma^2$. The cluster centers $\muVec_c$ can thus be optimized without requiring
knowledge about the cluster variances $\sigma^2$, i.e., the $\muVec_c$ optimization becomes independent of $\sigma^2$.
As the TV-EM updates for $\qcn$ and $\muVec_c$ are identical to the updates of $\scn$ and $\muVec_c$ in Alg.\,\ref{AlgKMeans},
the optimization procedure for the  $\muVec_c$ is given by the standard $k$-means algorithm.\\
\BOX\\[2mm]
A direct consequence of Prop.\,2 is that standard $k$-means provably monotonically increases the truncated free energy (\ref{EqnTruncatedF}) with $C'=1$. Notably, only for this choice of $C'$ the updates of cluster means and variance decouple. We can, of course, add the variance updates to
standard $k$-means but this does not effect the $\muVec_c$ updates. With or without $\sigma^2$ updates the free energy monotonically increases. If our goal
is the maximization of the free energy objective, the $\sigma^2$ updates should be included, however. According to the independence of $\muVec_c$-optimization from
$\sigma^2$, it would be sufficient to update $\sigma^2$ once and only after $k$-means has optimized the cluster centers.

Prop.\,2 shows that $k$-means is obtained from a variational free energy objective.
This free energy is in turn closely related to the likelihood objective of
GMMs (\ref{EqnGMMLikelihood}).
By analyzing the free energy for $C'=1$ more closely, we can make this relation more explicit.\\[3mm]
%
%
\noindent{}{\bf Proposition 3}\\
%
%
Consider a set of $N$ data points $\yVec^{\,(1:N)}\in\RRR^D$ and the $k$-means algorithm (Alg.\,\ref{AlgKMeans}) where $s^{(1:N)}_{1:C}$ and $\muVec_{1:C}$ denote, respectively, the cluster assignments and cluster centers computed in one iteration.
Furthermore, let $\sigma^2$ denote the variance computed with $s^{(1:N)}_{1:C}$ and $\muVec_{1:C}$ as in Eqn.\,\ref{EqnMStep}:
\begin{align}
\sigma^2 = \sigma^2(s^{(1:N)}_{1:C},\muVec_{1:C}) = \frac{1}{DN}\sum_{n=1}^{N}\sum_{c=1}^{C}\scn\, \|{}\yVecN\,-\,\muVec_c\|{}^2\,.\label{EqnSigmaKMeans}
\end{align}
It then follows that each $k$-means iteration monotonically increases the free energy $\FF(s^{(1:N)}_{1:C},\muVec_{1:C})$ given by:
\begin{align}
 \FF(s^{(1:N)}_{1:C},\muVec_{1:C}) &= -\log(C)\,-\,\frac{D}{2}\log(2\pi{}e\sigma^2)\,,
\label{EqnFreeEnergyKMeans}
\end{align}
where $e$ is Euler's number. The free energy (\ref{EqnFreeEnergyKMeans}) is a lower bound of the GMM log-likelihood (\ref{EqnGMMLikelihood}).
The difference between log-likelihood (\ref{EqnGMMLikelihood}) and free energy (\ref{EqnFreeEnergyKMeans}) is given by:
\begin{align}
\hspace{-2.0mm}D_{KL}(s^{(1:N)}_{1:C},\muVec_{1:C})\!=\!\frac{D}{2}\!+\!\frac{1}{N}\!\sum_{n=1}^{N} \hspace{-0.5mm}\log\Big(\hspace{-0.4mm} \sum_{c=1}^{C} \exp\!\big(\!-\!\frac{\|\yVecN-\muVec_c\|^2}{ 2\sigma^2 } \big) \Big).
\label{EqnDKLKMeans}
\end{align}
If for all $n$ and $c$ where $\scn=0$ applies: $\sigma\ll{}\|\yVecN-\muVec_c\|$, i.e., if clusters are well separable, then the bound becomes tight.\\[3mm]
\noindent{}{\bf Proof}\\
%
In the $k$-means case ($|\KKn|=C'=1$) each $\KKn$ only contains
one cluster which is given by the cluster assignments $\scn$ as: $\KKn=\{c\,|\, \scn=1\}$. If we abbreviate this cluster for $n$ with $c_o^{(n)}$,
it follows for the free energy (\ref{EqnTruncatedF}) after one $k$-means iteration:\vspace{-3pt}
\begin{align}
&\FF(\KK,\Theta) = \disT \frac{1}{N}\sum_{n}\log\!\big(\sum_{c\in\{c_o^{(n)}\}}\frac{1}{C}\,\Ncal(\yVecN;\,\muVec_{c},\sigma^2\One)\big) \nonumber\\[-1pt]
&= \disT \frac{1}{N}\sum_{n}\,\log\!\big(\frac{1}{C}\,\Ncal(\yVecN;\,\muVec_{c_o^{(n)}},\sigma^2\One)\big) \label{EqnProofA} \\[-2pt]
&= \disT -\log(C) - \frac{D}{2} \log(2\pi\sigma^2) \disT - \frac{1}{2\sigma^2} \frac{1}{N}\sum_{n=1}^{N}\sum_{c=1}^{C} \scn\|{}\yVecN\,-\,\muVec_{c}\|{}^2\,, \nonumber
\end{align}
\vspace{-14pt}\\
where we inserted the Gaussian density and then used $f(c_o^{(n)})=\sum_{c}\scn{}f(c)$. $\sigma^2$ and $\muVec_{c}$ are the parameters obtained after a single $k$-means iteration.
Following (\ref{EqnMStep}) we can therefore insert the expression $\frac{1}{DN}\sum_{n=1}^N{}\scn\,\|\yVecN-\muVec_c\|^2$ for $\sigma^2$, noting that the $\muVec_c$ are the same as in (\ref{EqnProofA}).
The last term of (\ref{EqnProofA}) then simplifies to $-\frac{D}{2}$.
If we now rewrite this as $-\frac{D}{2}\log(e)$ and combine with the second summand, we obtain (\ref{EqnFreeEnergyKMeans}).
%


The difference (\ref{EqnDKLKMeans}) between log-likelihood and free energy can be derived from the KL-divergence $\DKL{\qCN}{\rCN}$.
Using results of \citep[][]{Lucke2018} the KL-divergence for a truncated distribution is given by: $\DKL{\qCN}{\rCN}=-\sum_{n}\log(\sum_{c\in\KKn}\rcn)$.
Inserting $\rcn$ (Alg.\,\ref{EqnGMMAlgo}) for the GMM (\ref{EqnGMMIso}), we obtain:
\vspace{-4pt}
\begin{align*}
& \DKL{\qCN}{\rCN} = \disT -\frac{1}{N}\sum_{n}\log\!\Big(\sum_{c\in\KKn} \frac{\exp\!\big(-\frac{1}{2\sigma^2}\|\yVecN-\muVec_c\|^2\big)} {\sum_{c'} \exp\!\big(-\frac{1}{2\sigma^2}\|\yVecN-\muVec_{c'}\|^2\big)}\Big)\\[-4pt]
%
%
&= \disT \frac{1}{2N\sigma^2}\!\sum_{n,c}\!\scn\|\yVecN\!-\muVec_{c}\|^2 \disT + \frac{1}{N}\!\sum_{n}\! \log\!\Big(\! \sum_{c} \exp\!\big(\frac{-1}{2\sigma^2}\|\yVecN\!-\muVec_{c}\|^2\big)\Big)\\[-1pt]
&= \disT \frac{D}{2} + \frac{1}{N}\sum_{n} \log\!\Big( \sum_{c} \exp\!\big(-\frac{\|\yVecN-\muVec_{c}\|^2}{2\sigma^2} \big) \Big)\,,\vspace{1.5mm} \numberthis
\label{EqnDKL}
\end{align*}\vspace{-11pt}\\
using again expression (\ref{EqnSigmaKMeans}) for $\sigma^2$. 
If $\sigma^2\ll\|\yVecN-\muVec_c\|^2$ for all $n$,$c$ with $\scn=0$, then the last term of (\ref{EqnDKL}) is dominated by those $n$,$c$ with $\scn=1$, such that $\DKL{\qCN}{\rCN}\!\rightarrow\!0$.\\
\BOX\\[2mm]
%
\changedOld{Prop.\,3 makes explicit the difference to the GMM likelihood objective if $k$-means is used for parameter optimization (we elaborate in Suppl.\,\ref{sec:Difference}).}
\changedOld{Furthermore,} by using Prop.\,3, we can now directly link the GMM likelihood to the $k$-means objective.\\[3mm]
%
\noindent{}{\bf Corollary 2}\\
%
%
%
If $s^{(1:N)}_{1:C}$ and $\muVec_{1:C}$ are updated by $k$-means (Alg.\,1), then it applies for the GMM likelihood (\ref{EqnGMMLikelihood}) after each iteration that
\begin{equation}
\LL(\Theta) \geq{} -\log(C)\,-\,\frac{D}{2}\log\!\Big(\,\frac{2\pi{}e}{DN}\,\JJ(s^{(1:N)}_{1:C},\muVec_{1:C})\Big)\,,\label{EqnResult}
\end{equation}
where $\JJ(s^{(1:N)}_{1:C},\muVec_{1:C})$ is the $k$-means objective (\ref{EqnKMObjective}).
The lower free energy bound (right-hand-side of Eqn.\,\ref{EqnResult}) is strictly monotonically increased.\\[2mm]
\noindent{}{\bf Proof}\\
If $s^{(1:N)}_{1:C}$ are the cluster assignments of the first for-loop in Alg.\,1, and $\muVec_{1:C}$ the centers of the second for-loop, then
$\sigma^2$ in Prop.\,3 can directly be replaced by $(DN)^{-1}\JJ(s^{(1:N)}_{1:C},\muVec_{1:C})$. The free energy is thus a function of 
the $k$-means objective. As $k$-means has been shown to strictly monotonically decrease \changedOld{the objective $\JJ(s^{(1:N)}_{1:C},\muVec_{1:C})$ \citep[compare, e.g.,][]{Anderberg1973,InabaEtAl2000},
the lower free energy bound (\ref{EqnResult})} is strictly monotonically increased by $k$-means.
\\
\BOX

\vspace{-2mm}
\section{Applications of Theoretical Results}
\label{SecApp}
The principled link between $k$-means and variational GMMs can be used for a number of theoretical applications and interpretations of previous algorithms, including soft-$k$-means, lazy-$k$-means, fuzzy $k$-means, and previous GMM variants with `hard' posterior zeros. For such comparisons, let us first generalize Prop.\,3 for $k$-means-$C'$ with $C'>1$.\\[-2mm]
%
\ \\
\noindent{}{\bf Proposition 4}\\
%
%
Same prerequisites as for Prop.\,3. If $\muVec_{1:C}$ and $\sigma^2$ are updated using $k$-means-$C'$ (Alg.\,3),
then a lower free energy bound of the log-likelihood (\ref{EqnGMMLikelihood}) is monotonically increased. The bound
is after convergence given by:
%
%
\begin{align}
\FF(q^{(1:N)}_{1:C},\muVec_{1:C}) =&  -\log(C) - \frac{D}{2}\log(2\pi{}e\sigma^2) \nonumber \\[-4pt]
& - \frac{1}{N}\sum_{n=1}^{N}\sum_{c=1}^{C}\qcn{}\log(\qcn{})\,.\label{EqnFreeEnergyKMeansC}\vspace{-10mm}
\end{align}
%
%
\noindent{}{\bf Proof}\\
For GMM (\ref{EqnGMMIso}) the entropy of the noise distribution, $\HH(p(\yVec\,|\,c,\Theta))=\HH(\Ncal(\yVec;\muVec_c,\sigma^2\One))$,
does not change with $c$. The GMM therefore has an entropy limit \citep{LuckeHenniges2012} given by:\vspace{-5mm}
\begin{align*}
\phantom{wwwwwwww}\overline{Q}(\Theta) &= -\HH(p(c\,|\,\Theta))-\HH(p(\yVec\,|\,c,\Theta))\\
                     &= -\log(C)-\disT\frac{D}{2}\log(2\pi{}e\sigma^2)\,,\nonumber
\end{align*}
which is derived simply by inserting (\ref{EqnGMMIso}) into $\overline{Q}(\Theta)$. If we \citep[following][]{LuckeHenniges2012} reformulate the free energy (\ref{EqnTruncatedF}) such that it is expressed in terms of this entropy limit, we obtain:
$
\begin{array}{c}
\hspace{-1mm}\FF(\KK,\Theta)= \overline{Q}(\Theta) + \frac{D}{2} \big(1\,- \frac{\sigma^2_{\mathrm{new}}}{\sigma^2} \big) + \frac{1}{N}\sum_{n} \HH(q^{(n)}_c),\hspace{-1mm}
\end{array}
$
where $\sigma^2_{\mathrm{new}}$ is the variance after the M-step \changedOld{of $k$-means-$C'$}. 
At convergence, the ratio $\sigma^2_{\mathrm{new}}/\sigma^2$ converges to one and we obtain (\ref{EqnFreeEnergyKMeansC}).\\
%
\BOX\\[2mm]
%
%
Already by considering (\ref{EqnTruncatedF}), we can conclude that for the same $\Theta$ applies $\FF(\tilde{\KK},\Theta)\leq{}\FF(\KK,\Theta)$ if $\tilde{\KK}\subseteq\KK$.
Prop.\,4 now shows that the free energy difference is (after convergence) solely given by the entropy of the truncated distributions.
For $C'=1$ the entropy is zero, for $C'=C$ the entropy is
maximal and (\ref{EqnFreeEnergyKMeansC}) can be used to \changedOld{estimate} the likelihood during learning.

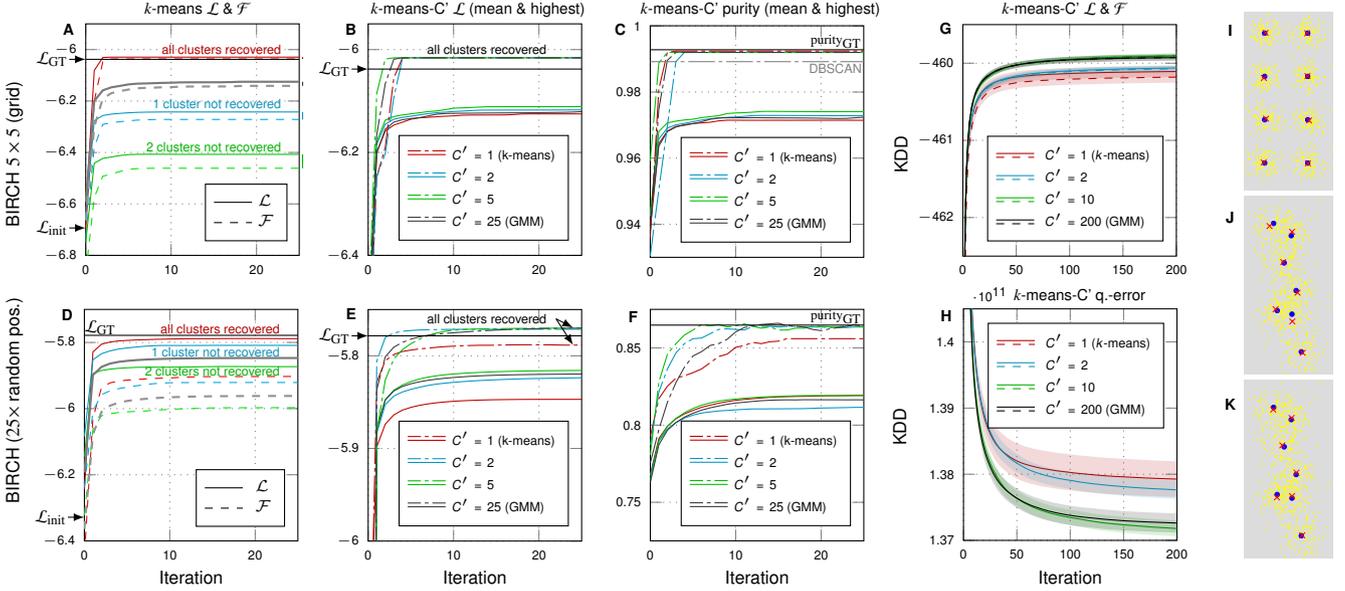
\begin{figure*}[bt]
\vspace{15pt}
	\begin{picture}(0,0)
		\put(2,-80){\rotatebox{90}{\rlap{\makebox[5.80cm]{\fontsize{7}{0}\selectfont\sffamily BIRCH~$5\times 5$ (grid)}}}}
	\end{picture}
	\begin{picture}(0,0)
		\put(0,55){\rlap{\makebox[5cm]{\fontsize{6}{0}\selectfont\sffamily  $k$-means $\mathcal{L}$ \& $\mathcal{F}$}}}
	\end{picture}
  \begin{subfigure}[c]{0.193\textwidth}
    \begin{adjustbox}{trim=0pt 10pt 0pt 14pt}
\tikzsetnextfilename{BIRCH-1-5-100_k-means_25}
\pgfplotsset{
	grid style={dotted,gray},
	minor grid style={dotted,lightgray},
  tick label style = {font=\tiny\sansmath\sffamily},
  legend style = {font=\sansmath\sffamily},
  xlabel style = {font=\sansmath\sffamily},
  ylabel style = {font=\sansmath\sffamily},
  legend image code/.code={
    \draw[mark repeat=2,mark phase=2]
    plot coordinates {
      (0cm,0cm)
      (0.25cm,0cm)        
      (0.5cm,0cm)         
    };%
  },
}	
\begin{tikzpicture}
	\tikzset{mark size={1.0}}
	\begin{axis}[
		colormap access=direct,
		width = 1.24\linewidth,
		height = 0.19 * \textheight,
		xmin=0,
		xmax=25,
		scaled x ticks = false,
		xlabel near ticks, xticklabel pos=lower,
    x label style={at={(0.5,-0.09)}},
		ymin = -6.8,
		ymax = -5.9,
		ylabel near ticks, yticklabel pos=left,
		grid = both,
		%
    legend entries={
      \fontsize{6}{0}\selectfont\sffamily $\mathcal{L}$,
      \fontsize{6}{0}\selectfont\sffamily $\mathcal{F}$
    },
		legend style={
			at={(0.945,0.31)},
			legend columns=1,
			row sep=-2pt,
		},
		legend cell align=left,
    after end axis/.code={
          \draw [-latex, shorten <=-3pt] (axis cs: -1,-6.6933935576) node[left]{\fontsize{6}{0}\selectfont\sffamily $\mathcal{L}_{\textnormal{init}}$} to[out=360,in=180] (axis cs: 0,-6.6933935576);
          \draw [solid] (axis cs: 0,-6.0378687151) to (axis cs: 25,-6.0378687151);
          \draw [-latex, shorten <=-3pt] (axis cs: -1,-6.0378687151) node[left]{\fontsize{6}{0}\selectfont\sffamily $\mathcal{L}_{\textnormal{GT}}$} to[out=360,in=180] (axis cs: 0,-6.0378687151);
          \node at (axis cs: -2,-5.92) {\fontsize{6}{0}\selectfont\sffamily\bfseries A};
					\draw [-, red!70!black] (axis cs: 25.4,-6.03638830146) to (axis cs: 25.4,-6.02933957757);
					\draw [-, cyan!70!black] (axis cs: 25.4,-6.27162484389) to (axis cs: 25.4,-6.24257018759);
					\draw [-, green!70!black] (axis cs: 25.4,-6.46130326354) to (axis cs: 25.4,-6.40734896632);
					\draw [-, grey!70!black] (axis cs: 25.4,-6.14110728288) to (axis cs: 25.4,-6.12502651431);
    }
  	]
	\addlegendimage{solid}
	\addlegendimage{dashed}
	\addplot+ [thin, solid, mark=None, color=red!75!black] table[x=n, y=loglikelihood] {./figs/BIRCH-1-5-100_k-means_25-2.txt};
	\addplot+ [thin, dashed, mark=None, color=red!90!black] table[x=n, y=free-energy] {./figs/BIRCH-1-5-100_k-means_25-2.txt};
	\node[red!70!black, anchor=east] at (axis cs: 24,-6.0) {\fontsize{5}{0}\selectfont\sffamily all clusters recovered};
	\addplot+ [thin, solid, mark=None, color=cyan!75!black] table[x=n, y=loglikelihood] {./figs/BIRCH-1-5-100_k-means_25-10.txt};
	\addplot+ [thin, dashed, mark=None, color=cyan!90!black] table[x=n, y=free-energy] {./figs/BIRCH-1-5-100_k-means_25-10.txt};	
	\node[cyan!70!black, anchor=east] at (axis cs: 24,-6.21) {\fontsize{5}{0}\selectfont\sffamily 1 cluster not recovered};
	\addplot+ [thin, solid, mark=None, color=green!75!black] table[x=n, y=loglikelihood] {./figs/BIRCH-1-5-100_k-means_25-1.txt};
	\addplot+ [thin, dashed, mark=None, color=green!90!black] table[x=n, y=free-energy] 
{./figs/BIRCH-1-5-100_k-means_25-1.txt};	
  \node[green!70!black, anchor=east] at (axis cs: 24,-6.38) {\fontsize{5}{0}\selectfont\sffamily 2 clusters not recovered};
	\addplot+ [thick, solid, mark=None, color=grey!75!white] table[x=n, y=loglikelihood] {./figs/BIRCH-1-5-100_k-means_25.txt};
	\addplot+ [thick, dashed, mark=None, color=grey!60!white] table[x=n, y=free-energy] {./figs/BIRCH-1-5-100_k-means_25.txt};	
  \end{axis}
\end{tikzpicture}
    \end{adjustbox}
  \end{subfigure}
	\begin{picture}(0,0)
		\put(0,55){\rlap{\makebox[5cm]{\fontsize{6}{0}\selectfont\sffamily  $k$-means-C' $\mathcal{L}$ (mean \& highest) }}}
	\end{picture}
  \begin{subfigure}[c]{0.193\textwidth}
    \begin{adjustbox}{trim=0pt 10pt 0pt 14pt}
\tikzsetnextfilename{BIRCH-1-5-100_k-means-Cprime}
\pgfplotsset{
	grid style={dotted,gray},
	minor grid style={dotted,lightgray},
  tick label style = {font=\tiny\sansmath\sffamily},
  legend style = {font=\sansmath\sffamily},
  xlabel style = {font=\sansmath\sffamily},
  ylabel style = {font=\sansmath\sffamily},
  legend image code/.code={
    \draw[mark repeat=2,mark phase=2]
    plot coordinates {
      (0cm,0cm)
      (0.25cm,0cm)        
      (0.5cm,0cm)         
    };%
  }
}	
\begin{tikzpicture}
	\tikzset{mark size={1.0}}
	\begin{axis}[
		colormap access=direct,
		width = 1.24\linewidth,
		height = 0.19 * \textheight,
		xmin=0,
		xmax=25,
		scaled x ticks = false,
		xlabel near ticks, xticklabel pos=lower,
    x label style={at={(0.5,-0.09)}},
		ymin = -6.4,
		ymax = -5.95,
		ylabel near ticks, yticklabel pos=left,
		grid = both,
   legend image code/.code={%
     \draw[dash pattern=on 0.175cm off 0.05cm on 0.05cm off 0.05cm on 0.175cm] (0cm,0.05cm) -- (0.5cm,0.05cm);
     \draw[solid]  (0cm, 0.0cm) -- (0.5cm, 0.0cm);
    },		
    legend entries={
      \fontsize{5}{0}\selectfont\sffamily $C' = 1$ (k-means),
      \fontsize{5}{0}\selectfont\sffamily  $C' = 2$,
      \fontsize{5}{0}\selectfont\sffamily  $C' = 5$,
      \fontsize{5}{0}\selectfont\sffamily $C' = 25$ (GMM),
    },
		legend style={
			at={(0.94,0.52)},
			legend columns=1,
			row sep=-2pt,
		},
		legend cell align=left,
    after end axis/.code={
          \draw [solid] (axis cs: 0,-6.0378687151) to (axis cs: 25,-6.0378687151);
          \draw [-latex, shorten <=-3pt] (axis cs: -1,-6.0378687151) node[left]{\fontsize{6}{0}\selectfont\sffamily $\mathcal{L}_{\textnormal{GT}}$} to[out=360,in=180] (axis cs: 0,-6.0378687151);
          \node at (axis cs: -2,-5.960) {\fontsize{6}{0}\selectfont\sffamily\bfseries B};
    }
  	]
	\addplot+ [thin, solid, mark=None, color=red!75!black] table[x=n, y=loglikelihood] {./figs/BIRCH-1-5-100_k-means_25.txt};
	\addplot+ [thin, solid, mark=None, color=cyan!75!black] table[x=n, y=L] {./figs/data2019/BIRCH_2_25_results.txt};
	\addplot+ [thin, solid, mark=None, color=green!75!black] table[x=n, y=L] {./figs/data2019/BIRCH_5_25_results.txt};
	\addplot+ [thin, solid, mark=None, color=grey!75!black] table[x=n, y=L] {./figs/data2019/BIRCH_25_25_results.txt};
	\addplot+ [thin, dash pattern=on 0.35cm off 0.05cm on 0.05cm off 0.05cm, mark=None, color=red!75!black] table[x=n, y=L] {./figs/data2019/BIRCH_1_25_best_LL.txt};  
	\addplot+ [thin, dash pattern=on 0.35cm off 0.05cm on 0.05cm off 0.05cm, mark=None, color=cyan!75!black] table[x=n, y=L] {./figs/data2019/BIRCH_2_25_best_LL.txt};  
	\addplot+ [thin, dash pattern=on 0.35cm off 0.05cm on 0.05cm off 0.05cm, mark=None, color=green!75!black] table[x=n, y=L] {./figs/data2019/BIRCH_5_25_best_LL.txt};  
	\addplot+ [thin, dash pattern=on 0.35cm off 0.05cm on 0.05cm off 0.05cm, mark=None, color=grey!75!black] table[x=n, y=L] {./figs/data2019/BIRCH_25_25_best_LL.txt};
  \node[black, anchor=east] at (axis cs: 22,-6.0) {\fontsize{5}{0}\selectfont\sffamily all clusters recovered};
  \end{axis}
\end{tikzpicture}
    \end{adjustbox}
  \end{subfigure}
	\begin{picture}(0,0)
		\put(0,55){\rlap{\makebox[5cm]{\fontsize{6}{0}\selectfont\sffamily  $k$-means-C' purity (mean \& highest) }}}
	\end{picture}
  \begin{subfigure}[c]{0.193\textwidth}
    \begin{adjustbox}{trim=-5pt 10pt 0pt 14pt}
\tikzsetnextfilename{BIRCH_k-means-Cprime_purity_main}
\pgfplotsset{
	grid style={dotted,gray},
	minor grid style={dotted,lightgray},
  tick label style = {font=\tiny\sansmath\sffamily},
  legend style = {font=\sansmath\sffamily},
  xlabel style = {font=\sansmath\sffamily},
  ylabel style = {font=\sansmath\sffamily},
  legend image code/.code={
    \draw[mark repeat=2,mark phase=2]
    plot coordinates {
      (0cm,0cm)
      (0.25cm,0cm)        
      (0.5cm,0cm)         
    };%
  }
}	
\begin{tikzpicture}
	\tikzset{mark size={1.0}}
	\begin{axis}[
		colormap access=direct,
		width = 1.24\linewidth,
		height = 0.19 * \textheight,
		xmin=0,
		xmax=25,
		xtick = {0,10,...,50},
		scaled x ticks = false,
		xlabel near ticks, xticklabel pos=lower,
    x label style={at={(0.5,-0.09)}},
		ymin = 0.93,
		ymax = 1.00005,
		ylabel near ticks, yticklabel pos=left,
		grid = both,
   legend image code/.code={%
     \draw[dash pattern=on 0.175cm off 0.05cm on 0.05cm off 0.05cm on 0.175cm] (0cm,0.05cm) -- (0.5cm,0.05cm);
     \draw[solid]  (0cm, 0.0cm) -- (0.5cm, 0.0cm);
    },		
    legend entries={
      \fontsize{5}{0}\selectfont\sffamily $C' = 1$ (k-means),
      \fontsize{5}{0}\selectfont\sffamily  $C' = 2$,
      \fontsize{5}{0}\selectfont\sffamily  $C' = 5$,
      \fontsize{5}{0}\selectfont\sffamily $C' = 25$ (GMM),
    },
		legend style={
			at={(0.94,0.52)},
			legend columns=1,
			row sep=-2pt,
		},
		legend cell align=left,
    after end axis/.code={
          \draw [solid] (axis cs: 0,0.9928) to (axis cs: 25,0.9928);
          \draw [] (axis cs:25,0.9928) node[inner sep=0.5pt, anchor=south east]{\fontsize{5}{0}\selectfont\sffamily $\mathrm{purity}_{\textnormal{GT}}$};
          \draw [dash pattern=on 0.35cm off 0.05cm on 0.05cm off 0.05cm, gray] (axis cs: 0,0.9892) to (axis cs: 25,0.9892);
          \draw [] (axis cs:25,0.9892) node[inner sep=0.5pt, anchor=north east, gray]{\fontsize{5}{0}\selectfont\sffamily DBSCAN};
          \node at (axis cs: -3.5,0.999) {\fontsize{6}{0}\selectfont\sffamily\bfseries C};
    }
  	]

	\addplot+ [thin, solid, mark=None, color=red!75!black] table[x=n, y=purity] {./figs/data2019/BIRCH_1_25_results.txt};
	\addplot+ [thin, solid, mark=None, color=cyan!75!black] table[x=n, y=purity] {./figs/data2019/BIRCH_2_25_results.txt};
	\addplot+ [thin, solid, mark=None, color=green!75!black] table[x=n, y=purity] {./figs/data2019/BIRCH_5_25_results.txt};
	\addplot+ [thin, solid, mark=None, color=grey!75!black] table[x=n, y=purity] {./figs/data2019/BIRCH_25_25_results.txt};
	\addplot+ [thin, dash pattern=on 0.35cm off 0.05cm on 0.05cm off 0.05cm, mark=None, color=red!75!black] table[x=n, y=purity] {./figs/data2019/BIRCH_1_25_best_p_NMI.txt};  
	\addplot+ [thin, dash pattern=on 0.35cm off 0.05cm on 0.05cm off 0.05cm, mark=None, color=cyan!75!black] table[x=n, y=purity] {./figs/data2019/BIRCH_2_25_best_p_NMI.txt};  
	\addplot+ [thin, dash pattern=on 0.35cm off 0.05cm on 0.05cm off 0.05cm, mark=None, color=green!75!black] table[x=n, y=purity] {./figs/data2019/BIRCH_5_25_best_p_NMI.txt};  
	\addplot+ [thin, dash pattern=on 0.35cm off 0.05cm on 0.05cm off 0.05cm, mark=None, color=grey!75!black] table[x=n, y=purity] {./figs/data2019/BIRCH_25_25_best_p_NMI.txt};
  \end{axis}
\end{tikzpicture}
    \end{adjustbox}
  \end{subfigure}
	\begin{picture}(0,0)
		\put(15,-80){\rotatebox{90}{\rlap{\makebox[5.80cm]{\fontsize{7}{0}\selectfont\sffamily KDD}}}}
	\end{picture}	
	\begin{picture}(0,0)
		\put(5,55){\rlap{\makebox[5cm]{\fontsize{6}{0}\selectfont\sffamily  $k$-means-C' $\mathcal{L}$ \& $\mathcal{F}$}}}
	\end{picture}
  \begin{subfigure}[c]{0.193\textwidth}
    \begin{adjustbox}{trim=3pt 10pt 0pt 14pt}
\tikzsetnextfilename{KDD_k-means_Cprime}
\pgfplotsset{
	grid style={dotted,gray},
	minor grid style={dotted,lightgray},
  tick label style = {font=\tiny\sansmath\sffamily},
  legend style = {font=\sansmath\sffamily},
  xlabel style = {font=\sansmath\sffamily},
  ylabel style = {font=\sansmath\sffamily},
  legend image code/.code={
    \draw[mark repeat=2,mark phase=2]
    plot coordinates {
      (0cm,0cm)
      (0.25cm,0cm)        
      (0.5cm,0cm)         
    };%
  }
}

\begin{tikzpicture}
	\tikzset{mark size={1.0}}
	\begin{axis}[
		colormap access=direct,
		width = 1.24\linewidth,
		height = 0.19 * \textheight,
    xmin=0,
		xmax=200,
		xtick = {0,50,...,200},
		scaled x ticks = false,
		xlabel near ticks, xticklabel pos=lower,
    x label style={at={(0.5,-0.09)}},
		ymin=-462.5,
		ymax = -459.5,
		ylabel={\scriptsize\sffamily \phantom{Likelihood / Free Energy}},
		ylabel near ticks,
		yticklabel pos=left,
		grid = both,
   legend image code/.code={%
     \draw[dashed] (0cm,-0.025cm) -- (0.5cm,-0.025cm);
     \draw[solid]  (0cm, 0.025cm) -- (0.5cm, 0.025cm);
    },		
    legend entries={
      \fontsize{5}{0}\selectfont\sffamily \!\! $C'=1$ ($k$-means),
      \fontsize{5}{0}\selectfont\sffamily \!\! $C'=2$,
      \fontsize{5}{0}\selectfont\sffamily \!\! $C'=10$,
      \fontsize{5}{0}\selectfont\sffamily \!\! $C'=200$ (GMM)
    },
		legend style={
			at={(0.94,0.52)},
			legend columns=1,
			row sep=-2pt,
			column sep=2pt,
		},
		legend cell align=left,
    after end axis/.code={
      \node at (axis cs: -16,-459.55) {\fontsize{6}{0}\selectfont\sffamily\bfseries G};
    }
  ]
  \addlegendimage{color=red!70!black}  	
	\addlegendimage{color=cyan!70!black}
  \addlegendimage{color=green!70!black}  	
  \addlegendimage{color=black}
  %
  
  \newcommand\filename{./figs/KDD_k-means-C'_200_1.txt}
  \addplot [draw=none, stack plots=y, forget plot] table[
    x=n,
    y expr=\thisrow{F}-\thisrow{F_err}
  ] {\filename};	
  
  \addplot [draw=none, fill=red!70!black, stack plots=y, fill opacity=0.15] table [
      x=n,
      y expr=2*\thisrow{F_err}
  ] {\filename}\closedcycle;
  
  \addplot [forget plot, stack plots=y,draw=none] table [x=n, y expr=-(\thisrow{F}+\thisrow{F_err})] {\filename};
    
  \addplot [draw=none, stack plots=y, forget plot] table[
    x=n,
    y expr=\thisrow{L}-\thisrow{L_err}
  ] {\filename};	
  
  \addplot [draw=none, fill=red!70!black, stack plots=y, fill opacity=0.15] table [
      x=n,
      y expr=2*\thisrow{L_err}
  ] {\filename}\closedcycle;
  
  \addplot [forget plot, stack plots=y,draw=none] table [x=n, y expr=-(\thisrow{L}+\thisrow{L_err})] {\filename};

	\addplot [solid, mark=None, color=red!70!black] table[x=n, y=L] {\filename};	
	\addplot [dashed, mark=None, color=red!70!black] table[x=n, y=F] {\filename};

  \renewcommand\filename{./figs/KDD_k-means-C'_200_2.txt}
  \addplot [draw=none, stack plots=y, forget plot] table[
    x=n,
    y expr=\thisrow{F}-\thisrow{F_err}
  ] {\filename};	
  
  \addplot [draw=none, fill=cyan!70!black, stack plots=y, fill opacity=0.15] table [
      x=n,
      y expr=2*\thisrow{F_err}
  ] {\filename}\closedcycle;
  
  \addplot [forget plot, stack plots=y,draw=none] table [x=n, y expr=-(\thisrow{F}+\thisrow{F_err})] {\filename};
    
  \addplot [draw=none, stack plots=y, forget plot] table[
    x=n,
    y expr=\thisrow{L}-\thisrow{L_err}
  ] {\filename};	
  
  \addplot [draw=none, fill=cyan!70!black, stack plots=y, fill opacity=0.15] table [
      x=n,
      y expr=2*\thisrow{L_err}
  ] {\filename}\closedcycle;
  
  \addplot [forget plot, stack plots=y,draw=none] table [x=n, y expr=-(\thisrow{L}+\thisrow{L_err})] {\filename};

	\addplot [solid, mark=None, color=cyan!70!black] table[x=n, y=L] {\filename};	
	\addplot [dashed, mark=None, color=cyan!70!black] table[x=n, y=F] {\filename};

  \renewcommand\filename{./figs/KDD_k-means-C'_200_10.txt}
  \addplot [draw=none, stack plots=y, forget plot] table[
    x=n,
    y expr=\thisrow{F}-\thisrow{F_err}
  ] {\filename};	
  
  \addplot [draw=none, fill=green!70!black, stack plots=y, fill opacity=0.15] table [
      x=n,
      y expr=2*\thisrow{F_err}
  ] {\filename}\closedcycle;
  
  \addplot [forget plot, stack plots=y,draw=none] table [x=n, y expr=-(\thisrow{F}+\thisrow{F_err})] {\filename};
    
  \addplot [draw=none, stack plots=y, forget plot] table[
    x=n,
    y expr=\thisrow{L}-\thisrow{L_err}
  ] {\filename};	
  
  \addplot [draw=none, fill=green!70!black, stack plots=y, fill opacity=0.15] table [
      x=n,
      y expr=2*\thisrow{L_err}
  ] {\filename}\closedcycle;
  
  \addplot [forget plot, stack plots=y,draw=none] table [x=n, y expr=-(\thisrow{L}+\thisrow{L_err})] {\filename};

	\addplot [solid, mark=None, color=green!70!black] table[x=n, y=L] {\filename};	
	\addplot [dashed, mark=None, color=green!70!black] table[x=n, y=F] {\filename};

%
%
%
%
%
%

  \renewcommand\filename{./figs/KDD_GMM.txt}
  \addplot [draw=none, stack plots=y, forget plot] table[
    x=n,
    y expr=\thisrow{F}-\thisrow{F_err}
  ] {\filename};	
  
  \addplot [draw=none, fill=black, stack plots=y, fill opacity=0.15] table [
      x=n,
      y expr=2*\thisrow{F_err}
  ] {\filename}\closedcycle;
  
  \addplot [forget plot, stack plots=y,draw=none] table [x=n, y expr=-(\thisrow{F}+\thisrow{F_err})] {\filename};
    
  \addplot [draw=none, stack plots=y, forget plot] table[
    x=n,
    y expr=\thisrow{L}-\thisrow{L_err}
  ] {\filename};	
  
  \addplot [draw=none, fill=black, stack plots=y, fill opacity=0.15] table [
      x=n,
      y expr=2*\thisrow{L_err}
  ] {\filename}\closedcycle;
  
  \addplot [forget plot, stack plots=y,draw=none] table [x=n, y expr=-(\thisrow{L}+\thisrow{L_err})] {\filename};

	\addplot [solid, mark=None, color=black] table[x=n, y=L] {\filename};	
	\addplot [dashed, mark=None, color=black] table[x=n, y=F] {\filename};

  \end{axis}
\end{tikzpicture}
    \end{adjustbox}
  \end{subfigure}
	\begin{picture}(0,0)
		\put(18,47){\rlap{\makebox[1cm]{\fontsize{6}{0}\selectfont\sffamily\bfseries I}}}
		\put(18,-23){\rlap{\makebox[1cm]{\fontsize{6}{0}\selectfont\sffamily\bfseries J}}}
		\put(18,-93){\rlap{\makebox[1cm]{\fontsize{6}{0}\selectfont\sffamily\bfseries K}}}
	\end{picture}	
  \begin{subfigure}[c]{0.13\textwidth}
		\vspace{15pt}
    \begin{adjustbox}{trim=-35pt 0pt 0pt 25pt}
			\rotatebox{90}{\resizebox{\linewidth}{!}{
				\includegraphics[trim=5.5cm 10cm 10cm 10.5cm, clip=true,]{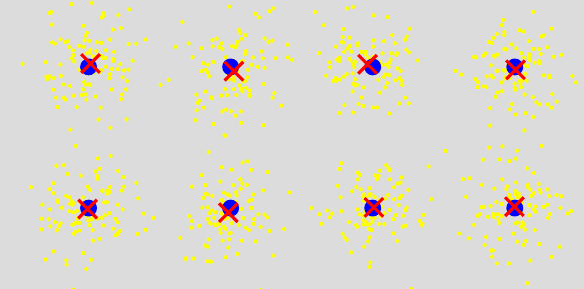}
			}}
    \end{adjustbox}
    \begin{adjustbox}{trim=-35pt 0pt 0pt 0pt}
			\rotatebox{90}{\resizebox{\linewidth}{!}{
				\includegraphics[trim=4cm 17cm 11.5cm 3.5cm, clip=true,]{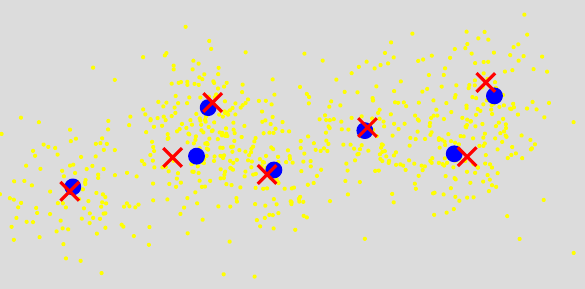}
			}}
    \end{adjustbox}
    \begin{adjustbox}{trim=-35pt 0pt 0pt 0pt}
			\rotatebox{90}{\resizebox{\linewidth}{!}{
				\includegraphics[trim=4cm 17cm 11.5cm 3.5cm, clip=true,]{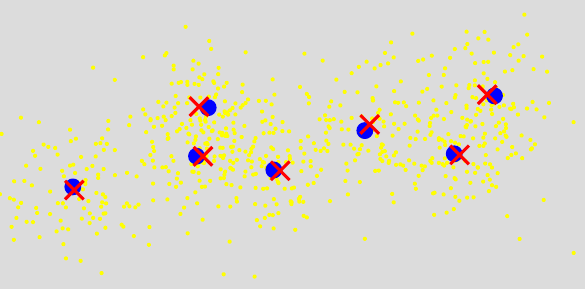}
			}}
    \end{adjustbox}
		\vspace{-120pt}
  \end{subfigure}
	%
	\newline
	%
	\begin{picture}(0,0)
		\put(2,-70){\rotatebox{90}{\rlap{\makebox[5.80cm]{\fontsize{7}{0}\selectfont\sffamily BIRCH ($25\times$ random pos.)}}}}
	\end{picture}	
  \begin{subfigure}[c]{0.193\textwidth}
    \begin{adjustbox}{trim=-2pt 10pt 0pt -5pt}
\tikzsetnextfilename{BIRCH_random-1-25-100_k-means_25}
\pgfplotsset{
	grid style={dotted,gray},
	minor grid style={dotted,lightgray},
  tick label style = {font=\tiny\sansmath\sffamily},
  legend style = {font=\sansmath\sffamily},
  xlabel style = {font=\sansmath\sffamily},
  ylabel style = {font=\sansmath\sffamily},
  legend image code/.code={
    \draw[mark repeat=2,mark phase=2]
    plot coordinates {
      (0cm,0cm)
      (0.25cm,0cm)        
      (0.5cm,0cm)         
    };%
  }
}	
\begin{tikzpicture}
	\tikzset{mark size={1.0}}
	\begin{axis}[
		colormap access=direct,
		width = 1.24\linewidth,
		height = 0.19 * \textheight,
		xmin=0,
		xmax=25,
		scaled x ticks = false,
		xlabel={\scriptsize\sffamily Iteration},
		xlabel near ticks, xticklabel pos=lower,
    x label style={at={(0.5,-0.09)}},
		ymin = -6.4,
		ymax = -5.7,
		ylabel near ticks, yticklabel pos=left,
		grid = both,
		%
    legend entries={
      \fontsize{6}{0}\selectfont\sffamily \, $\mathcal{L}$,
      \fontsize{6}{0}\selectfont\sffamily \, $\mathcal{F}$
    },
		legend style={
			at={(0.94,0.31)},
			legend columns=1,
			row sep=-2pt,
		},
		legend cell align=left,
    after end axis/.code={
          \draw [-latex, shorten <=-3pt] (axis cs: -1,-6.3283711253) node[left]{\fontsize{6}{0}\selectfont\sffamily $\mathcal{L}_{\textnormal{init}}$} to[out=360,in=180] (axis cs: 0,-6.3283711253);
          \draw [solid] (axis cs: 0,-5.77838223692) to (axis cs: 25,-5.77838223692);
          \draw [] (axis cs:4.8,-5.75338223692) node[left]{\fontsize{6}{0}\selectfont\sffamily $\mathcal{L}_{\textnormal{GT}}$};
          \node at (axis cs: -2,-5.715) {\fontsize{6}{0}\selectfont\sffamily\bfseries D};
    }
  	]
	\addlegendimage{solid}
	\addlegendimage{dashed}
	\addplot+ [thin, solid, mark=None, color=red!75!black] table[x=n, y=loglikelihood] {./figs/BIRCH_random-1-25-100_k-means_25-best.txt};
	\addplot+ [thin, dashed, mark=None, color=red!90!black] table[x=n, y=free-energy] {./figs/BIRCH_random-1-25-100_k-means_25-best.txt};
  \node[red!70!black, anchor=east] at (axis cs: 24,-5.76) {\fontsize{5}{0}\selectfont\sffamily all clusters recovered};
	\addplot+ [thin, solid, mark=None, color=cyan!75!black] table[x=n, y=loglikelihood] {./figs/BIRCH_random-1-25-100_k-means_25-1.txt};
	\addplot+ [thin, dashed, mark=None, color=cyan!90!black] table[x=n, y=free-energy] {./figs/BIRCH_random-1-25-100_k-means_25-1.txt};	
  \node[cyan!70!black, anchor=east] at(axis cs: 24,-5.828) {\fontsize{5}{0}\selectfont\sffamily 1 cluster not recovered};
	\addplot+ [thin, solid, mark=None, color=green!75!black] table[x=n, y=loglikelihood] {./figs/BIRCH_random-1-25-100_k-means_25-4.txt};
	\addplot+ [thin, dashed, mark=None, color=green!90!black] table[x=n, y=free-energy] {./figs/BIRCH_random-1-25-100_k-means_25-4.txt};	
	\node[green!70!black, anchor=east] at (axis cs: 24,-5.888) {\fontsize{5}{0}\selectfont\sffamily 2 clusters not recovered};
	\addplot+ [thick, solid, mark=None, color=grey!75!white] table[x=n, y=loglikelihood] {./figs/BIRCH_random-1-25-100_k-means_25.txt};
	\addplot+ [thick, dashed, mark=None, color=grey!60!white] table[x=n, y=free-energy] {./figs/BIRCH_random-1-25-100_k-means_25.txt};	
  \end{axis}
\end{tikzpicture}
    \end{adjustbox}
  \end{subfigure}
  \begin{subfigure}[c]{0.193\textwidth}
    \begin{adjustbox}{trim=-5pt 10pt 0pt -5pt}
\tikzsetnextfilename{BIRCH_random-1-25-100_k-means-Cprime}
\pgfplotsset{
	grid style={dotted,gray},
	minor grid style={dotted,lightgray},
  tick label style = {font=\tiny\sansmath\sffamily},
  legend style = {font=\sansmath\sffamily},
  xlabel style = {font=\sansmath\sffamily},
  ylabel style = {font=\sansmath\sffamily},
  legend image code/.code={
    \draw[mark repeat=2,mark phase=2]
    plot coordinates {
      (0cm,0cm)
      (0.25cm,0cm)        
      (0.5cm,0cm)         
    };%
  }
}	
\begin{tikzpicture}
	\tikzset{mark size={1.0}}
	\begin{axis}[
		colormap access=direct,
		width = 1.24\linewidth,
		height = 0.19 * \textheight,
		xmin=0,
		xmax=25,
		scaled x ticks = false,
		xlabel={\scriptsize\sffamily Iteration},
		xlabel near ticks, xticklabel pos=lower,
    x label style={at={(0.5,-0.09)}},
		ymin = -6.0,
		ymax = -5.75,
		ylabel near ticks, yticklabel pos=left,
		grid = both,
   legend image code/.code={%
     \draw[dash pattern=on 0.175cm off 0.05cm on 0.05cm off 0.05cm on 0.175cm] (0cm,0.05cm) -- (0.5cm,0.05cm);
     \draw[solid]  (0cm, 0.0cm) -- (0.5cm, 0.0cm);
    },		
    legend entries={
      \fontsize{5}{0}\selectfont\sffamily $C' = 1$ (k-means),
      \fontsize{5}{0}\selectfont\sffamily  $C' = 2$,
      \fontsize{5}{0}\selectfont\sffamily  $C' = 5$,
      \fontsize{5}{0}\selectfont\sffamily $C' = 25$ (GMM),
    },
		legend style={
			at={(0.94,0.52)},
			legend columns=1,
			row sep=-2pt,
		},
		legend cell align=left,
    after end axis/.code={
          \draw [solid] (axis cs: 0,-5.77838223692) to (axis cs: 25,-5.77838223692);
          \draw [-latex, shorten <=-3pt] (axis cs: -1,-5.77838223692) node[left]{\fontsize{6}{0}\selectfont\sffamily $\mathcal{L}_{\textnormal{GT}}$} to[out=360,in=180] (axis cs: 0,-5.77838223692);
          \node at (axis cs: -2,-5.755) {\fontsize{6}{0}\selectfont\sffamily\bfseries E};
    }
  	]
	\addplot+ [thin, solid, mark=None, color=red!75!black] table[x=n, y=loglikelihood] {./figs/BIRCH_random-1-25-100_k-means_25.txt};
	\addplot+ [thin, solid, mark=None, color=cyan!75!black] table[x=n, y=loglikelihood] {./figs/BIRCH_random-1-25-100_k-means-C'_25_2.txt};
	\addplot+ [thin, solid, mark=None, color=green!75!black] table[x=n, y=loglikelihood] {./figs/BIRCH_random-1-25-100_k-means-C'_25_5.txt};
	\addplot+ [thin, solid, mark=None, color=grey!75!black] table[x=n, y=loglikelihood] {./figs/BIRCH_random-1-25-100_k-means-C'_25_25.txt};
	\addplot+ [thin, dash pattern=on 0.35cm off 0.05cm on 0.05cm off 0.05cm, mark=None, color=red!75!black] table[x=n, y=loglikelihood] {./figs/BIRCH_random-1-25-100_k-means_25-best.txt};  
	\addplot+ [thin, dash pattern=on 0.35cm off 0.05cm on 0.05cm off 0.05cm, mark=None, color=cyan!75!black] table[x=n, y=loglikelihood] {./figs/BIRCH_random-1-25-100_k-means-C'_25_2-best.txt};  
	\addplot+ [thin, dash pattern=on 0.35cm off 0.05cm on 0.05cm off 0.05cm, mark=None, color=green!75!black] table[x=n, y=loglikelihood] {./figs/BIRCH_random-1-25-100_k-means-C'_25_5-best.txt};  
	\addplot+ [thin, dash pattern=on 0.35cm off 0.05cm on 0.05cm off 0.05cm, mark=None, color=grey!75!black] table[x=n, y=loglikelihood] {./figs/BIRCH_random-1-25-100_k-means-C'_25_25-best.txt};
	\draw [-latex, black!70!black] (axis cs: 22,-5.76) node[left]{\fontsize{5}{0}\selectfont\sffamily all clusters recovered} to (axis cs: 24,-5.7695);	
	\draw [-latex, black!70!black] (axis cs: 22,-5.76)  to (axis cs: 24,-5.78836487758);	
  \end{axis}
\end{tikzpicture}
    \end{adjustbox}
  \end{subfigure}
  \begin{subfigure}[c]{0.193\textwidth}
    \begin{adjustbox}{trim=-13pt 11pt 0pt -5pt}
\tikzsetnextfilename{BIRCH-random_k-means-Cprime_purity_main}
\pgfplotsset{
	grid style={dotted,gray},
	minor grid style={dotted,lightgray},
  tick label style = {font=\tiny\sansmath\sffamily},
  legend style = {font=\sansmath\sffamily},
  xlabel style = {font=\sansmath\sffamily},
  ylabel style = {font=\sansmath\sffamily},
  legend image code/.code={
    \draw[mark repeat=2,mark phase=2]
    plot coordinates {
      (0cm,0cm)
      (0.25cm,0cm)        
      (0.5cm,0cm)         
    };%
  }
}	
\begin{tikzpicture}
	\tikzset{mark size={1.0}}
	\begin{axis}[
		colormap access=direct,
		width = 1.24\linewidth,
		height = 0.19 * \textheight,
		xmin=0,
		xmax=25,
		xtick = {0,10,...,50},
		scaled x ticks = false,
		xlabel={\scriptsize\sffamily Iteration},
		xlabel near ticks, xticklabel pos=lower,
    x label style={at={(0.5,-0.09)}},
		ymin = 0.725,
		ymax = 0.875,
		ylabel near ticks, yticklabel pos=left,
		grid = both,
   legend image code/.code={%
     \draw[dash pattern=on 0.175cm off 0.05cm on 0.05cm off 0.05cm on 0.175cm] (0cm,0.05cm) -- (0.5cm,0.05cm);
     \draw[solid]  (0cm, 0.0cm) -- (0.5cm, 0.0cm);
    },		
    legend entries={
      \fontsize{5}{0}\selectfont\sffamily $C' = 1$ (k-means),
      \fontsize{5}{0}\selectfont\sffamily  $C' = 2$,
      \fontsize{5}{0}\selectfont\sffamily  $C' = 5$,
      \fontsize{5}{0}\selectfont\sffamily $C' = 25$ (GMM),
    },
		legend style={
			at={(0.94,0.52)},
			legend columns=1,
			row sep=-2pt,
		},
		legend cell align=left,
    after end axis/.code={
          \draw [solid] (axis cs: 0,0.8648) to (axis cs: 25,0.8648);
          \draw [] (axis cs:25,0.8648) node[inner sep=0.5pt, anchor=south east]{\fontsize{5}{0}\selectfont\sffamily $\mathrm{purity}_{\textnormal{GT}}$};
          \node at (axis cs: -2,0.87) {\fontsize{6}{0}\selectfont\sffamily\bfseries F};
    }
  	]
	\addplot+ [thin, solid, mark=None, color=red!75!black] table[x=n, y=purity] {./figs/data2019/BIRCH_random_1_25_results.txt};
	\addplot+ [thin, solid, mark=None, color=cyan!75!black] table[x=n, y=purity] {./figs/data2019/BIRCH_random_2_25_results.txt};
	\addplot+ [thin, solid, mark=None, color=green!75!black] table[x=n, y=purity] {./figs/data2019/BIRCH_random_5_25_results.txt};
	\addplot+ [thin, solid, mark=None, color=grey!75!black] table[x=n, y=purity] {./figs/data2019/BIRCH_random_25_25_results.txt};
	\addplot+ [thin, dash pattern=on 0.35cm off 0.05cm on 0.05cm off 0.05cm, mark=None, color=red!75!black] table[x=n, y=purity] {./figs/data2019/BIRCH_random_1_25_best_p_NMI.txt};  
	\addplot+ [thin, dash pattern=on 0.35cm off 0.05cm on 0.05cm off 0.05cm, mark=None, color=cyan!75!black] table[x=n, y=purity] {./figs/data2019/BIRCH_random_2_25_best_p_NMI.txt};  
	\addplot+ [thin, dash pattern=on 0.35cm off 0.05cm on 0.05cm off 0.05cm, mark=None, color=green!75!black] table[x=n, y=purity] {./figs/data2019/BIRCH_random_5_25_best_p_NMI.txt};  
	\addplot+ [thin, dash pattern=on 0.35cm off 0.05cm on 0.05cm off 0.05cm, mark=None, color=grey!75!black] table[x=n, y=purity] {./figs/data2019/BIRCH_random_25_25_best_p_NMI.txt};
  \end{axis}
\end{tikzpicture}
    \end{adjustbox}
  \end{subfigure}
	\begin{picture}(0,0)
		\put(22,-80){\rotatebox{90}{\rlap{\makebox[5.80cm]{\fontsize{7}{0}\selectfont\sffamily KDD}}}}
	\end{picture}	
	\begin{picture}(0,0)
		\put(18,50){\rlap{\makebox[5cm]{\fontsize{6}{0}\selectfont\sffamily  $k$-means-C' q.-error}}}
	\end{picture}
  \begin{subfigure}[c]{0.193\textwidth}
    \begin{adjustbox}{trim=-22pt 5pt 0pt -5pt}
\tikzsetnextfilename{KDD_k-means-Cprime_QE_main}
\pgfplotsset{
	grid style={dotted,gray},
	minor grid style={dotted,lightgray},
  tick label style = {font=\tiny\sansmath\sffamily},
  legend style = {font=\sansmath\sffamily},
  xlabel style = {font=\sansmath\sffamily},
  ylabel style = {font=\sansmath\sffamily},
  legend image code/.code={
    \draw[mark repeat=2,mark phase=2]
    plot coordinates {
      (0cm,0cm)
      (0.25cm,0cm)        
      (0.5cm,0cm)         
    };%
  }
}	
\begin{tikzpicture}
	\tikzset{mark size={1.0}}
	\begin{axis}[
		colormap access=direct,
		width = 1.24\linewidth,
		height = 0.19 * \textheight,
		xmin=0,
		xmax=200,
		scaled x ticks = false,
		xlabel={\scriptsize\sffamily Iteration},
		xlabel near ticks, xticklabel pos=lower,
    x label style={at={(0.5,-0.09)}},
		ymin = 1.37e11,
		ymax = 1.405e11,
		ylabel near ticks, yticklabel pos=left,
		grid = both,
   legend image code/.code={%
     \draw[dashed] (0cm,-0.025cm) -- (0.5cm,-0.025cm);
     \draw[solid]  (0cm, 0.025cm) -- (0.5cm, 0.025cm);
    },		
    legend entries={
      \fontsize{5}{0}\selectfont\sffamily \!\! $C'=1$ ($k$-means),
      \fontsize{5}{0}\selectfont\sffamily \!\! $C'=2$,
      \fontsize{5}{0}\selectfont\sffamily \!\! $C'=10$,
      \fontsize{5}{0}\selectfont\sffamily \!\! $C'=200$ (GMM)
    },
		legend style={
			at={(0.94,0.94)},
			legend columns=1,
			row sep=-2pt,
			column sep=2pt,
		},
		legend cell align=left,
    after end axis/.code={
      \node at (axis cs: -16,1.404e11) {\fontsize{6}{0}\selectfont\sffamily\bfseries H};
    }
  	]
  \addlegendimage{color=red!70!black}  	
	\addlegendimage{color=cyan!70!black}
  \addlegendimage{color=green!70!black}  	
  \addlegendimage{color=black}
	
  \newcommand\filename{./figs/KDD_k-means-C'_200_1.txt}
  \addplot [draw=none, stack plots=y, forget plot] table[
    x=n,
    y expr=\thisrow{phi}-\thisrow{phi_err}
  ] {\filename};	
  
  \addplot [draw=none, fill=red!70!black, stack plots=y, fill opacity=0.15] table [
      x=n,
      y expr=2*\thisrow{phi_err}
  ] {\filename}\closedcycle;
  
  \addplot [forget plot, stack plots=y,draw=none] table [x=n, y expr=-(\thisrow{phi}+\thisrow{phi_err})] {\filename};
    
	\addplot [solid, mark=None, color=red!70!black] table[x=n, y=phi] {\filename};	

  \renewcommand\filename{./figs/KDD_k-means-C'_200_2.txt}
  \addplot [draw=none, stack plots=y, forget plot] table[
    x=n,
    y expr=\thisrow{phi}-\thisrow{phi_err}
  ] {\filename};	
  
  \addplot [draw=none, fill=cyan!70!black, stack plots=y, fill opacity=0.15] table [
      x=n,
      y expr=2*\thisrow{phi_err}
  ] {\filename}\closedcycle;
  
  \addplot [forget plot, stack plots=y,draw=none] table [x=n, y expr=-(\thisrow{phi}+\thisrow{phi_err})] {\filename};
    
	\addplot [solid, mark=None, color=cyan!70!black] table[x=n, y=phi] {\filename};	

  \renewcommand\filename{./figs/KDD_k-means-C'_200_10.txt}
  \addplot [draw=none, stack plots=y, forget plot] table[
    x=n,
    y expr=\thisrow{phi}-\thisrow{phi_err}
  ] {\filename};	
  
  \addplot [draw=none, fill=green!70!black, stack plots=y, fill opacity=0.15] table [
      x=n,
      y expr=2*\thisrow{phi_err}
  ] {\filename}\closedcycle;
  
  \addplot [forget plot, stack plots=y,draw=none] table [x=n, y expr=-(\thisrow{phi}+\thisrow{phi_err})] {\filename};
    
	\addplot [solid, mark=None, color=green!70!black] table[x=n, y=phi] {\filename};	

  \renewcommand\filename{./figs/KDD_GMM.txt}
  \addplot [draw=none, stack plots=y, forget plot] table[
    x=n,
    y expr=\thisrow{phi}-\thisrow{phi_err}
  ] {\filename};	
  
  \addplot [draw=none, fill=black, stack plots=y, fill opacity=0.15] table [
      x=n,
      y expr=2*\thisrow{phi_err}
  ] {\filename}\closedcycle;
  
  \addplot [forget plot, stack plots=y,draw=none] table [x=n, y expr=-(\thisrow{phi}+\thisrow{phi_err})] {\filename};
    
	\addplot [solid, mark=None, color=black] table[x=n, y=phi] {\filename};	
  \end{axis}
\end{tikzpicture}
    \end{adjustbox}
  \end{subfigure}
  \vspace{1pt}
  \caption{
		\changed{\textsf{A\,-\,C} show results on a BIRCH data set with grid-positioned clusters, \textsf{D\,-\,F} on a BIRCH data set with randomly positioned clusters, and \textsf{G} and \textsf{H} on the KDD data set.
		The first column (plots \textsf{A} and \textsf{D}) shows log-likelihoods and free energies per data point for three individual runs of Alg.\,1 ($k$-means).
Red: all of the 25 clusters are found correctly; Blue: all but one; Green: all but two cluster. Additionally, the `grey' plot shows the mean of 100~independent runs. 
Plots \textsf{B} and \textsf{E} each show the mean log-likelihood (solid line) and the log-likelihood of the run with the highest final value (striped) based on 100~runs of Alg.\,3 ($k$-means-$C'$) for different $C'$.
		Plots \textsf{C} and \textsf{F} show the same for the purity, where we show the run with the highest sum of purity and NMI (additional plots for the NMI are given in Fig.\,\ref{fig:addExpA}).
		For comparison, we show results for DBSCAN, with free parameters optimized for highest combined purity and NMI. For BIRCH with random clusters, DBSCAN reaches a purity of $0.5$, 
which is hence not visible in plot \textsf{F}. Detailed results including comparisons with lazy-$k$-means, are given in Suppl.\,\ref{sec:suppl_details_experiments}.
		Plot \textsf{G} shows the mean log-likelihood and free energy (shaded with their respective SEMs) of $k$-means-$C'$ (Alg.\,3) with different $C'$ based on 10~individual runs each.
                \textsf{H}~visualizes the same runs but plots the quantization error instead. Visualizations of some ground truth cluster centers (blue circles) and found cluster centers of the best runs (red crosses) are shown in \textsf{I} ($k$-means, BIRCH $5\times 5$ grid), \textsf{J} ($k$-means, BIRCH $25\,\times$ random) and \textsf{K} ($k$-means-$C'$ with $C'=2$ on BIRCH $25\,\times$ random).}
    }
  \vspace{-10pt}    
	\label{ExpA}
\end{figure*}

Alg.\,3 ($k$-means-$C'$), for which Prop.\,4 applies, can be compared to soft-$k$-means \citep[][]{MacKay2003}, which was suggested as a `non-hard' $k$-means generalization.
Soft-$k$-means and $k$-means-$C'$ share an additional parameter for data variance.
For $k$-means-$C'$ this is the variance $\sigma^2$ itself, for soft-$k$-means this parameter is the `stiffness' parameter $\beta$, which also closely links to $\sigma^2$ (essentially $\beta=\frac{1}{2\sigma^2}$) of GMM (\ref{EqnGMMIso}).
However, $k$-means-$C'$ makes $k$-means `softer' by allowing for more than one non-zero value for the cluster assignments.
This is different from soft-$k$-means, which maintains non-zero values for all cluster assignments.
Related to this, problems with sensitivity to stiffness values and sensitivity to initial conditions compared to standard $k$-means \citep[][]{BarbakhFyfe2008} may be related to Prop.\,1 and Prop.\,2 which imply that for any approach with $C'>1$, updates of $\sigma^2$ should (in contrast to soft-$k$-means) not be neglected. 
%
$k$-means-$C'$ is itself closely related to the GMM algorithms of \citet[][]{SheltonEtAl2014} and \citet[][]{HughesSudderth2016}. But while \citet[][]{SheltonEtAl2014} and \citet[][]{HughesSudderth2016}
focus on EM acceleration, no proofs that their algorithms monotonically increase free energies are given (we elaborate in Suppl.\,\ref{sec:suppl_generalization}).

In general, other selection criteria than (\ref{EqnEuclid}) could be derived for other mixture models. Visa versa also other versions of $k$-means 
based on other criteria than (\ref{EqnEuclid}) can be interpreted as variational EM.
An example is {\em lazy-$k$-means} which is a relatively recent $k$-means generalization used to study convergence properties \citep[][]{HarPeledSadri2005}.
Lazy-$k$-means only reassigns a data point $n$ from a cluster $\tilde{c}$ to a new cluster $c$ if: 
\begin{align}
 (1+\eps)\,\|{}\yVecN\,-\,\muVec_c\|{}\,<\,\|{}\yVecN\,-\,\muVec_{\ct}\|{}\,, \label{EqnLazyKM}
\end{align}
where $\eps\geq{}0$ is small, and $k$-means is recovered for $\eps=0$. 

By considering Prop.\,1, any replacement of states in $\KK$ according to (\ref{EqnLazyKM}) would also increase the free energy (\ref{EqnTruncatedF}).
Based on our variational interpretation, lazy-$k$-means corresponds to a partial TV-E-step.
In analogy to Prop.\,1, we can show that (\ref{EqnTruncatedF}) is monotonically increased, but it is not necessarily maximized, i.e., Corollary\,1 does not apply.
However, the essential observation of a decoupled $\muVec$ and $\sigma^2$ update only depends on $C'$ being equal to one.
Prop.\,2 thus generalizes to the lazy-$k$-means case, and the same applies for Corollary\,2 (see Suppl.\,\ref{sec:suppl_lazyKM} for the proofs).
For lazy-$k$-means, polynomial running time bounds could be derived \citep[][]{HarPeledSadri2005}.
By virtue of Corollary\,2, this means that the corresponding log-likelihood bound can be optimized in polynomial time.
More generally, Corollary\,2 (as well as the other results) can serve for transferring many of the diverse run-time complexity results for $k$-means and $k$-means-like algorithms to results for GMM bounds.
Likelihood bounds are, on the other hand, of interest for theoretical studies of GMM optimization \citep[e.g.][]{KalaiEtAl2010,MoitraValiant2010,XuHsuMaleki2016}.
The here established link can thus serve to transfer results from $k$-means-like approaches \citep[e.g.][]{ArthurEtAl2009} to GMM clustering.
%

\changed{In this study, we have focused on $k$-means and its relation to GMMs with isotropic Gaussians and equal mixing proportions (Eqn.\,\ref{EqnGMMIso}).
The analytical tools applied here could be used similarly for general GMM densities. Also in the general case it would be possible
to define algorithms only considering the $C'$ most relevant clusters for updates. However, the criterion to assign clusters
to data points would diverge considerably from the closest cluster selection used by $k$-means. As a consequence, even when choosing
$C'$=$1$, a general GMM density would not result in a decoupling of $\muVec_c$ updates from the updates of the other model parameters.
We elaborate in Suppl.\,C.}

Finally, a very popular $k$-means version is {\em fuzzy $k$-means} \changedOld{\citep[e.g.,][for references]{Bezdek1981}}, which takes the form of a generalization
of the $k$-means objective (\ref{EqnKMObjective}) by using non-binary $\scn$ in the place of the $k$-means assignments.
Fuzzy $k$-means algorithms then update weighted cluster assignments and cluster centers in order to minimize such objectives.
%
Prop.\,4 serves best to highlight the differences between \changedOld{standard} fuzzy $k$-means and $k$-means-$C'$, because it shows that the average entropy
of the cluster assignments emerges in the context of GMMs as a term in addition to a softened objective. 
\changedOld{Standard algorithms for} fuzzy $k$-means \citep[e.g.][]{Bezdek1981,Yang1993} are different as they usually generalize the $k$-means objective
without an additional entropy term. \changedOld{Notably, newer versions of fuzzy $k$-means have been suggested to improve on earlier versions by introducing
additional regularization terms. One of these regularizations takes the form of the entropy of cluster assignments \citep[compare][Sec.\,2]{MiyamotoEtAl2008}.
Considering Eqn.\,\ref{EqnFreeEnergyKMeansC} of Prop.\,4, we could now relate the regularization constant of entropy regularized fuzzy $k$-means
to the GMM log-likelihood optimization, or introduce novel versions of fuzzy $k$-means with many weights set to `hard' zeros. Other, e.g., quadratic regularizations
\citep[see][]{MiyamotoEtAl2008} are, on the other hand, not as closely related to the GMM objective but may correspond to other data statistics.
}

\myvanish{
Fuzzy $k$-means is different from truncated variational generalizations as can be seen by considering the log-likelihood and the
free energy objectives derived here. Most notably, Prop.\,4 shows that the average entropy
of the cluster assignments emerges in the context of GMMs as a term in addition to a softened objective. 
Fuzzy $k$-means algorithms optimize objectives without assignment entropy, and are therefore different from the 
$k$-means-like algorithms considered here. Soft $k$-means, on the other hand, is more closely related than fuzzy $k$-means as it can be
shown to be derivable from an objective including an entropy term \citep[e.g.][]{KimEtAl2007}.
}
\section{Numerical Verification}
Before we conclude, we briefly numerically verify the main theoretical results of this work.
We use a BIRCH dataset 
with $C=25$ clusters on a $5\times{}5$ grid with $N=100$
data points per cluster (same data set for all runs) as partly shown in Fig.\,\ref{ExpA}I. Fig.\,\ref{ExpA}A shows different runs of standard $k$-means and the time course of the free energy and likelihood computed using (\ref{EqnFreeEnergyKMeans}) and (\ref{EqnGMMLikelihood}), respectively. The shown exemplary runs converge to different optima.
The run with highest final free energy recovers all cluster centers and results in a log-likelihood larger than the log-likelihood of the generating (ground-truth) parameters. We verified that this (small) overfitting effect decreases with increasing $N$.
The bound for the best run is relatively tight, which is consistent with (\ref{EqnDKLKMeans}) of Prop.\,3 for small $\sigma^2$.
The gap is larger for local optima, which have to have a larger $\sigma^2$ according to (\ref{EqnSigmaKMeans}) and consequently higher entropy for $\qcn$ of $C'>1$ including $C'=C$.
The gap also increases for clusters with larger overlap in Fig.\,\ref{ExpA}D/J, where we use the same setting as for Fig.\,\ref{ExpA}A but with randomly (uniformly) distributed cluster centers (see Fig.\,\ref{ExpA}J and \ref{ExpA}K).
Note that we use the seeding of 
$k$-means++ \citep[][]{ArthurVassilvitskii2006}
for Fig.\,\ref{ExpA}.
The initial values of $\LL(\Theta)$ are thus already relatively high (see $\LL_\text{init}$).
\comment{With uniform draws from the data points, initial likelihood values are lower ($\LL_\text{init}=...$ and $\LL_\text{init}=...$ for Fig.\,\ref{ExpA}A and B, respectively).}

Fig.\,\ref{ExpA}E shows different runs of $k$-means-$C'$ for the data as used for Fig.\,\ref{ExpA}J.
%
Using $k$-means-$C'$ with different numbers of winning clusters $C'$ can prevent shifted cluster centers caused by unsymmetrical cluster overlaps (compare Fig.\,\ref{ExpA}J and \ref{ExpA}K).
Final likelihoods of the best runs with $C' > 1$ can hence be higher than those for $k$-means. 
%
\changedOld{Fig.\,\ref{ExpA}E,J,K can also serve as numerical verification of the differences between free energies for different $C'$. Suppl.\,\ref{sec:suppl_details_experiments} elaborates on this.}
%
%
\changed{Figs.\,\ref{ExpA}C/F give additional results on the purity. Here we also compare to the popular DBSCAN method \citep{EsterEtAl1996}. While for the well separated grid data set the purity is comparably high, for the random set with larger overlaps, the purity for DBSCAN is with around 0.5 no longer comparable. More detailed results are given in Tab.\,\ref{tab:comparison}, where we also show results for lazy-$k$-means.}
Finally, Figs.\,\ref{ExpA}G and H verify our results using real and large scale data.
The KDD-Cup 2004 Protein Homology Task \citep[KDD,][]{CaruanaEtAl2004} comprises 145\,751 samples of 74-dimensional data points.
We observe tighter bounds between log-likelihood (\ref{EqnGMMLikelihood}) and free energy (\ref{EqnTruncatedF}) for better solutions of increasing $C'$.
Already for $C'=2$ the $D_{KL}$-gap decreases significantly relative to $k$-means and vanishes nearly completely for $C'=10$.

\removed{Remove this sentence: This also shows the applicability of Prop.\,1 for truncated posteriors, that regain close to non-truncated solutions already for small $C'$.}

%
%
%
%
\section{Conclusion}
We have established a novel and, we believe, very natural link between $k$-means and EM for GMMs by showing that $k$-means is a special case of a truncated variational EM approximations for GMMs.
The link can serve to transfer theoretical research between $k$-means-like and GMM clustering approaches (Sec.\,\ref{SecApp} treated some examples).
%
%
Of the many studies which consider $k$-means and data samples of GMMs \citep[e.g.][\& refs.\ therein]{ChaudhuriEtAl2009}, there is none that provides the close theoretical links and
free energy results provided here (also see Suppl.\,\ref{sec:Difference}). Earlier work by \citet[][]{Pollard1982} is maybe one of the most relevant studies,
as it proves a theorem which relates the convergence points of $k$-means to an
underlying distribution. In the sense of a central limit theorem, this distribution is given by a GMM with clusters of specific covariance.
Cluster overlap in the samples influences the cluster shapes via non-zero off-diagonal elements. The question of \citet[][]{Pollard1982} is thus how
to fit a GMM (in a central limit theorem sense) to correspond to $k$-means convergence points. Prop.\,3 may be related to the theorem of \citet[][]{Pollard1982}
but a closer inspection would require a more extensive analysis.\vspace{-1mm}

\changed{Other than the above discussed theoretical link of $k$-means to GMM clustering, our investigations may also be useful 
for the analysis and improvement of further aspects of $k$-means-like and GMM clustering. 
GMMs are used to address a wide range of tasks. Two examples may be image denoising \citep[e.g.][]{ZoranWeiss2011} and video tracking \citep[e.g.][]{JepsonEtAl2003,LanEtAl2015,LanEtAl2018}. Training $k$-means may, however, often be more efficient, which can be of importance, e.g., when a lot of data has to be processed in short times. By assigning a probabilistic interpretation to $k$-means, it may offer itself
as a faster alternative to GMMs in such settings. Similarly, $k$-means-$C'$ could be used as a compromise between GMMs and efficient $k$-means versions.
A further aspect our results can be related to is the estimation of cluster numbers. The standard $k$-means algorithm (Alg.\,1), standard EM for GMMs (Alg.\,2) as well
as $k$-means-$C'$ (Alg.\,3) require the number of clusters as input. A large number of studies have addressed this disadvantage of the standard approaches.
Model selection and fully Bayesian approaches \citep[][]{FraleyRaftery1998,Rasmussen2000,Neal2000} are common methods to estimate the cluster numbers of GMMs from data.
For $k$-means, well known contributions are the $X$-means algorithm \citep[][]{PellegMoore2000}, the $G$-means
algorithm \citep[][]{HamerlyElkan2004} as well as approaches based on clustering stability \citep[see][\& refs.\ therein]{Luxburg2010}. 
All the approaches for $k$-means use standard $k$-means iterations or full $k$-means runs as part of the complete algorithm, e.g., 
as subroutines in split-and-merge approaches \citep[][\& refs.\ therein]{UedaEtAl2000}. There are different options how the results of this study can be combined with these previous studies.
For $X$-means-like approaches, our results (e.g., Eqn.\,14) could be used to quantify how well the BIC selection criterion used by $X$-means can be expected to work.
If for a given data set $k$-means is not well approximating GMM solutions (e.g., for larger cluster overlaps), $k$-means-$C'$ iterations would offer
themselves as alternative iterations within an $X$-means setting. 
Less directly, $k$-means-$C'$ algorithms could (A)~be used in conjunction
with statistical tests for Gaussianity of projected data as in $G$-means, or (B)~they could be used (like $k$-means) to define stability scores for
stability-based selections of cluster numbers. Also in these two cases, improvements can be expected especially when cluster overlaps are large.
Finally, $k$-means and $k$-means-$C'$ could be combined with general Bayesian model selection \citep[][]{Schwarz1978} as their
free energies (Eqns.\,\ref{EqnResult} and \ref{EqnFreeEnergyKMeansC}, respectively) provide likelihood approximations.

More generally, $k$-means is usually not directly integrated into probabilistic frameworks as} the limit to zero cluster variance remained the most well known relation between $k$-means and GMMs.
From the probabilistic point of view, this limit is unsatisfactory, however, as the likelihood of data points under a GMM with $\sigma^2\rightarrow{}0$ also approaches zero. 
Truncated approaches (which allow for a $k$-means/GMM relation with finite variances $\sigma^2>0$) are novel compared to standard variational approaches which assume a-posteriori independence \citep[e.g.][]{SaulEtAl1996,Jaakkola2001}. Truncated EM approaches \citep[][]{LuckeEggert2010,SheikhEtAl2014,Lucke2018} aim at scalable and accurate approximations without assuming a-posteriori independence; a goal they share with many later approaches \citep[e.g.][]{MnihGregor2014,RezendeMohamed2015,SalismansEtAl2015,KucukelbirEtAl2016}.
Truncated EM is a natural variational approximation for $k$-means-like algorithms, and is here not only related but becomes, indeed, identical to standard $k$-means.\vspace{2mm} 


\vspace{-17pt}
\section*{Acknowledgements}
\vspace{-8pt}
We acknowledge funding by the DFG projects SFB 1330 (B2) and EXC 2177/1 (cluster of excellence H4a 2.0).
\vspace{-9pt}

%
%
%
%
%
%
%
\removed{Remove last sentence:
One of the implications following from this observations is a direct quantitative link between the $k$-means and GMM objectives (e.g.,\ Eqn.\,\ref{EqnResult}), which we believe may be of potentially high relevance for further theoretical and empirical studies.}
%
%
%
%
{
\renewcommand\baselinestretch{0.5}
\setlength{\itemsep}{-6.1pt}
%
%

}

\clearpage

\appendix
\renewcommand*{\thepage}{S\arabic{page}}
\setcounter{page}{1}  
\renewcommand*{\thefigure}{S\arabic{figure}}
\setcounter{figure}{0}  
\renewcommand*{\thetable}{S\arabic{table}}
\setcounter{table}{0}  
\renewcommand\appendixname{Supplementary}
\renewcommand\thesection{\Alph{section}}

\noindent{\Large \bfseries Supplement}

\section{Illustration of truncated posterior approximations}
\label{sec:Illustration}
\begin{figure*}[tb]
\begin{center}
	\resizebox{0.95\textwidth}{!}{
		\begin{adjustbox}{trim=0pt 20pt 0pt -0pt}%
      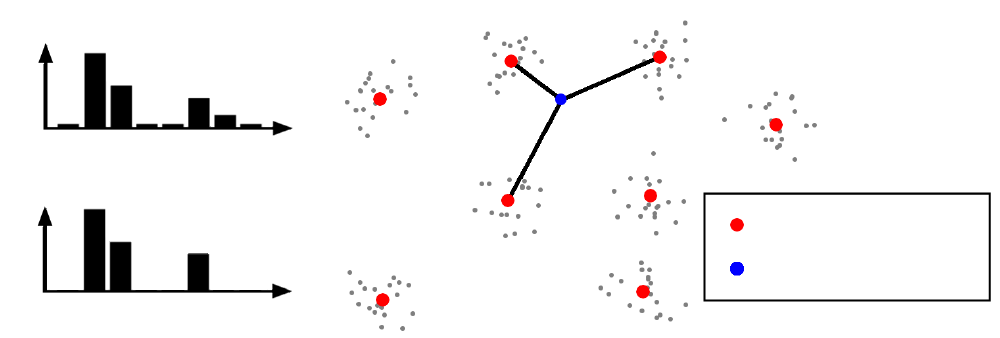
    \end{adjustbox}
	}
	\vspace{20pt}
\end{center}
\caption{Illustration of truncated distributions for a GMM (\ref{EqnGMMIso}) in $D=2$ dimensions. The figure considers a data point $\yVecN$ which lies (for illustrative purposes) to some extend in between some clusters. The full posterior (the responsibilities) $\rcn=p(c\,|\,\yVecN,\Theta)$ for the $C=8$ clusters are shown in the top-left. Below, a truncated approximation $\qcn$ with $|\KKn|=C'=3$ is shown for
the same data point. The truncated approximation maintains the $C'$ highest posterior values, sets all others to zero, and renormalizes the distribution to sum to one. The three closest clusters, which correspond to the three highest posterior values, are connected with black lines in the main figure.}
\label{FigVarGMM}
\end{figure*}

Fig.\,\ref{FigVarGMM} illustrates truncated distributions for an example with two-dimensional data points ($D=2$) with $C=8$ clusters. As can be observed, the truncated distributions with $C'=3$ is capturing
the posterior structure for data point $n$ well. For basically all data points (grey dots), truncated distributions with $C'=3$ are sufficiently exact; and for most data points $C'=2$ already represent
a very good approximations. Also the case $C'=1$, which correspond to the $k$-means case, will sufficiently well model the posterior because for most data points in this example
the posterior is dominated by the value of the closest cluster. Also see Figs.\,\ref{fig:addExpA} and \ref{fig:supp_ExpA} for numerical verifications.

\section{$k$-Means and hard cluster assignments for GMMs}
\label{sec:Difference}
Here we provide more details on how $k$-means or the $k$-means objective has previously been related to maximum likelihood optimization of GMMs.\\
%
%

\noindent{}{\em Classification expectation maximization.} The log-likelihood objective of GMMs (\ref{EqnGMMLikelihood}) and the quantization error (\ref{EqnKMObjective}) optimized by $k$-means are non-trivially related.
This is also the case for the GMMs with isotropic and equal Gaussian variances and equal mixing proportions as considered here (Eq.\,\ref{EqnGMMIso}).
For the purposes of our study we emphasize this point as earlier contributions reported results for clustering
criteria from which one may incorrectly infer a trivial relation between (\ref{EqnKMObjective}) and (\ref{EqnGMMLikelihood}). One example of such previous work \citep[see][and references therein]{CeleuxGovaert1992} does, for instance, consider a {\em classification expectation maximization} (CEM) algorithm for clustering. The paper defines a
{\em classification maximum likelihood} (CML) objective which is (in the notation of this paper) given by:
\begin{align}
  \LL^{\,\mathrm{CML}}(\Theta) &= \frac{1}{N}\sum_{n=1}^N\sum_{c=1}^C \scn \log\!\Big( \frac{1}{C}\,\Ncal\big(\yVecN;\muVec_c,\sigma^2\One\big)\Big),\label{EqnCML}
\end{align}
where the $\scn$ are the binary weights of Alg.\,1.
In the paper \citep[][]{CeleuxGovaert1992} it is then shown that the problem of maximizing the CML objective (\ref{EqnCML}) is equivalent to the problem of minimizing the quantization error (\ref{EqnKMObjective}).
Although (\ref{EqnCML}) is also referred to as a maximum likelihood (ML) objective \citep[see][and references therein]{CeleuxGovaert1992}, note the difference between this {\em classification} maximum likelihood (CML) objective (\ref{EqnCML}) and the standard ML objective for GMMs in (\ref{EqnGMMLikelihood}). Essentially, the sum over clusters in (\ref{EqnGMMLikelihood}) and the logarithm can not be trivially commuted to obtain (\ref{EqnCML}). 
Eq.\,(\ref{EqnDKLKMeans}) can be regarded as quantification of the
difference between (\ref{EqnCML}) and (\ref{EqnGMMLikelihood}) in terms of the ratio between data-to-cluster center distances and $\sigma$ \citep[compare initial discussion of][]{CeleuxGovaert1992}. Eq.\,(\ref{EqnDKLKMeans}) is ultimately a consequence of applying Jensen's inequality to commute logarithm and the sum over clusters, which gives rise to a lower free energy bound. 
Only if cluster centers are far apart compared to $\sigma$, the sum over $c$ will for each data point $n$ be dominated by the terms of within cluster distances. This is the case of well separable clusters, i.e., if `hard' data partitions \changed{are} representing a good approximation of `soft' {\em a-posterior assignments}. In that case the KL-divergence becomes zero. 
Also see Supplement\,\ref{sec:suppl_details_experiments} below for numerical experiments showing differences between the $k$-means and log-likelihood objectives.

\ \\
\noindent{}{\em Hard cluster assignments and variational distributions.} 
As discussed in the main text, the by far most common approach to relate $k$-means and Gaussian mixture models is to take the limit to zero cluster variances $\sigma^2\rightarrow{}0$.
This relation is very commonly used in text books as well as in the research literature itself. 
Alternatively, and related to this study, $k$-means is for didactic purposes also sometimes casually related to GMM optimization using variational EM. Such
a relation is usually confined to derivations that make the relation of $k$-means to GMM data models plausible. For instance, if the free energy w.r.t.\ a variational distribution is
in its standard form given by 
\begin{align}
\FF(q,\Theta) &= \disT \frac{1}{N}\sum_{n=1}^{N}\Big(\sum_{c=1}^C \qn(c) \log\big(p(c,\yVecN\,|\,\Theta)\big)\nonumber\\
&\phantom{iiiiiiiiiiiii}\disT\,-\,\sum_{c=1}^C \qn(c) \log\big(\qn(c)\big)\Big)
\end{align}
then one can informally define $\qn(c)$ to be equal to one if and only if $c$ corresponds to the maximal value of $\qn(c)$. For GMM~(\ref{EqnGMMIso}) $\qn(c)$ are then given by:
\begin{align}
\qn(c) & =\left\{\begin{array}{cl}
1 & \mbox{if }\forall{}c'\neq{}c:\| \yVecN -\muVec_c \|\hspace{0mm} < \hspace{0mm}\| \yVecN -\muVec_{c'} \|\\[1mm]
0 & \mbox{otherwise}
\end{array}
\right..\label{EqnHardQ}
\end{align}
As the entropy for such a distribution is equal to zero, the free energy reduces to 
\begin{align}
\FF(q,\Theta) &= \disT \frac{1}{N}\sum_{n=1}^{N} \log\big(p(c_o^{(n)},\yVecN\,|\,\Theta)\big)\label{EqnLastF}
\end{align}
where $c_o^{(n)}$ denotes the cluster closest to data point $\yVecN$. $k$-means is then often taken as optimizing this objective.  

In order to make any mathematically rigorous statements, the argumentation above lacks, at closer inspection, a solid theoretical foundation in two important aspects:
(A)~Derivations of the free energy using variational distributions all assume $\qn(c)$ to be strictly positive ($\qn(c)>0$ for all $n$ and $c$), which is violated for $\qn(c)$ defined 
as in Eq.\,(\ref{EqnHardQ}). (B)~Relating $k$-means to a free energy objective as (\ref{EqnLastF}) implicitly assumes the variational distributions $\qn(c)$ to be independent of the model parameters~$\Theta$ (i.e., independent of $\muVec_{1:C}$ and $\sigma^2$ in our case). Considering Eq.\,(\ref{EqnHardQ}) also this independence is not given (which is in contrast, e.g., to mean field distributions). The model parameters can also not simply be assumed
to be constant as is the case for full posteriors in standard EM: The proof verifying that values for the model parameters can be held fixed is given for full posteriors only \citep[see, e.g., Lemma~1 of][]{NealHinton1998} but it does not necessarily apply for general variational distributions $\qn(c)$ defined using model parameters $\Theta$.

The here applied results \citep[][]{Lucke2018} do address both these aspects: variational distributions with `hard' zeros are treated (Point~A), and variational distributions that can depend on the model parameters are explicitly considered (Point~B). Addressing any of these two points is non-trivial (see Propositions\,1 and 2 in \cite{Lucke2018}, for Point~A; and, e.g., Propositions\,\mbox{3-5} in \cite{Lucke2018}, for Point~B). However, if treated rigorously, results for a large class of distributions (which includes distributions of Eq.\,\ref{EqnHardQ}) can be derived, and the derived results apply for any generative model with discrete latents. Notably, also truncated variational distributions with non-zero entropy are included as well as distributions (\ref{EqnHardQ}) in which $\qn(c)=1$ does not necessarily apply for the closest cluster (such distributions are important, e.g., in relation to {\em lazy-$k$-means}, see Proposition\,4). In this paper we make use of results of \citet{Lucke2018} by applying them to GMMs given by
Eq.\,(\ref{EqnGMMIso}) (e.g., through Propositions\,1 and 4 which in turn use the simplified free energy (\ref{EqnTruncatedF}) of \cite{Lucke2018}). 

The difficulties to cleanly and rigorously treat distributions such as (\ref{EqnHardQ}) may explain why (to the knowledge of the authors) any relation of $k$-means and variational approaches is rather informally discussed
(compare, e.g., \cite{JordanEtAl1997}, who, e.g., relate Viterbi training to variational EM). If the relation of $k$-means to GMMs is made more explicit, the literature, including popular text books 
\citep[e.g.][]{MacKay2003,Barber2012}, drops back to the zero variance limit to derive $k$-means.

\section{Generalization of criterion (\ref{EqnEuclid}) for general GMMs}
\label{sec:suppl_generalization}
\vspace{-2mm}
Consider a general standard GMM of the form:
\begin{align}
  p(c\,|\,\Theta) &= \pi_c \quad \text{with } \disT\sum_{c=1}^C\pi_c=1,  \label{EqnGMMA}\\
  p(\yVec\,|\,c,\Theta) &= |2\pi\Sigma_c|^{-\frac{1}{2}} \exp\!\big(\!-\!\disT\frac{1}{2}\|\yVec-\muVec_c\|^2_{\Sigma_c}\big),\label{EqnGMMB}\\
  \|\yVec-\muVec_c\|^2_{\Sigma_c} &= (\yVec-\muVec_c)^\mathrm{T}\Sigma^{-1}_c(\yVec-\muVec_c),\label{EqnGMMC}
\end{align}
where $\pi_c\geq{}0$ are the mixing proportions, $\Sigma_c$ is a for each $c$ positive definite covariance matrix, and $|\cdot|$ denotes the determinant.
We denote by $\Theta=(\pi_{1:C},\muVec_{1:C},\Sigma_{1:C})$ the set of all parameters.
%
For GMM (\ref{EqnGMMA}) to (\ref{EqnGMMC}) a corresponding variational free energy is because of Eq.\,(\ref{EqnTruncatedF}) \changed{(first line)} increased if and only if:
\allowdisplaybreaks
\begin{align*}
&&&p(c,\yVec\,|\,\Theta) > p({\ct},\yVec\,|\,\Theta) \\
&\Leftrightarrow &&  \disT \pi_c\,|2\pi\Sigma_c|^{-\frac{1}{2}} \exp\!\big(\!-\!\frac{1}{2}\|\yVec-\muVec_c\|^2_{\Sigma_c}\big) \\[-3pt]
&& >\; & \disT \pi_{\ct}\,|2\pi\Sigma_{\ct}|^{-\frac{1}{2}} \exp\!\big(\!-\!\frac{1}{2}\|\yVec-\muVec_{\ct}\|^2_{\Sigma_{\ct}}\big)\\
&\Leftrightarrow && \disT \log(\pi_c) - \frac{1}{2}\log(|2\pi\Sigma_c|) -\frac{1}{2}\|\yVec-\muVec_c\|^2_{\Sigma_c} \\[-3pt]
&& >\; &\disT \log(\pi_{\ct}) - \frac{1}{2}\log(|2\pi\Sigma_{\ct}|) -\frac{1}{2}\|\yVec-\muVec_{\ct}\|^2_{\Sigma_{\ct}}\\
&\Leftrightarrow && \|\yVec-\muVec_c\|^2_{\Sigma_c} + \log(|2\pi\Sigma_c|) - 2\log(\pi_c)\\[-3pt]
&&<\; & \|\yVec-\muVec_{\ct}\|^2_{\Sigma_{\ct}} + \log(|2\pi\Sigma_{\ct}|) - 2\log(\pi_{\ct})\,. \label{EqnGenCriterion}\numberthis
\end{align*}
\vspace{0mm}

In comparison, \citet[][]{SheltonEtAl2014} use an estimated E-step, which does consequently not guarantee a monotonic increase of a free energy.
\citet[][]{HughesSudderth2016} do use a constrained likelihood optimization to find the best $C'$ clusters per data point (related to Corollary\,1), but a complete proof for general GMMs would require Proposition\,5 of \citet[][]{Lucke2018}, which warrants that M-steps can be derived while the \changedOld{parameters} $\Theta$ of the variational distributions $\qcn$ remain fixed.

\begin{table*}[tbh]
\caption{\changed{Log-likelihood per data point $\mathcal{L}$, quantization error $\phi$, purity and normalized mutual information (NMI) for $k$-means, $k$-means-C' with $C'=2$, lazy-$k$-means with $\epsilon=0.1$ and DBSCAN on the BIRCH data sets.
The free parameters of DBSCAN were optimized to maximize the sum of NMI and purity.
Using such a combination for parameter tuning prevents settings highly overfitted to one of the two criteria with high trade-offs on the other.
Given are the means over 100~independent runs as well as the values of the best single run.
The mean and the best are identical for DBSCAN given the same, optimized free parameters. The best values per column are written in bold.
}}
\vspace{-15pt}
\label{tab:comparison}
\begin{center}
\newcommand{\measa}{\mathcal{L}}
\newcommand{\measb}{\phi}
\newcommand{\measc}{\mathrm{purity}}
\newcommand{\measd}{\mathrm{NMI}}
\newcommand{\mean}{mean}
\newcommand{\best}{best}
\small
\resizebox{\textwidth}{!}{
\begin{tabular}{c | c c| c c| c c| c c| c c| c c| c c| c c}
\toprule
\multicolumn{1}{c}{Algorithm} & \multicolumn{8}{c}{BIRCH $5\times 5$ (grid)}&\multicolumn{8}{ c }{BIRCH ($25\,\times$ random positions)}\\
\cmidrule(lr){1-1}
\cmidrule(lr){2-9}
\cmidrule(lr){10-17}
\multicolumn{1}{c}{}&\multicolumn{2}{c}{$\measa$}&\multicolumn{2}{c}{$\measb$}&\multicolumn{2}{c}{$\measc$}&\multicolumn{2}{c}{$\measd$}&
\multicolumn{2}{c}{$\measa$}&\multicolumn{2}{c}{$\measb$}&\multicolumn{2}{c}{$\measc$}&\multicolumn{2}{c}{$\measd$}\\
\cmidrule(lr){2-3}
\cmidrule(lr){4-5}
\cmidrule(lr){6-7}
\cmidrule(lr){8-9}
\cmidrule(lr){10-11}
\cmidrule(lr){12-13}
\cmidrule(lr){14-15}
\cmidrule(lr){16-17}
\multicolumn{1}{c|}{}&\mean&\best&\mean&\best&\mean&\best*&\mean&\best*&\mean&\best&\mean&\best&\mean&\best*&\mean&\best*\\
\midrule
$k$-means      &-6.127 					&\textbf{-6.016} &5,503 				&\textbf{4,836} &0.971 					&\textbf{0.992} &0.977 					&\textbf{0.987} &-5.842 				 & -5.789 				 & \textbf{4,535} & \textbf{4,291} & \textbf{0.819} &0.856 					& 0.879 					&0.875\\
$k$-means-C'   &\textbf{-6.117} &\textbf{-6.016} &5,476 				&4,837 					&0.973 					&\textbf{0.992} &0.978 					&0.986 					&\textbf{-5.828} & \textbf{-5.771} & 4,637 					& 4,331 		 & 0.811 					&\textbf{0.864} & \textbf{0.880}  &\textbf{0.880}\\
lazy-$k$-means &\textbf{-6.117} &\textbf{-6.016} &\textbf{5,452}&4,837 					&0.974				  &\textbf{0.992} &0.978 					&\textbf{0.987} &-5.850 				 & -5.803 				 & 4,592 					& 4,351 		 & 0.809 					&0.846 					& 0.876 					&\textbf{0.880}\\
DBSCAN 	       & --    					& --    				 & --   				& --   					&\textbf{0.989} &0.989 					&\textbf{0.982} &0.982 					& --   					 & --     				 & --    					& --    	  & 0.502 				&0.502 				 & 0.800					& 0.800\\
\bottomrule
\end{tabular}
}
\end{center}
\vspace{-7pt}
\footnotesize
*: best values for purity and NMI are for all algorithms given as values of the run with the highest sum of purity and NMI to omit solutions highly overfitted to one of the two criteria
\vspace{2pt}
\end{table*}

\begin{figure*}[tb]
  \begin{adjustbox}{trim=-35pt 0pt 0pt 0pt}
\tikzsetnextfilename{BIRCH_k-means-Cprime_legend}
\pgfplotsset{
	grid style={dotted,gray},
	minor grid style={dotted,lightgray},
  tick label style = {font=\tiny\sansmath\sffamily},
  xlabel style = {font=\sansmath\sffamily},
  ylabel style = {font=\sansmath\sffamily},
  legend image code/.code={
    \draw[mark repeat=2,mark phase=2]
    plot coordinates {
      (0cm,0cm)
      (0.25cm,0cm)        
      (0.5cm,0cm)         
    };%
  },
}	
\begin{tikzpicture} 
			%
    \begin{axis}[%
    hide axis,
    xmin=10,
    xmax=50,
    ymin=0,
    ymax=0.4,
		legend columns=4,
		row sep=-2pt,
    legend image code/.code={%
      \draw[dash pattern=on 0.175cm off 0.05cm on 0.05cm off 0.05cm on 0.175cm] (0cm,0.05cm) -- (0.5cm,0.05cm);
      \draw[solid]  (0cm, 0.0cm) -- (0.5cm, 0.0cm);
      \draw[dashed] (0cm,-0.05cm) -- (0.5cm,-0.05cm);
    },				
    legend style={draw=white!15!black,legend cell align=left}
    ]
		\addlegendimage{color=red!70!black}  	
		\addlegendimage{color=cyan!70!black}
		\addlegendimage{color=green!70!black}  	
		\addlegendimage{color=black}
		\addlegendentry{\fontsize{5}{0}\selectfont\sffamily $C' = 1$ (k-means) \hphantom{\hspace{10pt}}};
    \addlegendentry{\fontsize{5}{0}\selectfont\sffamily $C' = 2$ \hphantom{\hspace{10pt}}};
    \addlegendentry{\fontsize{5}{0}\selectfont\sffamily $C' = 5$ \hphantom{\hspace{10pt}}};
    \addlegendentry{\fontsize{5}{0}\selectfont\sffamily $C' = 25$ (GMM)};
    \end{axis}		
\end{tikzpicture}
  \end{adjustbox}
	\newline
	\begin{picture}(0,0)
		\put(-0,-75){\rotatebox{90}{\rlap{\makebox[5.80cm]{\fontsize{8}{0}\selectfont\sffamily BIRCH~$5\times 5$ (grid)}}}}
	\end{picture}
	\hspace{18pt}
  \begin{subfigure}[c]{0.22\textwidth}
    \begin{adjustbox}{trim=10pt 5.5pt 0pt 0pt}
\tikzsetnextfilename{BIRCH-1-5-100_k-means-Cprime_supp}
\pgfplotsset{
	grid style={dotted,gray},
	minor grid style={dotted,lightgray},
  tick label style = {font=\tiny\sansmath\sffamily},
  legend style = {font=\sansmath\sffamily},
  xlabel style = {font=\sansmath\sffamily},
  ylabel style = {font=\sansmath\sffamily},
  legend image code/.code={
    \draw[mark repeat=2,mark phase=2]
    plot coordinates {
      (0cm,0cm)
      (0.25cm,0cm)        
      (0.5cm,0cm)         
    };%
  }
}	
\begin{tikzpicture}
	\tikzset{mark size={1.0}}
	\begin{axis}[
		colormap access=direct,
		width = 1.2\linewidth,
		height = 0.21 * \textheight,
		xmin=0,
		xmax=25,
		scaled x ticks = false,
		xlabel near ticks, xticklabel pos=lower,
    x label style={at={(0.5,-0.09)}},
		ymin = -6.4,
		ymax = -5.95,
		ylabel near ticks, yticklabel pos=left,
		grid = both,
    legend entries={
      \fontsize{5}{0}\selectfont\sffamily $\mathcal{L}$ (mean),
      \fontsize{5}{0}\selectfont\sffamily $\mathcal{L}$ (highest)
    },
		legend style={
			at={(0.94,0.26)},
			legend columns=1,
			row sep=-2pt,
		},
		legend cell align=left,
    after end axis/.code={
          \draw [solid] (axis cs: 0,-6.0378687151) to (axis cs: 25,-6.0378687151);
          \draw [-latex, shorten <=-3pt] (axis cs: -1,-6.0378687151) node[left]{\fontsize{6}{0}\selectfont\sffamily $\mathcal{L}_{\textnormal{GT}}$} to[out=360,in=180] (axis cs: 0,-6.0378687151);
    }
  	]
	\addlegendimage{solid}		
	\addlegendimage{dash pattern=on 0.175cm off 0.05cm on 0.05cm off 0.05cm on 0.175cm}
	\addplot+ [thin, solid, mark=None, color=red!75!black] table[x=n, y=loglikelihood] {./figs/BIRCH-1-5-100_k-means_25.txt};
	\addplot+ [thin, solid, mark=None, color=cyan!75!black] table[x=n, y=L] {./figs/data2019/BIRCH_2_25_results.txt};
	\addplot+ [thin, solid, mark=None, color=green!75!black] table[x=n, y=L] {./figs/data2019/BIRCH_5_25_results.txt};
	\addplot+ [thin, solid, mark=None, color=grey!75!black] table[x=n, y=L] {./figs/data2019/BIRCH_25_25_results.txt};
	\addplot+ [thin, dash pattern=on 0.35cm off 0.05cm on 0.05cm off 0.05cm, mark=None, color=red!75!black] table[x=n, y=L] {./figs/data2019/BIRCH_1_25_best_LL.txt};  
	\addplot+ [thin, dash pattern=on 0.35cm off 0.05cm on 0.05cm off 0.05cm, mark=None, color=cyan!75!black] table[x=n, y=L] {./figs/data2019/BIRCH_2_25_best_LL.txt};  
	\addplot+ [thin, dash pattern=on 0.35cm off 0.05cm on 0.05cm off 0.05cm, mark=None, color=green!75!black] table[x=n, y=L] {./figs/data2019/BIRCH_5_25_best_LL.txt};  
	\addplot+ [thin, dash pattern=on 0.35cm off 0.05cm on 0.05cm off 0.05cm, mark=None, color=grey!75!black] table[x=n, y=L] {./figs/data2019/BIRCH_25_25_best_LL.txt};
  \node[black, anchor=east] at (axis cs: 22,-6.0) {\fontsize{5}{0}\selectfont\sffamily all clusters recovered};
  \end{axis}
\end{tikzpicture}
    \end{adjustbox}
  \end{subfigure}
	\hspace{5pt}
  \begin{subfigure}[c]{0.22\textwidth}
    \begin{adjustbox}{trim=10pt 0pt 0pt 0pt}
\tikzsetnextfilename{BIRCH_k-means-Cprime_QE_supp}
\pgfplotsset{
	grid style={dotted,gray},
	minor grid style={dotted,lightgray},
  tick label style = {font=\tiny\sansmath\sffamily},
  xlabel style = {font=\sansmath\sffamily},
  ylabel style = {font=\sansmath\sffamily},
  legend image code/.code={
    \draw[mark repeat=2,mark phase=2]
    plot coordinates {
      (0cm,0cm)
      (0.25cm,0cm)        
      (0.5cm,0cm)         
    };%
  }
}	
\begin{tikzpicture}
	\tikzset{mark size={1.0}}
	\begin{axis}[
		colormap access=direct,
		width = 1.2\linewidth,
		height = 0.21 * \textheight,
		xmin=0,
		xmax=25,
		xtick = {0,10,...,50},
		scaled x ticks = false,
		xlabel near ticks, xticklabel pos=lower,
    x label style={at={(0.5,-0.09)}},
		ymin = 4750,
		ymax = 6000,
		ytick = {4750,5000,...,6000},
		ylabel near ticks, yticklabel pos=left,
		grid = both,
    legend entries={
      \fontsize{5}{0}\selectfont\sffamily $\phi$ (mean),
      \fontsize{5}{0}\selectfont\sffamily $\phi$ (lowest)
    },
		legend style={
			at={(0.94,0.94)},
			legend columns=1,
			row sep=-2pt,
		},
		legend cell align=left,
    after end axis/.code={
          \draw [solid] (axis cs: 0,4876.98093477) to (axis cs: 25,4876.98093477);
          \draw [] (axis cs:25,4876.98093477) node[inner sep=0.5pt, anchor=south east]{\fontsize{6}{0}\selectfont\sffamily $\phi_{\textnormal{GT}}$};
    }
  	]
	\addlegendimage{solid}		
	\addlegendimage{dash pattern=on 0.175cm off 0.05cm on 0.05cm off 0.05cm on 0.175cm}
	\addplot+ [thin, solid, mark=None, color=red!75!black] table[x=n, y=phi] {./figs/data2019/BIRCH_1_25_results.txt};
	\addplot+ [thin, solid, mark=None, color=cyan!75!black] table[x=n, y=phi] {./figs/data2019/BIRCH_2_25_results.txt};
	\addplot+ [thin, solid, mark=None, color=green!75!black] table[x=n, y=phi] {./figs/data2019/BIRCH_5_25_results.txt};
	\addplot+ [thin, solid, mark=None, color=grey!75!black] table[x=n, y=phi] {./figs/data2019/BIRCH_25_25_results.txt};
	\addplot+ [thin, dash pattern=on 0.35cm off 0.05cm on 0.05cm off 0.05cm, mark=None, color=red!75!black] table[x=n, y=phi] {./figs/data2019/BIRCH_1_25_best_qe.txt};  
	\addplot+ [thin, dash pattern=on 0.35cm off 0.05cm on 0.05cm off 0.05cm, mark=None, color=cyan!75!black] table[x=n, y=phi] {./figs/data2019/BIRCH_2_25_best_qe.txt};  
	\addplot+ [thin, dash pattern=on 0.35cm off 0.05cm on 0.05cm off 0.05cm, mark=None, color=green!75!black] table[x=n, y=phi] {./figs/data2019/BIRCH_5_25_best_qe.txt};  
	\addplot+ [thin, dash pattern=on 0.35cm off 0.05cm on 0.05cm off 0.05cm, mark=None, color=grey!75!black] table[x=n, y=phi] {./figs/data2019/BIRCH_25_25_best_qe.txt};
  \end{axis}
\end{tikzpicture}
    \end{adjustbox}
  \end{subfigure}
	\hspace{5pt}
  \begin{subfigure}[c]{0.22\textwidth}
    \begin{adjustbox}{trim=10pt 0pt 0pt 0pt}
\tikzsetnextfilename{BIRCH_k-means-Cprime_purity_supp}
\pgfplotsset{
	grid style={dotted,gray},
	minor grid style={dotted,lightgray},
  tick label style = {font=\tiny\sansmath\sffamily},
  legend style = {font=\sansmath\sffamily},
  xlabel style = {font=\sansmath\sffamily},
  ylabel style = {font=\sansmath\sffamily},
  legend image code/.code={
    \draw[mark repeat=2,mark phase=2]
    plot coordinates {
      (0cm,0cm)
      (0.25cm,0cm)        
      (0.5cm,0cm)         
    };%
  }
}	
\begin{tikzpicture}
	\tikzset{mark size={1.0}}
	\begin{axis}[
		colormap access=direct,
		width = 1.2\linewidth,
		height = 0.21 * \textheight,
		xmin=0,
		xmax=25,
		xtick = {0,10,...,50},
		scaled x ticks = false,
		xlabel near ticks, xticklabel pos=lower,
    x label style={at={(0.5,-0.09)}},
		ymin = 0.93,
		ymax = 1.00005,
		ylabel near ticks, yticklabel pos=left,
		grid = both,
    legend entries={
      \fontsize{5}{0}\selectfont\sffamily $\mathrm{purity}$ (mean),
      \fontsize{5}{0}\selectfont\sffamily $\mathrm{purity}$ (highest comb.)
    },
		legend style={
			at={(0.94,0.26)},
			legend columns=1,
			row sep=-2pt,
		},
		legend cell align=left,
    after end axis/.code={
          \draw [solid] (axis cs: 0,0.9928) to (axis cs: 25,0.9928);
          \draw [] (axis cs:25,0.9928) node[inner sep=0.5pt, anchor=south east]{\fontsize{5}{0}\selectfont\sffamily $\mathrm{purity}_{\textnormal{GT}}$};
    }
  	]
	\addlegendimage{solid}		
	\addlegendimage{dash pattern=on 0.175cm off 0.05cm on 0.05cm off 0.05cm on 0.175cm}		
	\addplot+ [thin, solid, mark=None, color=red!75!black] table[x=n, y=purity] {./figs/data2019/BIRCH_1_25_results.txt};
	\addplot+ [thin, solid, mark=None, color=cyan!75!black] table[x=n, y=purity] {./figs/data2019/BIRCH_2_25_results.txt};
	\addplot+ [thin, solid, mark=None, color=green!75!black] table[x=n, y=purity] {./figs/data2019/BIRCH_5_25_results.txt};
	\addplot+ [thin, solid, mark=None, color=grey!75!black] table[x=n, y=purity] {./figs/data2019/BIRCH_25_25_results.txt};
	\addplot+ [thin, dash pattern=on 0.35cm off 0.05cm on 0.05cm off 0.05cm, mark=None, color=red!75!black] table[x=n, y=purity] {./figs/data2019/BIRCH_1_25_best_p_NMI.txt};  
	\addplot+ [thin, dash pattern=on 0.35cm off 0.05cm on 0.05cm off 0.05cm, mark=None, color=cyan!75!black] table[x=n, y=purity] {./figs/data2019/BIRCH_2_25_best_p_NMI.txt};  
	\addplot+ [thin, dash pattern=on 0.35cm off 0.05cm on 0.05cm off 0.05cm, mark=None, color=green!75!black] table[x=n, y=purity] {./figs/data2019/BIRCH_5_25_best_p_NMI.txt};  
	\addplot+ [thin, dash pattern=on 0.35cm off 0.05cm on 0.05cm off 0.05cm, mark=None, color=grey!75!black] table[x=n, y=purity] {./figs/data2019/BIRCH_25_25_best_p_NMI.txt};
  \end{axis}
\end{tikzpicture}
    \end{adjustbox}
  \end{subfigure}
	\hspace{5pt}
  \begin{subfigure}[c]{0.22\textwidth}
    \begin{adjustbox}{trim=10pt 0pt 0pt 0pt}
\tikzsetnextfilename{BIRCH_k-means-Cprime_NMI_supp}
\pgfplotsset{
	grid style={dotted,gray},
	minor grid style={dotted,lightgray},
  tick label style = {font=\tiny\sansmath\sffamily},
  xlabel style = {font=\sansmath\sffamily},
  ylabel style = {font=\sansmath\sffamily},
  legend image code/.code={
    \draw[mark repeat=2,mark phase=2]
    plot coordinates {
      (0cm,0cm)
      (0.25cm,0cm)        
      (0.5cm,0cm)         
    };%
  }
}	
\begin{tikzpicture}
	\tikzset{mark size={1.0}}
	\begin{axis}[
		colormap access=direct,
		width = 1.2\linewidth,
		height = 0.21 * \textheight,
		xmin=0,
		xmax=25,
		xtick = {0,10,...,50},
		scaled x ticks = false,
		xlabel near ticks, xticklabel pos=lower,
    x label style={at={(0.5,-0.09)}},
		ymin = 0.95,
		ymax = 1.00005,
		ytick = {0.95,0.96,...,1.01},
		ylabel near ticks, yticklabel pos=left,
		grid = both,
    legend entries={
      \fontsize{5}{0}\selectfont\sffamily $\mathrm{NMI}$ (mean),
      \fontsize{5}{0}\selectfont\sffamily $\mathrm{NMI}$ (highest comb.)
    },
		legend style={
			at={(0.94,0.26)},
			legend columns=1,
			row sep=-2pt,
		},
		legend cell align=left,
    after end axis/.code={
          \draw [solid] (axis cs: 0,0.987646436635) to (axis cs: 25,0.987646436635);
          \draw [] (axis cs:25,0.987646436635) node[inner sep=0.5pt, anchor=south east]{\fontsize{5}{0}\selectfont\sffamily $\mathrm{NMI}_{\textnormal{GT}}$};
    }
  	]
	\addlegendimage{solid}		
	\addlegendimage{dash pattern=on 0.175cm off 0.05cm on 0.05cm off 0.05cm on 0.175cm}				
	\addplot+ [thin, solid, mark=None, color=red!75!black] table[x=n, y=NMI] {./figs/data2019/BIRCH_1_25_results.txt};
	\addplot+ [thin, solid, mark=None, color=cyan!75!black] table[x=n, y=NMI] {./figs/data2019/BIRCH_2_25_results.txt};
	\addplot+ [thin, solid, mark=None, color=green!75!black] table[x=n, y=NMI] {./figs/data2019/BIRCH_5_25_results.txt};
	\addplot+ [thin, solid, mark=None, color=grey!75!black] table[x=n, y=NMI] {./figs/data2019/BIRCH_25_25_results.txt};
	\addplot+ [thin, dash pattern=on 0.35cm off 0.05cm on 0.05cm off 0.05cm, mark=None, color=red!75!black] table[x=n, y=NMI] {./figs/data2019/BIRCH_1_25_best_p_NMI.txt};  
	\addplot+ [thin, dash pattern=on 0.35cm off 0.05cm on 0.05cm off 0.05cm, mark=None, color=cyan!75!black] table[x=n, y=NMI] {./figs/data2019/BIRCH_2_25_best_p_NMI.txt};  
	\addplot+ [thin, dash pattern=on 0.35cm off 0.05cm on 0.05cm off 0.05cm, mark=None, color=green!75!black] table[x=n, y=NMI] {./figs/data2019/BIRCH_5_25_best_p_NMI.txt};  
	\addplot+ [thin, dash pattern=on 0.35cm off 0.05cm on 0.05cm off 0.05cm, mark=None, color=grey!75!black] table[x=n, y=NMI] {./figs/data2019/BIRCH_25_25_best_p_NMI.txt};
  \end{axis}
\end{tikzpicture}
    \end{adjustbox}
  \end{subfigure}
	\vspace{4pt}
	\newline
	\begin{picture}(0,0)
		\put(-0,-70){\rotatebox{90}{\rlap{\makebox[5.80cm]{\fontsize{8}{0}\selectfont\sffamily BIRCH ($25\times$ random pos.)}}}}
	\end{picture}
	\hspace{18pt}
  \begin{subfigure}[c]{0.22\textwidth}
    \begin{adjustbox}{trim=10pt 5.5pt 0pt 0pt}
\tikzsetnextfilename{BIRCH_random-1-25-100_k-means-Cprime_supp}
\pgfplotsset{
	grid style={dotted,gray},
	minor grid style={dotted,lightgray},
  tick label style = {font=\tiny\sansmath\sffamily},
  legend style = {font=\sansmath\sffamily},
  xlabel style = {font=\sansmath\sffamily},
  ylabel style = {font=\sansmath\sffamily},
  legend image code/.code={
    \draw[mark repeat=2,mark phase=2]
    plot coordinates {
      (0cm,0cm)
      (0.25cm,0cm)        
      (0.5cm,0cm)         
    };%
  }
}	
\begin{tikzpicture}
	\tikzset{mark size={1.0}}
	\begin{axis}[
		colormap access=direct,
		width = 1.2\linewidth,
		height = 0.21 * \textheight,
		xmin=0,
		xmax=25,
		scaled x ticks = false,
		xlabel={\scriptsize\sffamily Iteration},
		xlabel near ticks, xticklabel pos=lower,
    x label style={at={(0.5,-0.09)}},
		ymin = -6.0,
		ymax = -5.75,
		ylabel near ticks, yticklabel pos=left,
		grid = both,
    legend entries={
      \fontsize{5}{0}\selectfont\sffamily $\mathcal{L}$ (mean),
      \fontsize{5}{0}\selectfont\sffamily $\mathcal{L}$ (highest)
    },
		legend style={
			at={(0.94,0.26)},
			legend columns=1,
			row sep=-2pt,
		},
		legend cell align=left,
    after end axis/.code={
          \draw [solid] (axis cs: 0,-5.77838223692) to (axis cs: 25,-5.77838223692);
          \draw [-latex, shorten <=-3pt] (axis cs: -1,-5.77838223692) node[left]{\fontsize{6}{0}\selectfont\sffamily $\mathcal{L}_{\textnormal{GT}}$} to[out=360,in=180] (axis cs: 0,-5.77838223692);
    }
  	]
	\addlegendimage{solid}		
	\addlegendimage{dash pattern=on 0.175cm off 0.05cm on 0.05cm off 0.05cm on 0.175cm}
	\addplot+ [thin, solid, mark=None, color=red!75!black] table[x=n, y=loglikelihood] {./figs/BIRCH_random-1-25-100_k-means_25.txt};
	\addplot+ [thin, solid, mark=None, color=cyan!75!black] table[x=n, y=loglikelihood] {./figs/BIRCH_random-1-25-100_k-means-C'_25_2.txt};
	\addplot+ [thin, solid, mark=None, color=green!75!black] table[x=n, y=loglikelihood] {./figs/BIRCH_random-1-25-100_k-means-C'_25_5.txt};
	\addplot+ [thin, solid, mark=None, color=grey!75!black] table[x=n, y=loglikelihood] {./figs/BIRCH_random-1-25-100_k-means-C'_25_25.txt};
	\addplot+ [thin, dash pattern=on 0.35cm off 0.05cm on 0.05cm off 0.05cm, mark=None, color=red!75!black] table[x=n, y=loglikelihood] {./figs/BIRCH_random-1-25-100_k-means_25-best.txt};  
	\addplot+ [thin, dash pattern=on 0.35cm off 0.05cm on 0.05cm off 0.05cm, mark=None, color=cyan!75!black] table[x=n, y=loglikelihood] {./figs/BIRCH_random-1-25-100_k-means-C'_25_2-best.txt};  
	\addplot+ [thin, dash pattern=on 0.35cm off 0.05cm on 0.05cm off 0.05cm, mark=None, color=green!75!black] table[x=n, y=loglikelihood] {./figs/BIRCH_random-1-25-100_k-means-C'_25_5-best.txt};  
	\addplot+ [thin, dash pattern=on 0.35cm off 0.05cm on 0.05cm off 0.05cm, mark=None, color=grey!75!black] table[x=n, y=loglikelihood] {./figs/BIRCH_random-1-25-100_k-means-C'_25_25-best.txt};
	\draw [-latex, black!70!black] (axis cs: 22,-5.76) node[left]{\fontsize{5}{0}\selectfont\sffamily all clusters recovered} to (axis cs: 24,-5.7695);	
	\draw [-latex, black!70!black] (axis cs: 22,-5.76)  to (axis cs: 24,-5.78836487758);	
  \end{axis}
\end{tikzpicture}
    \end{adjustbox}
  \end{subfigure}
	\hspace{5pt}
  \begin{subfigure}[c]{0.22\textwidth}
    \begin{adjustbox}{trim=10pt 0pt 0pt 0pt}
\tikzsetnextfilename{BIRCH-random_k-means-Cprime_QE_supp}
\pgfplotsset{
	grid style={dotted,gray},
	minor grid style={dotted,lightgray},
  tick label style = {font=\tiny\sansmath\sffamily},
  xlabel style = {font=\sansmath\sffamily},
  ylabel style = {font=\sansmath\sffamily},
  legend image code/.code={
    \draw[mark repeat=2,mark phase=2]
    plot coordinates {
      (0cm,0cm)
      (0.25cm,0cm)        
      (0.5cm,0cm)         
    };%
  }
}	
\begin{tikzpicture}
	\tikzset{mark size={1.0}}
	\begin{axis}[
		colormap access=direct,
		width = 1.2\linewidth,
		height = 0.21 * \textheight,
		xmin=0,
		xmax=25,
		xtick = {0,10,...,50},
		scaled x ticks = false,
		xlabel={\scriptsize\sffamily Iteration},
		xlabel near ticks, xticklabel pos=lower,
    x label style={at={(0.5,-0.09)}},
		ymin = 4250,
		ymax = 5000,
		ytick = {4250,4500,...,5000},
		ylabel near ticks, yticklabel pos=left,
		grid = both,
    legend entries={
      \fontsize{5}{0}\selectfont\sffamily $\phi$ (mean),
      \fontsize{5}{0}\selectfont\sffamily $\phi$ (lowest)
    },
		legend style={
			at={(0.94,0.94)},
			legend columns=1,
			row sep=-2pt,
		},
		legend cell align=left,
    after end axis/.code={
          \draw [solid] (axis cs: 0,4379.48933854) to (axis cs: 25,4379.489338547);
          \draw [] (axis cs:25,4379.48933854) node[inner sep=0.5pt, anchor=south east]{\fontsize{6}{0}\selectfont\sffamily $\phi_{\textnormal{GT}}$};
    }
  	]
	\addlegendimage{solid}		
	\addlegendimage{dash pattern=on 0.175cm off 0.05cm on 0.05cm off 0.05cm on 0.175cm}
	\addplot+ [thin, solid, mark=None, color=red!75!black] table[x=n, y=phi] {./figs/data2019/BIRCH_random_1_25_results.txt};
	\addplot+ [thin, solid, mark=None, color=cyan!75!black] table[x=n, y=phi] {./figs/data2019/BIRCH_random_2_25_results.txt};
	\addplot+ [thin, solid, mark=None, color=green!75!black] table[x=n, y=phi] {./figs/data2019/BIRCH_random_5_25_results.txt};
	\addplot+ [thin, solid, mark=None, color=grey!75!black] table[x=n, y=phi] {./figs/data2019/BIRCH_random_25_25_results.txt};
	\addplot+ [thin, dash pattern=on 0.35cm off 0.05cm on 0.05cm off 0.05cm, mark=None, color=red!75!black] table[x=n, y=phi] {./figs/data2019/BIRCH_random_1_25_best_qe.txt};  
	\addplot+ [thin, dash pattern=on 0.35cm off 0.05cm on 0.05cm off 0.05cm, mark=None, color=cyan!75!black] table[x=n, y=phi] {./figs/data2019/BIRCH_random_2_25_best_qe.txt};  
	\addplot+ [thin, dash pattern=on 0.35cm off 0.05cm on 0.05cm off 0.05cm, mark=None, color=green!75!black] table[x=n, y=phi] {./figs/data2019/BIRCH_random_5_25_best_qe.txt};  
	\addplot+ [thin, dash pattern=on 0.35cm off 0.05cm on 0.05cm off 0.05cm, mark=None, color=grey!75!black] table[x=n, y=phi] {./figs/data2019/BIRCH_random_25_25_best_qe.txt};
  \end{axis}
\end{tikzpicture}
    \end{adjustbox}
  \end{subfigure}
	\hspace{5pt}
  \begin{subfigure}[c]{0.22\textwidth}
    \begin{adjustbox}{trim=10pt 5pt 0pt 0pt}
\tikzsetnextfilename{BIRCH-random_k-means-Cprime_purity_supp}
\pgfplotsset{
	grid style={dotted,gray},
	minor grid style={dotted,lightgray},
  tick label style = {font=\tiny\sansmath\sffamily},
  legend style = {font=\sansmath\sffamily},
  xlabel style = {font=\sansmath\sffamily},
  ylabel style = {font=\sansmath\sffamily},
  legend image code/.code={
    \draw[mark repeat=2,mark phase=2]
    plot coordinates {
      (0cm,0cm)
      (0.25cm,0cm)        
      (0.5cm,0cm)         
    };%
  }
}	
\begin{tikzpicture}
	\tikzset{mark size={1.0}}
	\begin{axis}[
		colormap access=direct,
		width = 1.2\linewidth,
		height = 0.21 * \textheight,
		xmin=0,
		xmax=25,
		xtick = {0,10,...,50},
		scaled x ticks = false,
		xlabel={\scriptsize\sffamily Iteration},
		xlabel near ticks, xticklabel pos=lower,
    x label style={at={(0.5,-0.09)}},
		ymin = 0.725,
		ymax = 0.875,
		ylabel near ticks, yticklabel pos=left,
		grid = both,
    legend entries={
      \fontsize{5}{0}\selectfont\sffamily $\mathrm{purity}$ (mean),
      \fontsize{5}{0}\selectfont\sffamily $\mathrm{purity}$ (highest  comb.)
    },
		legend style={
			at={(0.94,0.26)},
			legend columns=1,
			row sep=-2pt,
		},
		legend cell align=left,
    after end axis/.code={
          \draw [solid] (axis cs: 0,0.8648) to (axis cs: 25,0.8648);
          \draw [] (axis cs:25,0.8648) node[inner sep=0.5pt, anchor=south east]{\fontsize{5}{0}\selectfont\sffamily $\mathrm{purity}_{\textnormal{GT}}$};
    }
  	]
	\addlegendimage{solid}		
	\addlegendimage{dash pattern=on 0.175cm off 0.05cm on 0.05cm off 0.05cm on 0.175cm}		
	\addplot+ [thin, solid, mark=None, color=red!75!black] table[x=n, y=purity] {./figs/data2019/BIRCH_random_1_25_results.txt};
	\addplot+ [thin, solid, mark=None, color=cyan!75!black] table[x=n, y=purity] {./figs/data2019/BIRCH_random_2_25_results.txt};
	\addplot+ [thin, solid, mark=None, color=green!75!black] table[x=n, y=purity] {./figs/data2019/BIRCH_random_5_25_results.txt};
	\addplot+ [thin, solid, mark=None, color=grey!75!black] table[x=n, y=purity] {./figs/data2019/BIRCH_random_25_25_results.txt};
	\addplot+ [thin, dash pattern=on 0.35cm off 0.05cm on 0.05cm off 0.05cm, mark=None, color=red!75!black] table[x=n, y=purity] {./figs/data2019/BIRCH_random_1_25_best_p_NMI.txt};  
	\addplot+ [thin, dash pattern=on 0.35cm off 0.05cm on 0.05cm off 0.05cm, mark=None, color=cyan!75!black] table[x=n, y=purity] {./figs/data2019/BIRCH_random_2_25_best_p_NMI.txt};  
	\addplot+ [thin, dash pattern=on 0.35cm off 0.05cm on 0.05cm off 0.05cm, mark=None, color=green!75!black] table[x=n, y=purity] {./figs/data2019/BIRCH_random_5_25_best_p_NMI.txt};  
	\addplot+ [thin, dash pattern=on 0.35cm off 0.05cm on 0.05cm off 0.05cm, mark=None, color=grey!75!black] table[x=n, y=purity] {./figs/data2019/BIRCH_random_25_25_best_p_NMI.txt};
  \end{axis}
\end{tikzpicture}
    \end{adjustbox}
  \end{subfigure}
	\hspace{5pt}
  \begin{subfigure}[c]{0.22\textwidth}
    \begin{adjustbox}{trim=12.5pt 0pt 0pt 0pt}
\tikzsetnextfilename{BIRCH-random_k-means-Cprime_NMI_supp}
\pgfplotsset{
	grid style={dotted,gray},
	minor grid style={dotted,lightgray},
  tick label style = {font=\tiny\sansmath\sffamily},
  xlabel style = {font=\sansmath\sffamily},
  ylabel style = {font=\sansmath\sffamily},
  legend image code/.code={
    \draw[mark repeat=2,mark phase=2]
    plot coordinates {
      (0cm,0cm)
      (0.25cm,0cm)        
      (0.5cm,0cm)         
    };%
  }
}	
\begin{tikzpicture}
	\tikzset{mark size={1.0}}
	\begin{axis}[
		colormap access=direct,
		width = 1.2\linewidth,
		height = 0.21 * \textheight,
		xmin=0,
		xmax=25,
		xtick = {0,10,...,50},
		scaled x ticks = false,
		xlabel={\scriptsize\sffamily Iteration},
		xlabel near ticks, xticklabel pos=lower,
    x label style={at={(0.5,-0.09)}},
		ymin = 0.865,
		ymax = 0.885,
		/pgf/number format/precision=3,
		ytick = {0.865,0.87,...,0.895},
		ylabel near ticks, yticklabel pos=left,
		grid = both,
    legend entries={
      \fontsize{5}{0}\selectfont\sffamily $\mathrm{NMI}$ (mean),
      \fontsize{5}{0}\selectfont\sffamily $\mathrm{NMI}$ (highest  comb.)
    },
		legend style={
			at={(0.94,0.26)},
			legend columns=1,
			row sep=-2pt,
		},
		legend cell align=left,
    after end axis/.code={
          \draw [solid] (axis cs: 0,0.87913094305) to (axis cs: 25,0.87913094305);
          \draw [] (axis cs:25,0.87913094305) node[inner sep=0.5pt, anchor=south east]{\fontsize{5}{0}\selectfont\sffamily $\mathrm{NMI}_{\textnormal{GT}}$};
    }
  	]
	\addlegendimage{solid}		
	\addlegendimage{dash pattern=on 0.175cm off 0.05cm on 0.05cm off 0.05cm on 0.175cm}				
	\addplot+ [thin, solid, mark=None, color=red!75!black] table[x=n, y=NMI] {./figs/data2019/BIRCH_random_1_25_results.txt};
	\addplot+ [thin, solid, mark=None, color=cyan!75!black] table[x=n, y=NMI] {./figs/data2019/BIRCH_random_2_25_results.txt};
	\addplot+ [thin, solid, mark=None, color=green!75!black] table[x=n, y=NMI] {./figs/data2019/BIRCH_random_5_25_results.txt};
	\addplot+ [thin, solid, mark=None, color=grey!75!black] table[x=n, y=NMI] {./figs/data2019/BIRCH_random_25_25_results.txt};
	\addplot+ [thin, dash pattern=on 0.35cm off 0.05cm on 0.05cm off 0.05cm, mark=None, color=red!75!black] table[x=n, y=NMI] {./figs/data2019/BIRCH_random_1_25_best_p_NMI.txt};  
	\addplot+ [thin, dash pattern=on 0.35cm off 0.05cm on 0.05cm off 0.05cm, mark=None, color=cyan!75!black] table[x=n, y=NMI] {./figs/data2019/BIRCH_random_2_25_best_p_NMI.txt};  
	\addplot+ [thin, dash pattern=on 0.35cm off 0.05cm on 0.05cm off 0.05cm, mark=None, color=green!75!black] table[x=n, y=NMI] {./figs/data2019/BIRCH_random_5_25_best_p_NMI.txt};  
	\addplot+ [thin, dash pattern=on 0.35cm off 0.05cm on 0.05cm off 0.05cm, mark=None, color=grey!75!black] table[x=n, y=NMI] {./figs/data2019/BIRCH_random_25_25_best_p_NMI.txt};
  \end{axis}
\end{tikzpicture}
    \end{adjustbox}
  \end{subfigure}
	\vspace{-5pt}
  \caption{
    \changed{The four columns show from left to right: the log-likelihood $\mathcal{L}$, quantization error $\phi$, purity, and normalized mutual information (NMI) of the $k$-means-C' algorithms on BIRCH data sets as in Fig.\,\ref{ExpA}.
		The plots show the mean over 100~independent runs (solid line). For the log-likelihood and quantization error, the single run with the best respective final value is shown in striped. For the purity and NMI plots, the single run with the highest sum of purity and NMI is shown in striped. Such a selection criterion omits runs that are highly overfitted to either purity or NMI.}
    }
  \vspace{-10pt}    
	\label{fig:addExpA}
\end{figure*}
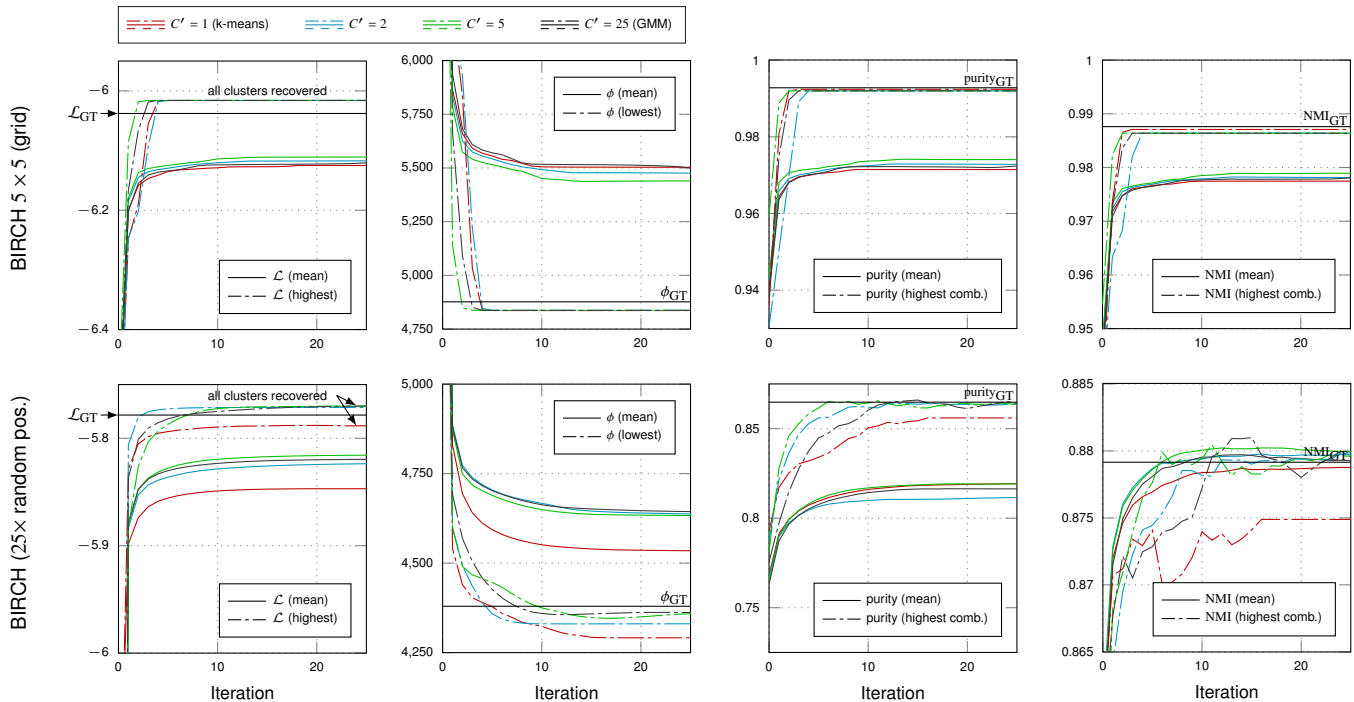

\changed{Considering (\ref{EqnGenCriterion}), note that the criterion to select clusters now depends on all model parameters (in contrast to the criterion of Eq.\,\ref{EqnEuclid}).
If algorithms for parameter updates are defined based on (\ref{EqnGenCriterion}), all current parameter values have to be considered in E-steps which compute 
generalizations of the responsibilities $\qcn$ (compare Eq.\,\ref{EqnQCN}). Notably, even if these responsibilities $\qcn$ become binary for the choice $C'=1$, the selection
of the non-zero values of $\qcn$ would still require the other parameter values. There would consequently not be a $k$-means-like decoupling from other parameter updates
like for the GMM defined by Eq.\,(\ref{EqnGMMIso}).
}

\begin{figure*}[p]
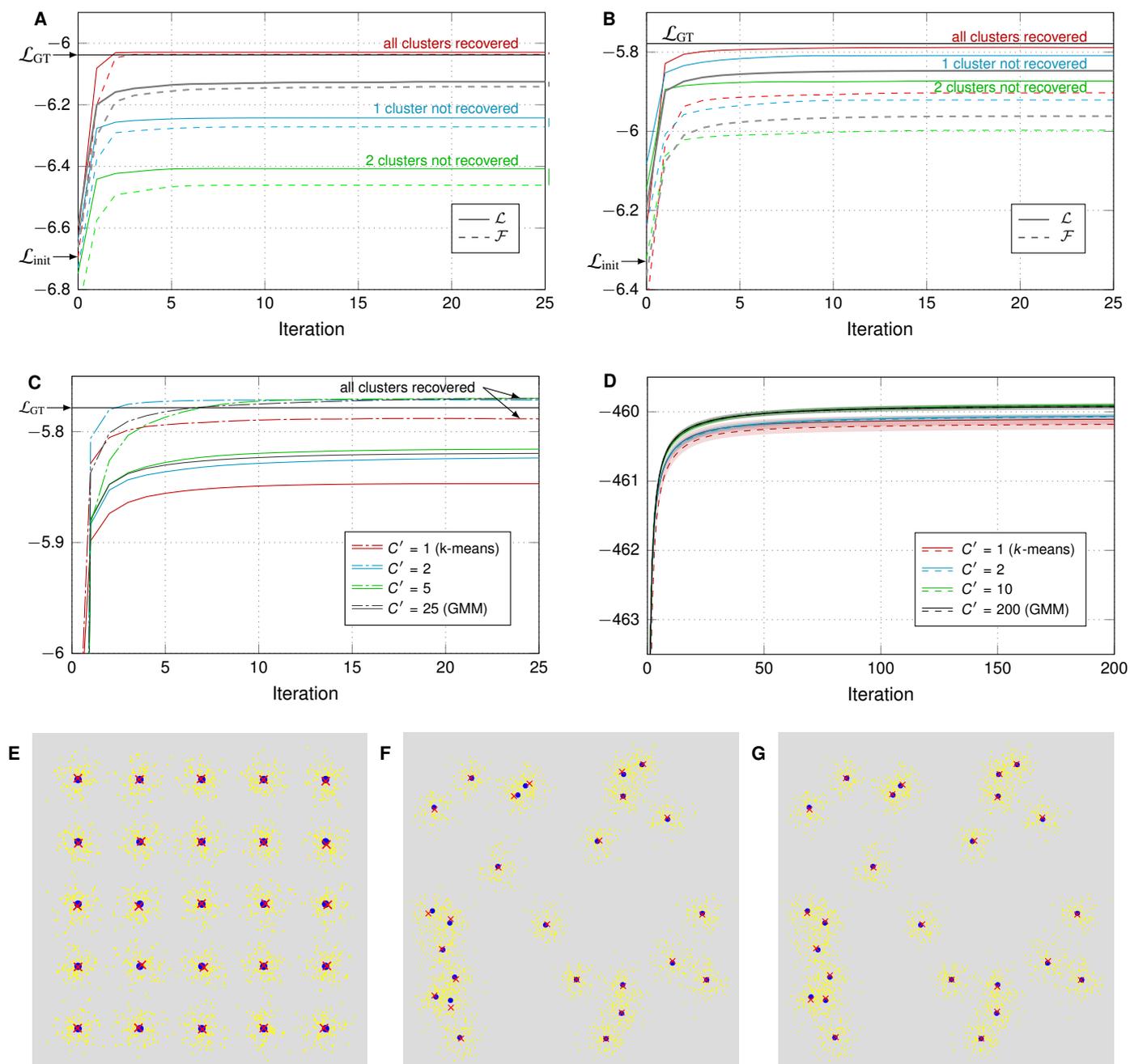

  \begin{subfigure}[c]{0.5\textwidth}
    \begin{adjustbox}{trim=0pt 0pt 0pt 0pt}
\tikzsetnextfilename{BIRCH-1-5-100_k-means_25_large}
\pgfplotsset{
	grid style={dotted,gray},
	minor grid style={dotted,lightgray},
  tick label style = {font=\footnotesize\sansmath\sffamily},
  legend style = {font=\sansmath\sffamily},
  xlabel style = {font=\sansmath\sffamily},
  ylabel style = {font=\sansmath\sffamily},
  legend image code/.code={
    \draw[mark repeat=2,mark phase=2]
    plot coordinates {
      (0cm,0cm)
      (0.25cm,0cm)        
      (0.5cm,0cm)         
    };%
  },
}	
\begin{tikzpicture}
	\tikzset{mark size={1.0}}
	\begin{axis}[
		colormap access=direct,
		width = 1 * \linewidth,
		height = 0.25 * \textheight,
		xmin=0,
		xmax=25,
		scaled x ticks = false,
		xlabel={\small\sffamily Iteration},
		xlabel near ticks, xticklabel pos=lower,
    x label style={at={(0.5,-0.09)}},
		ymin = -6.8,
		ymax = -5.9,
		ylabel near ticks, yticklabel pos=left,
		grid = both,
		%
    legend entries={
      \fontsize{7}{0}\selectfont\sffamily $\mathcal{L}$,
      \fontsize{7}{0}\selectfont\sffamily $\mathcal{F}$
    },
		legend style={
			at={(0.945,0.31)},
			legend columns=1,
			row sep=-2pt,
		},
		legend cell align=left,
    after end axis/.code={
          \draw [-latex, shorten <=-3pt] (axis cs: -1,-6.6933935576) node[left]{\fontsize{9}{0}\selectfont\sffamily $\mathcal{L}_{\textnormal{init}}$} to[out=360,in=180] (axis cs: 0,-6.6933935576);
          \draw [solid] (axis cs: 0,-6.0378687151) to (axis cs: 25,-6.0378687151);
          \draw [-latex, shorten <=-3pt] (axis cs: -1,-6.0378687151) node[left]{\fontsize{9}{0}\selectfont\sffamily $\mathcal{L}_{\textnormal{GT}}$} to[out=360,in=180] (axis cs: 0,-6.0378687151);
          \node at (axis cs: -2,-5.92) {\fontsize{9}{0}\selectfont\sffamily\bfseries A};
					\draw [-, red!70!black] (axis cs: 25.2,-6.03638830146) to (axis cs: 25.2,-6.02933957757);
					\draw [-, cyan!70!black] (axis cs: 25.2,-6.27162484389) to (axis cs: 25.2,-6.24257018759);
					\draw [-, green!70!black] (axis cs: 25.2,-6.46130326354) to (axis cs: 25.2,-6.40734896632);
					\draw [-, grey!70!black] (axis cs: 25.2,-6.14110728288) to (axis cs: 25.2,-6.12502651431);
    }
  	]
	\addlegendimage{solid}
	\addlegendimage{dashed}
	\addplot+ [thin, solid, mark=None, color=red!75!black] table[x=n, y=loglikelihood] {./figs/BIRCH_k-means_LL-0.txt};
	\addplot+ [thin, dashed, mark=None, color=red!90!black] table[x=n, y=free-energy] {./figs/BIRCH_k-means_LL-0.txt};
	\node[red!70!black, anchor=east] at (axis cs: 24,-6.0) {\fontsize{7}{0}\selectfont\sffamily all clusters recovered};
	\addplot+ [thin, solid, mark=None, color=cyan!75!black] table[x=n, y=loglikelihood] {./figs/BIRCH_k-means_LL-1.txt};
	\addplot+ [thin, dashed, mark=None, color=cyan!90!black] table[x=n, y=free-energy] {./figs/BIRCH_k-means_LL-1.txt};	
	\node[cyan!70!black, anchor=east] at (axis cs: 24,-6.21) {\fontsize{7}{0}\selectfont\sffamily 1 cluster not recovered};
	\addplot+ [thin, solid, mark=None, color=green!75!black] table[x=n, y=loglikelihood] {./figs/BIRCH_k-means_LL-2.txt};
	\addplot+ [thin, dashed, mark=None, color=green!90!black] table[x=n, y=free-energy] {./figs/BIRCH_k-means_LL-2.txt};	
  \node[green!70!black, anchor=east] at (axis cs: 24,-6.38) {\fontsize{7}{0}\selectfont\sffamily 2 clusters not recovered};
	\addplot+ [thick, solid, mark=None, color=grey!75!white] table[x=n, y=loglikelihood] {./figs/BIRCH_k-means_LL.txt};
	\addplot+ [thick, dashed, mark=None, color=grey!60!white] table[x=n, y=free-energy] {./figs/BIRCH_k-means_LL.txt};	
  \end{axis}
\end{tikzpicture}
    \end{adjustbox}
  \end{subfigure}
  \vspace{10pt}
  \begin{subfigure}[c]{0.5\textwidth}
    \begin{adjustbox}{trim=0pt 0pt 0pt 0pt}
\tikzsetnextfilename{BIRCH_random-1-25-100_k-means_25_large}
\pgfplotsset{
	grid style={dotted,gray},
	minor grid style={dotted,lightgray},
  tick label style = {font=\footnotesize\sansmath\sffamily},
  legend style = {font=\sansmath\sffamily},
  xlabel style = {font=\sansmath\sffamily},
  ylabel style = {font=\sansmath\sffamily},
  legend image code/.code={
    \draw[mark repeat=2,mark phase=2]
    plot coordinates {
      (0cm,0cm)
      (0.25cm,0cm)        
      (0.5cm,0cm)         
    };%
  }
}	
\begin{tikzpicture}
	\tikzset{mark size={1.0}}
	\begin{axis}[
		colormap access=direct,
		width = 1. *\linewidth,
		height = 0.25 * \textheight,
		xmin=0,
		xmax=25,
		scaled x ticks = false,
		xlabel={\small\sffamily Iteration},
		xlabel near ticks, xticklabel pos=lower,
    x label style={at={(0.5,-0.09)}},
		ymin = -6.4,
		ymax = -5.7,
		ylabel near ticks, yticklabel pos=left,
		grid = both,
		%
    legend entries={
      \fontsize{7}{0}\selectfont\sffamily \, $\mathcal{L}$,
      \fontsize{7}{0}\selectfont\sffamily \, $\mathcal{F}$
    },
		legend style={
			at={(0.94,0.31)},
			legend columns=1,
			row sep=-2pt,
		},
		legend cell align=left,
    after end axis/.code={
          \draw [-latex, shorten <=-3pt] (axis cs: -1,-6.3283711253) node[left]{\fontsize{9}{0}\selectfont\sffamily $\mathcal{L}_{\textnormal{init}}$} to[out=360,in=180] (axis cs: 0,-6.3283711253);
          \draw [solid] (axis cs: 0,-5.77838223692) to (axis cs: 25,-5.77838223692);
          \draw [] (axis cs:3,-5.75338223692) node[left]{\fontsize{9}{0}\selectfont\sffamily $\mathcal{L}_{\textnormal{GT}}$};
          \node at (axis cs: -2,-5.715) {\fontsize{9}{0}\selectfont\sffamily\bfseries B};
    }
  	]
	\addlegendimage{solid}
	\addlegendimage{dashed}
	\addplot+ [thin, solid, mark=None, color=red!75!black] table[x=n, y=loglikelihood] {./figs/BIRCH_random-1-25-100_k-means_25-best.txt};
	\addplot+ [thin, dashed, mark=None, color=red!90!black] table[x=n, y=free-energy] {./figs/BIRCH_random-1-25-100_k-means_25-best.txt};
  \node[red!70!black, anchor=east] at (axis cs: 24,-5.76) {\fontsize{7}{0}\selectfont\sffamily all clusters recovered};
	\addplot+ [thin, solid, mark=None, color=cyan!75!black] table[x=n, y=loglikelihood] {./figs/BIRCH_random-1-25-100_k-means_25-1.txt};
	\addplot+ [thin, dashed, mark=None, color=cyan!90!black] table[x=n, y=free-energy] {./figs/BIRCH_random-1-25-100_k-means_25-1.txt};	
  \node[cyan!70!black, anchor=east] at(axis cs: 24,-5.828) {\fontsize{7}{0}\selectfont\sffamily 1 cluster not recovered};
	\addplot+ [thin, solid, mark=None, color=green!75!black] table[x=n, y=loglikelihood] {./figs/BIRCH_random-1-25-100_k-means_25-4.txt};
	\addplot+ [thin, dashed, mark=None, color=green!90!black] table[x=n, y=free-energy] {./figs/BIRCH_random-1-25-100_k-means_25-4.txt};	
	\node[green!70!black, anchor=east] at (axis cs: 24,-5.888) {\fontsize{7}{0}\selectfont\sffamily 2 clusters not recovered};
	\addplot+ [thick, solid, mark=None, color=grey!75!white] table[x=n, y=loglikelihood] {./figs/BIRCH_random-1-25-100_k-means_25.txt};
	\addplot+ [thick, dashed, mark=None, color=grey!60!white] table[x=n, y=free-energy] {./figs/BIRCH_random-1-25-100_k-means_25.txt};	
  \end{axis}
\end{tikzpicture}
    \end{adjustbox}
  \end{subfigure}
  \begin{subfigure}[c]{0.5\textwidth}
    \begin{adjustbox}{trim=0pt 0pt 0pt 0pt}
\tikzsetnextfilename{BIRCH_random-1-25-100_k-means-Cprime_large}
\pgfplotsset{
	grid style={dotted,gray},
	minor grid style={dotted,lightgray},
  tick label style = {font=\footnotesize\sansmath\sffamily},
  legend style = {font=\sansmath\sffamily},
  xlabel style = {font=\sansmath\sffamily},
  ylabel style = {font=\sansmath\sffamily},
  legend image code/.code={
    \draw[mark repeat=2,mark phase=2]
    plot coordinates {
      (0cm,0cm)
      (0.25cm,0cm)        
      (0.5cm,0cm)         
    };%
  }
}	
\begin{tikzpicture}
	\tikzset{mark size={1.0}}
	\begin{axis}[
		colormap access=direct,
		width = 1. *\linewidth,
		height = 0.25 * \textheight,
		xmin=0,
		xmax=25,
		scaled x ticks = false,
		xlabel={\small\sffamily Iteration},
		xlabel near ticks, xticklabel pos=lower,
    x label style={at={(0.5,-0.09)}},
		ymin = -6.0,
		ymax = -5.75,
		ylabel near ticks, yticklabel pos=left,
		grid = both,
   legend image code/.code={%
     \draw[dash pattern=on 0.175cm off 0.05cm on 0.05cm off 0.05cm on 0.175cm] (0cm,0.05cm) -- (0.5cm,0.05cm);
     \draw[solid]  (0cm, 0.0cm) -- (0.5cm, 0.0cm);
    },		
    legend entries={
      \fontsize{7}{0}\selectfont\sffamily $C' = 1$ (k-means),
      \fontsize{7}{0}\selectfont\sffamily $C' = 2$,
      \fontsize{7}{0}\selectfont\sffamily $C' = 5$,
      \fontsize{7}{0}\selectfont\sffamily $C' = 25$ (GMM),
    },
		legend style={
			at={(0.94,0.44)},
			legend columns=1,
			row sep=-2pt,
		},
		legend cell align=left,
    after end axis/.code={
          \draw [solid] (axis cs: 0,-5.77838223692) to (axis cs: 25,-5.77838223692);
          \draw [-latex, shorten <=-3pt] (axis cs: -1,-5.77838223692) node[left]{\fontsize{7}{0}\selectfont\sffamily $\mathcal{L}_{\textnormal{GT}}$} to[out=360,in=180] (axis cs: 0,-5.77838223692);
          \node at (axis cs: -2,-5.755) {\fontsize{9}{0}\selectfont\sffamily\bfseries C};
    }
  	]
	\addplot+ [thin, solid, mark=None, color=red!75!black] table[x=n, y=loglikelihood] {./figs/BIRCH_random-1-25-100_k-means_25.txt};
	\addplot+ [thin, solid, mark=None, color=cyan!75!black] table[x=n, y=loglikelihood] {./figs/BIRCH_random-1-25-100_k-means-C'_25_2.txt};
	\addplot+ [thin, solid, mark=None, color=green!75!black] table[x=n, y=loglikelihood] {./figs/BIRCH_random-1-25-100_k-means-C'_25_5.txt};
	\addplot+ [thin, solid, mark=None, color=grey!75!black] table[x=n, y=loglikelihood] {./figs/BIRCH_random-1-25-100_k-means-C'_25_25.txt};
	\addplot+ [thin, dash pattern=on 0.35cm off 0.05cm on 0.05cm off 0.05cm, mark=None, color=red!75!black] table[x=n, y=loglikelihood] {./figs/BIRCH_random-1-25-100_k-means_25-best.txt};  
	\addplot+ [thin, dash pattern=on 0.35cm off 0.05cm on 0.05cm off 0.05cm, mark=None, color=cyan!75!black] table[x=n, y=loglikelihood] {./figs/BIRCH_random-1-25-100_k-means-C'_25_2-best.txt};  
	\addplot+ [thin, dash pattern=on 0.35cm off 0.05cm on 0.05cm off 0.05cm, mark=None, color=green!75!black] table[x=n, y=loglikelihood] {./figs/BIRCH_random-1-25-100_k-means-C'_25_5-best.txt};  
	\addplot+ [thin, dash pattern=on 0.35cm off 0.05cm on 0.05cm off 0.05cm, mark=None, color=grey!75!black] table[x=n, y=loglikelihood] {./figs/BIRCH_random-1-25-100_k-means-C'_25_25-best.txt};
	\draw [-latex, black!70!black] (axis cs: 22,-5.76) node[left]{\fontsize{7}{0}\selectfont\sffamily all clusters recovered} to (axis cs: 24,-5.7695);	
	\draw [-latex, black!70!black] (axis cs: 22,-5.76)  to (axis cs: 24,-5.78836487758);	
  \end{axis}
\end{tikzpicture}
    \end{adjustbox}
  \end{subfigure}
  \vspace{10pt}
  \begin{subfigure}[c]{0.5\textwidth}
    \begin{adjustbox}{trim=10pt 0pt 0pt 0pt}
\tikzsetnextfilename{KDD_k-means_Cprime_large}
\pgfplotsset{
	grid style={dotted,gray},
	minor grid style={dotted,lightgray},
  tick label style = {font=\footnotesize\sansmath\sffamily},
  legend style = {font=\sansmath\sffamily},
  xlabel style = {font=\sansmath\sffamily},
  ylabel style = {font=\sansmath\sffamily},
  legend image code/.code={
    \draw[mark repeat=2,mark phase=2]
    plot coordinates {
      (0cm,0cm)
      (0.25cm,0cm)        
      (0.5cm,0cm)         
    };%
  }
}

\begin{tikzpicture}
	\tikzset{mark size={1.0}}
	\begin{axis}[
		colormap access=direct,
		width = 1. *\linewidth,
		height = 0.25 * \textheight,
    xmin=0,
		xmax=200,
		xtick = {0,50,...,200},
		scaled x ticks = false,
		xlabel={\small\sffamily Iteration},
		xlabel near ticks, xticklabel pos=lower,
    x label style={at={(0.5,-0.09)}},
		ymin=-463.5,
		ymax = -459.5,
		ylabel={\small\sffamily \phantom{Likelihood / Free Energy}},
		ylabel near ticks,
		yticklabel pos=left,
		grid = both,
   legend image code/.code={%
     \draw[dashed] (0cm,-0.025cm) -- (0.5cm,-0.025cm);
     \draw[solid]  (0cm, 0.025cm) -- (0.5cm, 0.025cm);
    },		
    legend entries={
      \fontsize{7}{0}\selectfont\sffamily \!\! $C'=1$ ($k$-means),
      \fontsize{7}{0}\selectfont\sffamily \!\! $C'=2$,
      \fontsize{7}{0}\selectfont\sffamily \!\! $C'=10$,
      \fontsize{7}{0}\selectfont\sffamily \!\! $C'=200$ (GMM)
    },
		legend style={
			at={(0.94,0.44)},
			legend columns=1,
			row sep=-2pt,
			column sep=2pt,
		},
		legend cell align=left,
    after end axis/.code={
      \node at (axis cs: -16,-459.55) {\fontsize{9}{0}\selectfont\sffamily\bfseries D};
    }
  ]
  \addlegendimage{color=red!70!black}  	
	\addlegendimage{color=cyan!70!black}
  \addlegendimage{color=green!70!black}  	
  \addlegendimage{color=black}
  %
  
  \newcommand\filename{./figs/KDD_k-means-C'_200_1.txt}
  \addplot [draw=none, stack plots=y, forget plot] table[
    x=n,
    y expr=\thisrow{F}-\thisrow{F_err}
  ] {\filename};	
  
  \addplot [draw=none, fill=red!70!black, stack plots=y, fill opacity=0.15] table [
      x=n,
      y expr=2*\thisrow{F_err}
  ] {\filename}\closedcycle;
  
  \addplot [forget plot, stack plots=y,draw=none] table [x=n, y expr=-(\thisrow{F}+\thisrow{F_err})] {\filename};
    
  \addplot [draw=none, stack plots=y, forget plot] table[
    x=n,
    y expr=\thisrow{L}-\thisrow{L_err}
  ] {\filename};	
  
  \addplot [draw=none, fill=red!70!black, stack plots=y, fill opacity=0.15] table [
      x=n,
      y expr=2*\thisrow{L_err}
  ] {\filename}\closedcycle;
  
  \addplot [forget plot, stack plots=y,draw=none] table [x=n, y expr=-(\thisrow{L}+\thisrow{L_err})] {\filename};

	\addplot [solid, mark=None, color=red!70!black] table[x=n, y=L] {\filename};	
	\addplot [dashed, mark=None, color=red!70!black] table[x=n, y=F] {\filename};

  \renewcommand\filename{./figs/KDD_k-means-C'_200_2.txt}
  \addplot [draw=none, stack plots=y, forget plot] table[
    x=n,
    y expr=\thisrow{F}-\thisrow{F_err}
  ] {\filename};	
  
  \addplot [draw=none, fill=cyan!70!black, stack plots=y, fill opacity=0.15] table [
      x=n,
      y expr=2*\thisrow{F_err}
  ] {\filename}\closedcycle;
  
  \addplot [forget plot, stack plots=y,draw=none] table [x=n, y expr=-(\thisrow{F}+\thisrow{F_err})] {\filename};
    
  \addplot [draw=none, stack plots=y, forget plot] table[
    x=n,
    y expr=\thisrow{L}-\thisrow{L_err}
  ] {\filename};	
  
  \addplot [draw=none, fill=cyan!70!black, stack plots=y, fill opacity=0.15] table [
      x=n,
      y expr=2*\thisrow{L_err}
  ] {\filename}\closedcycle;
  
  \addplot [forget plot, stack plots=y,draw=none] table [x=n, y expr=-(\thisrow{L}+\thisrow{L_err})] {\filename};

	\addplot [solid, mark=None, color=cyan!70!black] table[x=n, y=L] {\filename};	
	\addplot [dashed, mark=None, color=cyan!70!black] table[x=n, y=F] {\filename};

  \renewcommand\filename{./figs/KDD_k-means-C'_200_10.txt}
  \addplot [draw=none, stack plots=y, forget plot] table[
    x=n,
    y expr=\thisrow{F}-\thisrow{F_err}
  ] {\filename};	
  
  \addplot [draw=none, fill=green!70!black, stack plots=y, fill opacity=0.15] table [
      x=n,
      y expr=2*\thisrow{F_err}
  ] {\filename}\closedcycle;
  
  \addplot [forget plot, stack plots=y,draw=none] table [x=n, y expr=-(\thisrow{F}+\thisrow{F_err})] {\filename};
    
  \addplot [draw=none, stack plots=y, forget plot] table[
    x=n,
    y expr=\thisrow{L}-\thisrow{L_err}
  ] {\filename};	
  
  \addplot [draw=none, fill=green!70!black, stack plots=y, fill opacity=0.15] table [
      x=n,
      y expr=2*\thisrow{L_err}
  ] {\filename}\closedcycle;
  
  \addplot [forget plot, stack plots=y,draw=none] table [x=n, y expr=-(\thisrow{L}+\thisrow{L_err})] {\filename};

	\addplot [solid, mark=None, color=green!70!black] table[x=n, y=L] {\filename};	
	\addplot [dashed, mark=None, color=green!70!black] table[x=n, y=F] {\filename};

%
%
%
%
%
%

  \renewcommand\filename{./figs/KDD_GMM.txt}
  \addplot [draw=none, stack plots=y, forget plot] table[
    x=n,
    y expr=\thisrow{F}-\thisrow{F_err}
  ] {\filename};	
  
  \addplot [draw=none, fill=black, stack plots=y, fill opacity=0.15] table [
      x=n,
      y expr=2*\thisrow{F_err}
  ] {\filename}\closedcycle;
  
  \addplot [forget plot, stack plots=y,draw=none] table [x=n, y expr=-(\thisrow{F}+\thisrow{F_err})] {\filename};
    
  \addplot [draw=none, stack plots=y, forget plot] table[
    x=n,
    y expr=\thisrow{L}-\thisrow{L_err}
  ] {\filename};	
  
  \addplot [draw=none, fill=black, stack plots=y, fill opacity=0.15] table [
      x=n,
      y expr=2*\thisrow{L_err}
  ] {\filename}\closedcycle;
  
  \addplot [forget plot, stack plots=y,draw=none] table [x=n, y expr=-(\thisrow{L}+\thisrow{L_err})] {\filename};

	\addplot [solid, mark=None, color=black] table[x=n, y=L] {\filename};	
	\addplot [dashed, mark=None, color=black] table[x=n, y=F] {\filename};

  \end{axis}
\end{tikzpicture}
    \end{adjustbox}
  \end{subfigure}
  \begin{subfigure}[c]{\textwidth}
    \begin{adjustbox}{trim=0pt 0pt 0pt 0pt}
			\begin{tikzpicture}
				\node[] at (0,0){\fontsize{9}{0}\selectfont\sffamily\bfseries E};
			\end{tikzpicture}		
    \end{adjustbox}
    \begin{adjustbox}{trim=0pt 140pt 0pt 0pt}
			\rotatebox{90}{\resizebox{0.3\linewidth}{!}{
				\includegraphics[trim=5.5cm 5cm 7cm 7.5cm, clip=true,]{./figs/BIRCH-1-5-100_k-means_25-best.pdf}
			}}
    \end{adjustbox}
    \begin{adjustbox}{trim=0pt 0pt 0pt 0pt}
			\begin{tikzpicture}
				\node[] at (0,0){\fontsize{9}{0}\selectfont\sffamily\bfseries F};
			\end{tikzpicture}		
    \end{adjustbox}
    \begin{adjustbox}{trim=0pt 140pt 0pt 0pt}
			\rotatebox{90}{\resizebox{0.3\linewidth}{!}{
				\includegraphics[trim=3.5cm 3cm 3cm 3.5cm, clip=true,]{./figs/BIRCH_random-1-25-100_k-means_25-best.pdf}
			}}
    \end{adjustbox}
    \begin{adjustbox}{trim=0pt 0pt 0pt 0pt}
			\begin{tikzpicture}
				\node[] at (0,0){\fontsize{9}{0}\selectfont\sffamily\bfseries G};
			\end{tikzpicture}		
    \end{adjustbox}
    \begin{adjustbox}{trim=0pt 140pt 0pt 0pt}
			\rotatebox{90}{\resizebox{0.3\linewidth}{!}{
				\includegraphics[trim=3.5cm 3cm 3cm 3.5cm, clip=true,]{./figs/BIRCH_random-1-25-100_k-means-Cprime_25_2-best.pdf}
			}}
    \end{adjustbox}
  \end{subfigure}
  \vspace{145pt}
  \caption{\textsf{A} shows experiments of Alg.\,1 ($k$-means) on a BIRCH data set with grid-positioned clusters, as visualized in \textsf{E}. Shown are the log-likelihood and free energy per iteration for three individual runs (red, blue and green) and the mean of 100 independent runs (gray). The individual runs show convergences to different optima. 
    \textsf{B}~shows the same experiments as \textsf{A} on uniform randomly positioned clusters, as visualized in \textsf{F}.
		In \textsf{A} and \textsf{B}, the $D_{KL}$-gap at convergence is visualized as colored vertical lines next to the plot and can in \textsf{A} be clearly observed to increase for less optimal solutions.
		In \textsf{B}, due to higher cluster overlaps, the $D_{KL}$-gap is here overall larger compared to \textsf{A}.
    \textsf{C}~shows the mean log-likelihood (solid line) and the log-likelihood of the best run (striped) of 100 runs of Alg.\,3 ($k$-means-$C'$) for different $C'$.
		For $C'\geq 2$, the best solutions are close to identical for the different settings, although some tend to find these best solutions more frequently.
    \textsf{D}~shows the mean log-likelihood (solid line) and free energy (dashed) on the KDD data set over 10 runs, shaded with their respective SEMs.
    Visualization of some ground truth cluster centers (blue circles) and found cluster centers of the best runs (red crosses) on BIRCH data sets are shown in \textsf{E}, \textsf{F} (for $k$-means) and \textsf{G} (for $k$-means-$C'$ with $C'=2$).
    Comparison of \textsf{F} and \textsf{G} shows the difference between using $C'=1$ ($k$-means) and $C'>1$. Especially for regions with higher cluster overlap, $k$-means tends to push close-by clusters away from each other, due to the hard assignment of data points to only a single cluster. This effect can be observed on the groups of two and three clusters in the \changed{upper half} as well as on the group of clusters in the \changed{bottom} left corner. For $C'=2$, this effect is already greatly reduced.
    }
  \vspace{-10pt}    
	\label{fig:supp_ExpA}
\end{figure*}

\begin{figure*}[p]
  \begin{subfigure}[c]{\textwidth}
 		\includegraphics[width=0.65\textwidth]{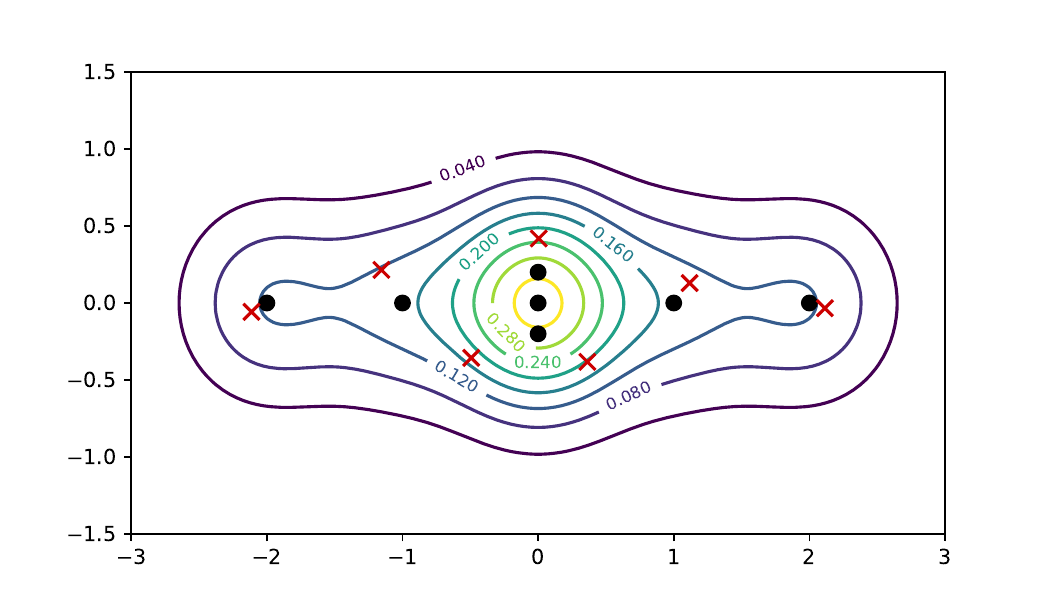}
    \begin{adjustbox}{trim=5pt 2pt 0pt 0pt}
\tikzsetnextfilename{Counterexample_k-means_plot}

\pgfplotsset{
	grid style={dotted,gray},
	minor grid style={dotted,lightgray},
  tick label style = {font=\tiny\sansmath\sffamily},
  legend style = {font=\sansmath\sffamily},
	/tikz/font=\sansmath\sffamily,
  xlabel style = {font=\sansmath\sffamily},
  ylabel style = {font=\sansmath\sffamily},
  every non boxed x axis/.style={} 	
}	

\begin{tikzpicture}
\begin{groupplot}[
    group style={
        group name=my fancy plots,
        group size=1 by 2,
        xticklabels at=edge bottom,
        vertical sep=0pt
    },
    width=0.33\linewidth,
    xmin=0,
    xmax=10,
		grid = both,    
		clip=false,
]

\nextgroupplot[
  ymin=-2.4,
  ymax=-2.25,
  ytick={-2.3,-2.4},
  axis x line=top, 
  height=3.cm * 1.05,
	legend style={
		at={(0.95,0.85)},
		legend columns=1,
		row sep=-2pt,
      inner sep=1.3pt,
	},
	legend cell align=left,
	legend entries={\hspace{1pt}\fontsize{6}{0}\selectfont\sffamily Likelihood},  
]
\addplot+ [line width=1pt, mark=None, color=blue!80!black] table[x=n, y=LL] {./figs/Counterexample_k-means.txt};

\nextgroupplot[
  ymin=-2.8,
  ymax=-2.55,
  ytick={-2.6,-2.7,-2.8},
  axis x line=bottom,
  axis y discontinuity=parallel,
  height=5.cm * 1.05,
	xlabel={\scriptsize\sffamily Iteration},
	xlabel near ticks, xticklabel pos=lower,
	legend style={
		at={(0.95,0.3)},
		legend columns=1,
		row sep=-2pt,
      inner sep=1.3pt,
	},
	legend cell align=left,
	legend entries={\hspace{1pt}\fontsize{6}{0}\selectfont\sffamily Free Energy},  
]
\addplot+ [dashed, line width=1pt, mark=None, color=red!80!black] table[x=n, y=FE] {./figs/Counterexample_k-means.txt};

\draw [white, fill=white] (axis cs:-0.1,-2.8045) rectangle (axis cs:10.1,-2.8055);

\end{groupplot}
\end{tikzpicture}
    \end{adjustbox}
  \caption{$k$-means}
  \end{subfigure}
  \begin{subfigure}[c]{\textwidth}
		\includegraphics[width=0.65\textwidth]{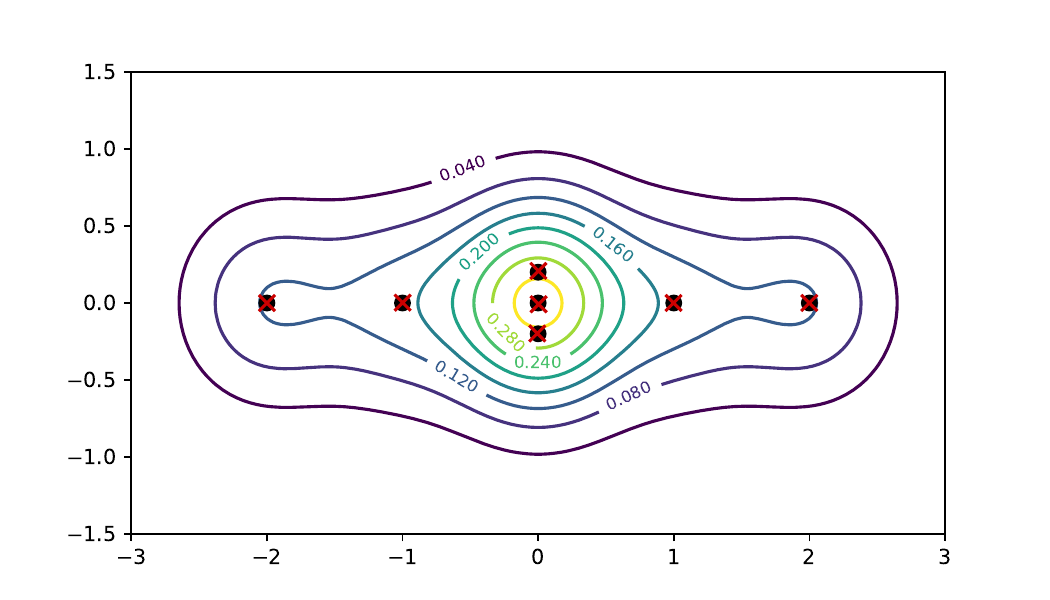}
    \begin{adjustbox}{trim=18pt 3pt 0pt 0pt}
\tikzsetnextfilename{Counterexample_GMM_plot}

\pgfplotsset{
	grid style={dotted,gray},
	minor grid style={dotted,lightgray},
  tick label style = {font=\tiny\sansmath\sffamily},
  legend style = {font=\sansmath\sffamily},
	/tikz/font=\sansmath\sffamily,
  xlabel style = {font=\sansmath\sffamily},
  ylabel style = {font=\sansmath\sffamily},
  every non boxed x axis/.style={} 	
}	

\begin{tikzpicture}
	\begin{axis}[
  width=0.33\linewidth,
  xmin=0,
  xmax=10,
	xlabel={\scriptsize\sffamily Iteration},
	xlabel near ticks, xticklabel pos=lower,
	grid = both,    
	clip=false,
  ymin=-2.26105,
  ymax=-2.26100,
  yticklabel style={/pgf/number format/fixed,
                  /pgf/number format/precision=8},
  ytick={-2.26100,-2.26101,-2.26102,-2.26103,-2.26104,-2.26105},
  height=6.9cm,
	legend style={
		at={(0.95,0.25)},
		legend columns=1,
		row sep=-2pt,
      inner sep=1.3pt,
	},
	legend cell align=left,
	legend entries={\hspace{1pt}\fontsize{6}{0}\selectfont\sffamily Likelihood,
	                \hspace{1pt}\fontsize{6}{0}\selectfont\sffamily Free Energy
	},  
]
\addplot+ [line width=1pt, mark=None, color=blue!80!black] table[x=n, y=LL] {./figs/Counterexample_GMM.txt};

\addplot+ [dashed, line width=1pt, mark=None, color=red!80!black] table[x=n, y=FE] {./figs/Counterexample_GMM.txt};

\end{axis}
\end{tikzpicture}
    \end{adjustbox}
  \caption{GMM}
  \end{subfigure}
  \caption{%
This example illustrates that the free energy and the GMM likelihood objective are not trivially related.
We generate data from seven overlapping, equal and isotropic Gaussians arranged as above (black circles), drawing 100\,000 data samples per Gaussian.
The contour lines show the underlying probability density distribution of which the data points are drawn.
We compare $k$-means in (a) with EM for isotropic (non-truncated) GMMs in (b).
For both, we use the ground-truth generating cluster centers (and variances for the GMM) as initialization.
If we now run $k$-means, we observe that while the free energy increases the log-likelihood decreases.
For this example the final cluster centers (red crosses) obtained by $k$-means differ very significantly from the ground-truth.
But also, e.g., for just two overlapping Gaussians, $k$-means results in final cluster centers significantly different from ground truth as can be observed in Fig.\,\ref{fig:supp_ExpA}\,(F/G).
The higher the cluster overlap, the more pronounced this effect becomes.
EM for GMM does on the other hand (as expected) result in final cluster centers (red crosses) very similar to ground truth (note the different scales of the plots; the higher initial likelihood value for the GMM compared to $k$-means is not due to different initial cluster centers, but due to a different $\sigma^2$-value for $k$-means as a result of applying Eq.\,(\ref{EqnSigmaKMeans}) with $k$-means activations).
Our example also provides a counterexample for the $k$-means objective~(\ref{EqnKMObjective}) and the likelihood objective for GMMs~(\ref{EqnGMMLikelihood}) \changedOld{giving rise to the same optimization problem}: here, the quantization error decreases, but the GMM likelihood gets worse.
The optimization of equally sized, isotropic GMMs (\ref{EqnGMMIso}) and of the $k$-means objective are sometimes regarded as equivalent; Feldman~et~al.~(2011), for instance, write ``[...] their result requires that the Gaussians are identical spheres, in which case the maximum likelihood problem is identical to the $k$-means problem''.
Also results of \citet[][]{Pollard1982}, who is often cited for showing that $k$-means is a GMM maximum likelihood estimator, seem to be misinterpreted sometimes.
$k$-means becomes an increasingly good maximum likelihood estimator if we additionally demand increasingly separable clusters. Increased separability is in turn
closely related to the $\sigma^2\rightarrow{}0$ limit, in which the $k$-means and GMM objectives become increasingly similar.\\[2mm] 
Feldman, D., Faulkner, M., Krause, A., 2011, Scalable training of mixture models via coresets, NIPS, 2142--2150.\\
Pollard, D., 1982. A central limit theorem for k-means clustering. The Annals of Probability, 919–926.
}
	\label{fig:supp_counterexample}
\end{figure*}

\section{Generalizations for lazy-$k$-means}
\label{sec:suppl_lazyKM}
If we change the cluster selection criterion (\ref{EqnEuclid}) to the criterion for {\em lazy-$k$-means} (\ref{EqnLazyKM}), then it follows from Proposition\,1 that each cluster assignment in lazy-$k$-means increases the free energy~(\ref{EqnTruncatedF}).
As the M-steps (equal to the $k$-means M-steps) then increase the free energy w.r.t.\ $\Theta$, it follows that lazy-$k$-means monotonically increases the same free energy objective.
Corollary\,1 does not apply but we can generalize Proposition\,2.\\[2mm]

\noindent{\bf Proposition} (Generalization of Proposition\,2 for lazy-$k$-means)\\
Consider the TV-EM algorithm (Alg.\,\ref{AlgTVEMforGMM}) but with criterion (\ref{EqnLazyKM}) instead of criterion (\ref{EqnEuclid}).
If we set $C'=1$, then the TV-EM updates of the cluster centers $\muVec_c$ (\ref{EqnMStep}) become independent of the variance $\sigma^2$ and are given by the lazy-$k$-means algorithm.\\[2mm]
{\bf Proof}\\
The proof is analogous to the one of Proposition\,2 with the only difference that $c_o^{(n)}$ is now a cluster of $\yVecN$ for which applies:
$\forall\ct\neq{}c_o^{(n)}: \|\yVecN\,-\,\muVec_{c_o^{(n)}}\| < (1+\eps)\,\|\yVecN\,-\,\muVec_{\ct{}}\|$.
The cluster assignments thus become those of lazy-$k$-means, while the  parameter updates remain those of standard $k$-means (i.e., the same as used for lazy-$k$-means).\\
\BOX\\[1mm]
As Propositions\,1 and 2 can be generalized, the fact that lazy-$k$-means optimizes the same free energy as $k$-means does also imply that Corollary\,2 can be used to relate lazy-$k$-means
to the GMM objective (\ref{EqnGMMLikelihood}).
\enlargethispage{\baselineskip}

\section{More details on the numerical experiments}
\label{sec:suppl_details_experiments}
\changed{Fig.\,\ref{fig:addExpA} shows additional results of $k$-means-$C'$ on the BIRCH data sets, namely the log-likelihood, quantization error, purity and NMI (where likelihood and purity
values are the same as those in Fig.\,\ref{ExpA}, but shown here again for easier comparison).
Tab.\,\ref{tab:comparison} gives a numerical comparison of these results to the DBSCAN and lazy-$k$-means algorithms.
The results on the quantization error compared to the likelihoods in Fig.\,\ref{fig:addExpA} and Tab.\,\ref{tab:comparison} highlight the fact that optimization of the $k$-means criterion (i.e., the quantization error) does generally not directly coincide with optimization of free energies by $k$-means-$C'$ with $C'>1$ (including optimization of likelihoods by EM for GMM for $C'=C$).
For the NMI and purity scores, we find that on the BIRCH set with random clusters (and therefore larger overlaps) $k$-means is prone to trade off NMI with decreasing purity scores (which results in a lower than average NMI score for the shown run with the highest combined score of NMI and purity).
The $k$-means-$C'$ algorithm, on the other hand, already results for $C'=2$ in high NMI and purity scores near the ground truth.\footnote{\changed{For the formulars of purity and NMI, see Manning et al. 2008, chapter: Evaluation of clustering, \url{https://nlp.stanford.edu/IR-book/html/htmledition/evaluation-of-clustering-1.html}\newline Manning, C. D., Raghavan, P. and Sch\"utze, H., 2008, Introduction to Information Retrieval, Cambridge University Press.}}}

\balance

Fig.\,\ref{fig:supp_ExpA} shows enlarged versions of \changed{plots A, D, E, G} of Fig.\,\ref{ExpA} with more details in the caption.
In addition to these results we also verified that the free energies (\ref{EqnTruncatedF}), (\ref{EqnFreeEnergyKMeans}) and the right-hand-side of (\ref{EqnResult}) are numerically equal for $k$-means.
For $k$-means-$C'$ we verified that free energies (\ref{EqnTruncatedF}) and (\ref{EqnFreeEnergyKMeansC}) 
are equal at convergence.

Note that Fig.\,\ref{fig:supp_ExpA} can also be interpreted as numerically verifying that the free energies and the likelihood objective of the GMM (\ref{EqnGMMIso}) are not trivially related. This
includes the free energy (\ref{EqnFreeEnergyKMeans}) which is optimized by $k$-means and which corresponds to $C'=1$. Comparison of the means of Figs.\,\ref{fig:supp_ExpA}(F) and (G) already shows the difference
when comparing results between $C'=1$ ($k$-means) to $C'=2$. Finally, the numerical experiment of Fig.\,\ref{fig:supp_counterexample} is deliberately chosen to highlight the difference between the
$k$-means objective (\ref{EqnKMObjective}) and the GMM log-likelihood (\ref{EqnGMMLikelihood}). By applying $k$-means, the $k$-means free energy increases (the quantization error gets smaller) but the log-likelihood gets worse. Results for the cluster centers recovered by $k$-means and EM for GMM (\ref{EqnGMMIso}) are very different.

\end{document}